# 3D Video Quality Assessment

by

Amin Banitalebi Dehkordi

B.Sc., University of Tehran, 2008

M.Sc., University of Tehran, 2011

A THESIS SUBMITTED IN PARTIAL FULFILLMENT OF
THE REQUIREMENTS FOR THE DEGREE OF

**Doctor of Philosophy**

in

THE FACULTY OF GRADUATE AND POSTDOCTORAL STUDIES

(Electrical & Computer Engineering)

THE UNIVERSITY OF BRITISH COLUMBIA

(Vancouver)

August 2015

# Abstract

A key factor in designing 3D systems is to understand how different visual cues and distortions affect the perceptual quality of 3D video. The ultimate way to assess video quality is through subjective tests. However, subjective evaluation is time consuming, expensive, and in most cases not even possible. An alternative solution is objective quality metrics, which attempt to model the Human Visual System (HVS) in order to assess the perceptual quality. The potential of 3D technology to significantly improve the immersiveness of video content has been hampered by the difficulty of objectively assessing Quality of Experience (QoE). A no-reference (NR) objective 3D quality metric, which could help determine capturing parameters and improve playback perceptual quality, would be welcomed by camera and display manufactures. Network providers would embrace a full-reference (FR) 3D quality metric, as they could use it to ensure efficient QoE-based resource management during compression and Quality of Service (QoS) during transmission.

In this thesis, we investigate the objective quality assessment of stereoscopic 3D video. First, we propose a full-reference Human-Visual-system based 3D (HV3D) video quality metric, which efficiently takes into account the fusion of the two views as well as depth map quality. Subjective experiments verified the performance of the proposed method. Next, we investigate the No-Reference quality assessment of stereoscopic video. To this end, we investigate the importance of various visual saliency attributes in 3D video. Based on the results gathered from our study, we design a learning based visual saliency prediction model for 3D video. Eye-tracking experiments helped verify the performance of the proposed 3D Visual Attention Model (VAM). A benchmark dataset containing 61 captured stereo videos, their eye fixation data, and performance evaluations of 50 state-of-the-art VAMs is created and made publicly available online. Finally, we incorporate the saliency maps generated by our 3D VAM in the design of the state-of-the-art no-reference (NR) and also full-reference (FR) 3D quality metrics.

# Preface

This thesis presents research conducted by Amin Banitalebi Dehkordi, under the guidance of Dr. Panos Nasiopoulos. A list of publications resulting from the work presented in this thesis is provided on the following page.

The main body of Chapter 2 is taken from our previous publications in [P1]-[P11]. A provisional patent application was filled based on the material of this chapter [P4] and a contribution was made to MPEG video compression standardization activities [P5]. The content of Chapter 3 appears in one conference [P12] and one journal publication [P13]. Portions of Chapter 4 appear in [P14]-[P16] while the main body of Chapter 5 appears in [P17]. The work presented in all of these manuscripts was performed by Amin Banitalebi Dehkordi, including literature review, designing and implementing the proposed algorithms, performing all experiments, analyzing the results and writing the manuscripts. Dr. M. T. Pourazad provided editorial input and high-level consultation. E. Nasiopoulos helped with the eye-tracking experiments in [P15] and Y. Dong provided intermediate data files for work conducted in [P16]. The entire work was conducted under the supervision and with editorial input from Dr. Panos Nasiopoulos. The first and last chapters of this thesis were written by Amin Banitalebi Dehkordi, with editing assistance from Dr. Nasiopoulos.

List of Publications Based on Work Presented in This Thesis

[P1] A. Banitalebi-Dehkordi, M. T. Pourazad, and P. Nasiopoulos, "A human visual system based 3D video quality metric," $2^{nd}$ International Conference on 3D Imaging, IC3D, Dec. 2012, Belgium.

[P2] A. Banitalebi-Dehkordi, M. T. Pourazad, and P. Nasiopoulos, "3D video quality metric for mobile applications," $38^{th}$ International Conference on Acoustic, Speech, and Signal Processing, ICASSP, May 2013, Vancouver, Canada.

[P3] A. Banitalebi-Dehkordi, M. T. Pourazad, and P. Nasiopoulos, "3D video quality metric for 3D video compression," $11^{th}$ IEEE IVMSP Workshop: 3D Image/Video Technologies and Applications, June 2013, Seoul, Korea.

[P4] 61/731,864 "An efficient human visual system based quality metric for 3D video," A. Banitalebi-Dehkordi, M. T. Pourazad, and P. Nasiopoulos, US Provisional Patent Application, filed Nov. 30, 2012.

# Table of Contents

# List of Tables

# List of Figures

# List of Acronyms

| | |
|---|---|
| ALF | Adaptive Loop Filter |
| AUC | Area Under the Curve |
| AVC | Advanced Video Coding |
| CSF | Contrast Sensitivity Function |
| DCT | Discrete Cosine Transform |
| DERS | Depth Estimation Reference Software |
| DoG | Differences of Gaussians |
| EDISON | Edge Detection and Image Segmentation |
| EMD | Earth Movers Distance |
| FDM | Fixation Density Map |
| FPR | False Positive Rate |
| GOP | Group Of Pictures |
| HD | High Definition |
| HDR | High Dynamic Range |
| HEVC | High Efficiency Video Coding |
| HVS | Human Visual System |
| JNB | Just Noticeable Blur |
| kbps | Kilobits per second |
| KLD | Kullback-Leibler Divergence |
| MAD | Mean of Absolute Differences |
| NCJ | Numerical Categorical Judgement |
| NSS | Natural Scan-path Saliency |
| MOS | Mean Opinion Score |
| MPEG | Moving Picture Experts Group |
| MSE | Mean Square Error |
| OR | Outlier Ratio |
| PCC | Pearson Correlation Coefficient |
| PSNR | Peak Signal to Noise Ratio |
| QP | Quantization Parameter |
| RDOQ | Rate-Distortion Optimized Quantization |
| RF | Random Forests |
| RMSE | Root Mean Square Error |
| ROC | Receiver Operating Characteristics |
| SAO | Sample Adaptive Offset |
| SCC | Spearman Correlation Coefficient |
| SI | Spatial Information |
| STD | Standard Deviation |
| SVM | Support Vector Machine |
| SVR | Support Vector Regression |
| TI | Temporal Information |
| TPR | True Positive Rate |
| VAM | Visual Attention Model |
| VCEG | Video Coding Experts Group |
| VSRS | View Synthesis Reference Software |

# Acknowledgements

I offer my enduring gratitude to my supervisor, Dr. Panos Nasiopoulos, for his exceptional guidance and constant support throughout my PhD studies. I thank Dr. Nasiopoulos for his encouragement, enthusiasm, motivation, insightful suggestions, and for enlarging my vision of science and life. Panos has always been a great mentor, a role model, and a friend.

I also give my sincere thanks to my colleagues at the Digital Media Lab, Dr. Mahsa T. Pourazad, Dr. Lino Coria, Dr. Di Xu, Sima Valizadeh, Anahita Shojae, Basak Oztas, Maryam Azimi, Mohsen Amiri, Hamidreza Tohidypour, Bambang Sarif, Pedram Mohammadi, Stelios Ploumis, Ilya Ganelin, Fujun Xie, and all other lab mates. It was a pleasure working with you all in such a vibrant and eclectic environment.

I thank my friends in the Electrical and Computer Engineering department, Mohammad Ghasemi, Hossein Bashashati, Hamid Palangi, Burak Yoldemir, and others.

To my friends in Vancouver who added different dimensions to my life, Nasim and Zainab Zolaktaf.

I would like to thank the Natural Sciences and Engineering Research Council of Canada (NSERC) and the University of British Columbia for providing financial support in the form of research grants, scholarships, and tuition awards.

Last but not least, I thank my family for their constant love and unconditional support. "Family is not an important thing. It's everything."

At the end, thanks to you, reader. If you have made it this far, you at least read one page of my thesis. Thank You.

*"Nothing in the world can take the place of persistence. Talent will not; nothing is more common than unsuccessful men with talent. Genius will not; unrewarded genius is almost a proverb. Education will not; the world is full of educated derelicts. Persistence and determination alone are omnipotent. The slogan, 'press on' has solved, and always will solve, the problems of the human race."*

Calvin Coolidge

*To my family*

# 1  Introduction and Overview

3D video technologies have been introduced to the consumer market in the past few years. As these technologies mature, they will not only affect specialized fields such as the entertainment, education, training, and medical imaging industries, but also change the quality of the viewing experience of the average consumer by bringing life-like video into gaming, theater, and television. Delivering the highest possible quality of experience to end-users is crucial in 3D content creation and delivery.

The ultimate way to assess 3D video quality is through subjective tests. However, subjective evaluation is time consuming, expensive, and in some cases not possible. An alternative solution is developing objective quality metrics, which attempt to model the Human Visual System in order to assess perceptual quality. While in some applications e.g., 3D video compression, a reference video of original quality is available, in other applications such as 3D cameras, set-top boxes, 3D TVs, or 3D Cinema, there is no reference to compare against. Objective quality assessment metrics, in general, are categorized as Full-Reference (when a reference video is available), No-Reference (when no reference video is available), and Reduced-Reference (when only partial information or features from the reference video are available). Although several 2D quality metrics have been proposed for still images and videos, in the case of 3D efforts are only at the initial stages.

This thesis investigates the Full-Reference (FR) and No-Reference (NR) quality assessment of stereoscopic 3D video. In Chapter 2, a FR quality metric for stereo video is designed and its performance is verified through subjective user studies. Chapters 3 and 4 are related to NR quality assessment, where we first study the various attributes contributing to NR 3D quality. More specifically, in Chapter 3 we investigate the effect of the frame rate on 3D video quality and bitrate. We combine the knowledge acquired from these experiments with several other 3D saliency attributes to design a 3D Visual Attention Model (VAM) in Chapter 4. We conduct eye-tracking experiments to evaluate the performance of the proposed VAM. We also create a benchmark 3D eye-tracking database containing a large scale stereo video dataset, the corresponding eye-tracking data, and the performance evaluation of 50 existing VAMs over this dataset. In Chapter 5, we incorporate the 3D saliency information to modify the design of existing 3D quality metrics. We show in Chapter 5 that using the saliency information in quality assessment not only significantly boosts the performance of NR 3D video quality metrics, but it also improves the quality assessment performance of FR 3D video quality metrics.

The following sections in this introductory chapter provide a literature review of the topics addressed in each of the research chapters. Section 1.1 reviews the state-of-the-art in FR and NR 3D video quality metrics. Existing studies on the effect of frame rate on video quality are reviewed in Section 1.2. Section 1.3 provides an overview of the existing 3D saliency prediction models as well as the available eye-tracking datasets. Section 1.4 reviews the recent saliency inspired FR and NR quality metrics. Finally, Section 1.5 concludes the introduction with an overview of the research contributions presented in this thesis.

## 1.1 Overview of Existing 3D Video Quality Metrics

With the introduction of 3D technology to the consumer market in the recent years, one of the challenges industry has to face is assessing the quality of 3D content and evaluating the viewer's quality of experience (QoE). While several accurate quality metrics have been designed for 2D content, there is still room for improvement when it comes to objective assessment of 3D video quality. Assessing the quality of 3D content is much more difficult than that of 2D. In the case of 2D, there are well-known factors such as brightness, contrast, and sharpness that affect perceptual quality. In the case of 3D, depth perception changes the impact that the above factors have on the overall perceived 3D video quality. Although the effect of these factors on 2D video quality has been extensively studied, we have a limited understanding of how these factors affect 3D perceptual quality. In addition, 3D related factors such as the scene's depth range, display size and the type of 3D display technology used (i.e., active or passive glasses, glasses-free auto-stereoscopic displays, etc.) solely affect 3D video quality, while they have no effect on 2D video quality [1]-[3]. The study presented by Seuntiens [4] identifies some additional 3D quality factors such as "presence" and "naturalness" that also affect 3D perception. These factors are of particular interest nowadays, since they play an important role in the design and evaluation of interactive media. Chen et al. [5] suggest including two other 3D quality factors, "depth quantity" and "visual comfort", in the overall quality of 3D content. The work presented in [6] and [7] shows that 3D quality factors have closer correlation with the overall 3D video quality compared to 2D quality factors. Considering that the existing 2D quality metrics are entirely based on 2D quality factors, it is not surprising that they are not accurate when used for evaluating 3D quality [8]-[11]. This has been verified by applying existing 2D quality metrics on the right and left views separately, and averaging the values over two views, then comparing the results with subjective evaluations [8]-[11]. The rest of this section provides a brief survey on the state-of-the-art FR and NR quality metrics for stereoscopic video.

*1.1.1 Existing FR 3D Quality Metrics*

To account for the effect of 3D quality factors when evaluating the quality of 3D content, Ha and Kim [12] proposed a 3D quality metric that is solely based on the temporal and spatial disparity variations. However, the proposed metric does not include the effect of 2D-associated quality factors such as contrast and sharpness. It is known that the overall perceived 3D quality is dependent on both 3D depth perception (3D factors) and the general picture quality (2D factors) [1]-[3]. In [12], the Mean Square Error (MSE) is used as a quality measure, which is known to have low performance in accurately representing the human visual system [13]-[14]. Moreover, this method utilizes disparity information instead of the actual depth map, which may result in inaccuracies in the case of occlusions. In addition, when using disparity instead of depth, the same amount of disparity may correspond to different perceived depths, depending on the viewing conditions. Boev et al. categorized the distortions of 3D content to monoscopic and stereoscopic types and proposed separate metrics for each type of distortions [15]. In this approach, the monoscopic quality metric quantitatively measures the distortions caused by blur, noise and contrast-change, while the stereoscopic metric exclusively measures the distortions caused by depth inaccuracies. The main drawback of this approach is that it does not attempt to fuse the 2D and 3D associated factors into one index, which in turn limits the accuracy of measuring 3D quality.

Considering that the '3D quality of experience' refers to the overall palatability of a stereo pair, which is not limited to image impairments, some quality assessment studies propose to combine a measure of depth perception with the quality of the individual views. The study in [16] formulates the 3D quality as an efficient combination of the depth map quality and the quality of individual views. The depth map quality is evaluated based on the error between the squared disparities of the reference 3D pictures and that of the distorted one and the quality of each view is measured via structural similarity (SSIM) index [13]. Another approach forms a 3D quality index by combining the SSIM index of individual views and the Mean Absolute Difference (MAD) between the squared disparities of the reference 3D content and the distorted one [17]. In both methods presented in [16] and [17], the quality of the individual views is directly used to assess the overall 3D quality. However, when watching 3D, the brain fuses the two views to a single mental view known as cyclopean view. This suggests incorporating the quality of cyclopean view instead of that of individual views in order to design more accurate 3D quality metrics [18]-[19]. To this end, Shao et al. [20] considered the binocular visual characteristics of the Human Visual System (HVS) to design a full-reference image quality metric. In this approach, a left-right consistency check is performed to classify each view to non-corresponding, binocular fusion, and binocular suppression regions. Quality of each region is evaluated separately using the local amplitude and phase features of the reference and distorted views and then combined into an overall score [20]. In another work, Chen et al. propose a full-reference 3D quality metric, which assesses the quality of the cyclopean view image instead of individual right and left view images. In their approach the cyclopean view is generated using a binocular rivalry model [18]. The energy of Gabor filter bank responses on the left and right images is utilized to model the stimulus strength and imitate the rivalrous selection of cyclopean image quality. This work was later extended in [21] by taking into account various 2D quality metrics for cyclopean view quality evaluation and generating disparity map correspondences. The results of this study show that by taking into account the binocular rivalry in the objective 3D content quality assessment process, the correlation between the objective and subjective quality scores increases, especially in the case of asymmetrically distorted content [21]. Similarly, in the study by Jin et al. the quality of cyclopean view was taken into account in the design of a full-reference 3D quality metric for mobile applications, which is called PHVS3D [19]. In this case, the information of the left and right channels is fused using the 3D-DCT transform to generate a cyclopean view. Then, a map of local block dissimilarities between the reference and distorted cyclopean views is estimated using the MSE of block structures. The weighted average of this map is used as the PHVS3D quality index. Although the proposed schemes in [18] and [19] take into account the quality of cyclopean view, they ignore the depth effect of the scene. The quality of cyclopean view on its own does not fully represent what is being perceived from watching a stereo pair as it only reflects the impairment associated with the cyclopean image. The overall human judgment of the 3D quality changes depending on the scene's depth level [4]-[7]. The impairments of 3D content are more noticeable if they occur in areas where depth level of the scene changes. Thus, a measure of depth quality in addition to the quality of cyclopean view has to be taken to account in the design of a 3D quality metric. One example of such approach is the full-reference PHSD quality metric proposed by Jin et al. [9]. Similar to PHVS3D, PHSD fuses the information of the left and right views to simulate the cyclopean view and measure its quality. Then to take into account the effect of depth, the MSE between the disparity maps of the reference and distorted 3D content as well as the local disparity variance are incorporated. PHSD has been specifically designed for video compression applications on mobile 3D devices. Although PHSD has been proposed for evaluating the quality of 3D video, it does not utilize a temporal pooling strategy to address the effect of temporal variations of the quality in the video. The temporal pooling schemes map a series of fidelity scores associated to different frames to a single quality score that represents an entire video sequence [22]. Most of the existing full-reference stereoscopic quality metrics are either designed for 3D images or do not utilize temporal pooling.

### 1.1.2 Existing NR 3D Quality Metrics

No reference quality assessment is generally a much more difficult task than full reference quality assessment, as no information is available about the reference data. As a consequence, NR quality metrics usually aim at evaluating the quality when only a specific type of distortion is present. Compared to FR 3D video quality metrics, there is smaller number of NR 3D metrics proposed in the literature [23]-[28]. Akhtar et al. proposed a NR quality assessment framework for JPEG coded stereoscopic images based on segmented local features of artifacts and disparity [23]. This method takes into account the blockiness and blur at the location of edge and non-edge areas. While subjective experiments by the authors show the efficiency of this metric, the practicality of this approach is limited as it is only applicable to JPEG coded images (only blur and blockiness are considered). In addition, this method is for stereo images and does not consider the temporal aspects of quality. Shao and colleagues proposed a no-reference stereoscopic image quality assessment (NR-SIQA) method by extracting distortion-specific features [24]. In a similar work, Chen et al. proposed a NR quality metric for stereo images by extracting natural statistics from the views and a generated cyclopean view picture [25]. Both of these methods consider Gaussian blur, White noise, and JPEG and JPEG2000 compression and utilize Support Vector Regression (SVR) for quality assessment. In another work by Hasan et al., a NR video quality metric for stereo images, QA3D, is designed to assess transmission artifacts (blockiness, sharpness, edginess) [26]. In this method, the difference between average disparity values for consecutive frames is used to define a dissimilarity index.

In addition to the mentioned NR metrics, there are also a few studies that investigate the usage of saliency prediction in NR quality assessment. Gu et al. proposed a saliency inspired parallax compensation based distortion metric (NOSPDM) for JPEG compressed stereoscopic images [27]. This method uses the correlation between 2D saliency maps of the views as a term in the overall formulation of NR quality. In another work, Ryu and Sohn proposed a NR quality metric for stereoscopic images, which takes into account blurriness and blockiness of an image pair [28]. In this approach, a pair of blurriness and blockiness maps are generated for each view, and combined with 2D saliency maps generated from each of the views. 2D saliency information is incorporated as a weighting factor to the blurriness and blockiness maps.

A major drawback of all mentioned approaches is that they are designed for stereo images and the experiments are based on 2D image domain distortions. The use of these metrics for video quality assessment is, therefore, not verified. In addition, distortions associated to 3D video delivery pipeline, i.e., 3D video compression, view synthesizing artifacts, depth map compression, and depth map packet loss, are not considered.

## 1.2 Overview of State-of-the-Art Studies on Video Frame Rate

In order to achieve the best possible 3D viewing experience, the factors that attribute to 3D quality need to be carefully taken into consideration throughout the 3D content delivery pipeline (capturing, transmission, and display). There are several factors and parameters that affect the perceptual quality of 3D content. Some of these factors are explicit to 3D content and do not affect the quality of 2D content or do not exist in the case of 2D, such as disparity, display size, 3D display technology (active, passive, glasses-free), and binocular properties of Human Visual System. Other factors are common attributes between 2D and 3D, but may have different effect on these two types of media, such as brightness [29] and color saturation [30]. Although fast moving objects may have a negative impact in the visual quality of 2D content, the quality degradation effect caused by fast motion seems to be significantly magnified in the case of 3D [31][32]. The reason is that fast motion in 3D results in rapid change in the perceived depth of objects, if the motion direction is perpendicular to the screen. This in turn leads to fast decoupling of vergence and accommodation (which is the main source of visual fatigue when watching 3D), resulting in degradation of the overall 3D quality of experience [31],[32].

Considering that motion of objects cannot be controlled in live 3D videos and that in the case of movies, motion might be one of the key elements of the story line, the need for new tools/methods that make motion of objects appear smoother and improve the 3D viewing Quality of Experience (QoE) has become noticed by the research community. To address this need, recently the film industry has introduced higher frame rates for 3D video capturing [33].

Existing studies and subjective evaluations on 2D videos show that the subjective quality of 2D videos with standard conventional frame rates (25 fps (PAL), 30 fps (NTSC), and 24 fps (cinema films)) is slightly worse than the ones with higher frame rates, increasing with the amount of motion [35]-[40]. However, in the case of stereoscopic video, higher frame rates than the traditional 24 frames per second (fps), yield sharper, less blurred, and more natural content, resulting in significantly improved overall viewing experience [34].

While the effect of frame rate on 2D perceptual video quality has been studied for many years [35]-[40], in the case of 3D there are still unanswered questions. The main question is identifying what frame rate yields the best visual quality for 3D. Recent feedback on 3D movies captured at 48 fps indicates improved 3D viewing experience [34],[41], but the question that remains is how much improvement is achieved by increasing the frame rate from 24 fps to 48 fps, or if any frame rate increase above 48 fps will result in additional visual quality improvement. For the case of 3D content broadcasting, in addition to the questions regarding the impact of frame-rate increase on 3D video quality, there are questions on bandwidth requirements for the transmission of high frame rate 3D content. Considering that the required bandwidth for the transmission of 3D video is generally higher than that of 2D video, it is important to perform feasibility studies on high frame rate 3D content transmission and come up with bitrate adaptation guidelines for variable bitrate networks. These guidelines, similar to the existing ones for the transmission of 2D video [35]-[40],[42]-[44], will help adjust the content frame-rate adaptively (dropping frame rate or frame rate up-conversion), according to channel capacity and required video quality. Limited work has been done in this regard for 3D video and, therefore, there is still room for improvement [45],[46].

## 1.3 Existing Literature on 3D Visual Attention Modeling

When watching natural scenes, an overwhelming amount of information is delivered to the human eye, with the optic nerve receiving an estimated $10^8$ bits of information per second [47]. In order for the human visual system to process this volume of visual data, it separates the data into pre-attentive and attentive levels [48]. The former is responsible for identifying the regions worth of attention, while the latter involves in-depth processing of limited portions of the visual information [48].

In computer vision, there is a strong interest in designing models inspired by HVS that narrow down a large amount of visual data to a smaller amount of more visually important data. Generally, eye-tracking experiments are used to help us understand what catches human attention in a scene. However, eye-tracking devices are not a viable option in many automated applications. Instead, visual attention models have been developed to mimic the layered perception mechanism of the human visual system by automatically detecting Regions Of Interest (ROIs) in a scene.

Psychological findings suggest that, in the pre-attentive stage, the visual information of a scene is represented by several retinotopic maps, each of them illustrating one visual attribute [49]. These attributes along with higher-level scene-dependent information are then analyzed by the visual cortex. Motivated by this fact, visual attention models also predict the locations of salient regions using three different approaches: bottom-up, top-down, and integration of the two. Bottom-up saliency detection models adopt rapid low-level visual attributes such as brightness, color, motion, and texture to generate a stimulus driven saliency map. Top-down approaches, however, utilize high-level context-dependent information such as humans, faces, animals, cars, and text for saliency detection in specific tasks. Integrated methods utilize bottom-up and top-down attributes for saliency detection [50].

There has been a great deal of research done in the field of 2D images and video saliency analysis that resulted in developing many successful visual attention models for 2D content [51]-[67]. However, two-dimensional VAMs are usually not accurate enough in predicting the salient regions in 3D content, as they do not incorporate depth information [68]-[71]. One reason is that depth perception changes the impact of the 2D visual saliency attributes (e.g., brightness, color, texture, motion). Also, there are several other visual attributes such as depth range, display size, the technology used in 3D display (i.e., active or passive glasses, glasses-free auto-stereoscopic displays, etc.), naturalness [4], and visual comfort [5], that solely affect 3D attention while they don't have any impact on 2D visual attention [1],[3]. As a result, in order to truly mimic the visual attention analysis of the HVS, we need to use 3D-exclusive saliency prediction mechanisms. The points made above and the rapid expansion of 3D image and video technologies emphasize the necessity to either extend the current 2D saliency detection mechanisms to 3D data, or develop novel 3D-specific saliency prediction methods.

In the following subsections, we give an overview of the state-of-the-art VAMs as well as the most recent eye-tracking datasets.

*1.3.1  Overview of the State-of-the-Art 3D VAMs*

The existing literature for 3D saliency prediction offers two main groups of solutions. The first (earliest attempts for 3D saliency prediction) directly uses the depth map (or disparity map) as a weighting factor in conjunction with an existing 2D saliency detection model. In other words, a 2D saliency map is created first, then each pixel (or region) in the resulting map is assigned a weight according to its disparity value. Maki et al. [72], Zhang et al. [73], and Chamaret et al. [74] used this approach to design their computational model for 3D saliency. These methods are based on the idea that generally the objects that are closer to the observer are considered to be more salient. Although they observed qualitative improvements compared to 2D saliency mechanisms, they didn't provide a quantitative evaluation of the proposed methods. Moreover, objects closer to viewers are not necessarily more salient.

The second group of solutions for 3D visual attention prediction utilizes the depth information of a scene to create a depth saliency map. The depth saliency map is usually combined with the 2D conspicuity maps (using the existing visual attention models) to construct a computational model of 3D visual saliency. Using this approach, Ouerhani and Hugli [75] proposed an attention model that takes into account the depth gradient features as well as the surface curvature. They performed qualitative assessment, but no quantitative assessment against eye-tracking data was done. Lang et al. [76] proposed a depth saliency map in which they evaluated the statistical probability of saliency ratio at different depth ranges using a training database. To validate their method, they also integrated the resulting depth saliency map to some other 2D models by summation or element-wise multiplication. Wang et al. [69] incorporated a Bayesian approach of depth saliency map generation and combined their map with some existing 2D models through averaging. Fang et al. [68] proposed a computational model of saliency for stereoscopic images by taking into account four different attributes: brightness, color, texture, and depth. They partitioned each image into patches and considered DC and AC coefficients of the DCT transform of each patch as its corresponding features. They generated several feature maps and linearly combined them with an emphasis on the compactness property of feature maps. Unlike the above methods that were designed for stereo images, Kim et al.'s work [71] is among a few saliency prediction models, which were proposed for stereoscopic videos. They adopted a scene type classification mechanism and incorporated several saliency attributes as well as concepts like saliency compactness, depth discontinuities, and visual discomfort. The generated feature maps were combined through summation or element-wise multiplication. It is common practice for saliency prediction methods to calculate various feature maps and then average them into one final map. However, it is not exactly known how the human brain fuses the different visual attributes. Examining the importance of each of the features and determining how to properly fuse them to closely imitate the human visual system remains a challenge.

*1.3.2  Overview on the Existing Eye-Tracking Datasets*

Eye-tracking datasets are usually used as ground-truth to validate the performance of different saliency prediction methods. In a recent study by Engelke et al., it was found that Fixation Density Maps (FDMs) resulting from independent eye-tracking experiments were very similar [77]. In other words, independently conducted eye-tracking experiments provided similar results for saliency prediction, quality assessment, and image retargeting purposes. This suggests that a benchmark eye-gaze dataset can be robustly used as a reference ground-truth point for various applications.

The research community has made publicly available over a dozen 2D image and video datasets, to facilitate testing the performance of automatic VAMs for predicting human fixation points [78]-[102]. In the case of 3D, however, there are very few stereoscopic image datasets available [69][76],[103]. While there are plenty of eye-tracking datasets available for 2D VAM studies, there is only a couple of stereoscopic video dataset available so far, which contain 8 and 47 stereoscopic sequences [104],[105]. The lack of such 3D datasets is an additional obstacle in evaluating and comparing 3D-VAMs and saliency prediction mechanisms for 3D video content. Available eye-tracking datasets are categorized as: 2D image/video and 3D image/video datasets. In this section we introduce the representative datasets for each category.

*1.3.2.1  2D Image Eye-Tracking Datasets*

One of the first eye-tracking datasets for 2D images is the "IRCCyN Image 1" [106] dataset, which was prepared using 27 scenes and 40 users in a free-viewing condition. Since then, several other eye-tracking experiments were conducted and collected results are made available to the public. Among those, the MIT "LowRes" [85], "CVCL" [79], "CSAIL" [53], NUSEF [78], "MIT Benchmark" [83], "McGill ImgSal" [86], and FiFa [87] are the ones that contain a large number of scenes (over 250 images). Several other datasets with a lower number of scenes are also available [84], e.g., "GazeCom Image" [88], "IRCCyN Image 2" [89], KTH [90], "LIVE DOVES" [91], Toronto [63],[92], "TUD Image 1&2" [93],[94], "TUD Interactions" [95], and VAIQ [96].

*1.3.2.2   2D Video Eye-Tracking Datasets*

In addition to the 2D image banks, there are also several public 2D video databases for visual attention studies. These datasets mainly vary in the number of video sequences and resolutions. The "Actions" [81], DIEM [97], "IRCCyN Video 1 & 2" [98],[99], "TUD Task" [100], "USC CRCNS Original" [101], "USC CRCNS MTV" [82], and "USC VAGBA" [102] datasets contain a large number of 2D video sequences (over 50), while ASCMN [107], "GazeCom Video" [88], and SFU dataset [80] contain a lower number of videos. To the best of our knowledge, the only large-scale high definition (1080×1920) 2D video eye-tracking dataset presently available is the "USC VAGBA" dataset. This dataset was originally prepared for validation purposes in a study on visual attention guided video compression [102].

*1.3.2.3   3D Image Eye-Tracking Datasets*

There are only a few 3D eye-tracking datasets publically available. The NUS3D dataset collected human eye fixations from 600 stereopairs viewed by 80 subjects [76]. To create this dataset, a Kinect sensor was utilized to capture depth at 640×480 resolution. The depth maps were then used along with a captured color view (left view) to synthesize the right view. The purpose of this study was to validate a visual attention model for 3D stereopairs. The "3DGaze database" is another source for eye tracking information of 3D stereopairs. This dataset contains gaze information of 18 stereoscopic images at various resolutions viewed by 35 users [69]. The stereoscopic images used in this dataset were captured using 3D cameras. Disparity maps were later generated by automatic disparity map generation algorithms. In a study by Khaustova et al. [103], a dataset of stereopairs was introduced that contains eye fixation data corresponding to 54 captured stereoscopic images at full-HD resolution (1920×1080) viewed by 15 subjects.

This dataset was then used to investigate the possible changes in visual attention with respect to depth and texture variations.

*1.3.2.4   3D Video Eye-Tracking Datasets*

To the best of our knowledge, to this date, the only publicly available 3D video eye-tracking datasets are the EyeC3D [104] and the IRCCyN [105] datasets, which contain 8 and 47 stereoscopic videos, respectively.

## 1.4  Existing Saliency Inspired Quality Assessment Methods

One of the major applications of VAMs is in the design of quality metrics where they guide the quality assessment techniques towards the most salient regions of an image or video. Then, quality metrics treat visible distortions in the salient regions differently from the ones existing in the non-salient regions [108]-[119].

Designing VAMs for 2D video goes back to over two decades ago and their applications are well explored [51]-[67]. Saliency measurement is already being integrated into various Full-Reference and No-Reference quality metrics. However, since 3D video technologies have entered the consumer market only in the past few years, the existing literature on 3D VAMs and their applications is yet to be as complete as the 2D case. Consequently, the application of 3D VAMs in designing 3D video quality metrics needs to be addressed. This makes more sense by considering the fact that quality measurement techniques usually calculate the amount of local visible distortions, similarities, or image statistics and perform pooling to generate the final metric value. Saliency maps resulted from VAMs can be integrated to the quality assessment pipeline in the pooling stage by emphasizing on the most salient regions in the content.

In order to integrate the saliency prediction into the full-reference stereoscopic quality assessment task, Zhang et al. proposed to use their 3D saliency map as a weighting factor for the Structural Similarity (SSIM) [13] in Depth Image Based Rendering applications [120]. In their method, 3D saliency is modeled as average of the 2D image saliency and 2D saliency map resulted from depth map. In another work by Chu et al., saliency is extracted using 2D VAM of [56] and is used as a weighting factor in the quality metric design [121]. Jiang et al. [122] used the 2D spectral residual VAM [123] along with a method of foreground and background depth maps for saliency prediction on stereoscopic images. They used a hard-threshold value for the depth map to split it into foreground and background depth and combined these two maps with the 2D saliency map of [123]. A major drawback of the mentioned approaches is that 2D VAMs are being used for saliency measurement of 3D content, while experiments have shown that 2D VAMs fail to accurately predict 3D human visual saliency [68],[124]. In addition to using 2D VAMs, the methods mentioned above do not take into consideration the temporal aspects of the video as they are solely based on single image quality assessment.

In the case of no-reference stereoscopic video quality assessment, Gu et al. proposed to add a saliency based term in the formulation of their sharpness metric [27]. This term is defined as a linear correlation between the 2D saliency maps of the two views. In another work by Ryu and Sohn [28], a no-reference quality assessment method for stereoscopic images is proposed by modeling the binocular quality perception in the context of blurriness and blockiness [28]. In this method, 2D VAM of GBVS [56] is used for saliency evaluations. The NR metrics proposed by Gu et al., Ryu and Sohn use 2D visual attention models to predict saliency for stereoscopic content. As mentioned, it is proven that 2D VAMs lack accuracy in saliency prediction for 3D data. The two NR metrics are designed for quality assessment of stereoscopic images and do not consider the temporal aspects of the quality evaluation.

## 1.5 Thesis Contributions

In this thesis, we present novel methods for quality assessment of stereoscopic 3D video content. In Chapter 2, we propose a full-reference 3D quality metric, which combines the quality of the cyclopean view and the quality of the depth map. In order to assess the degradation caused by 3D factors in the cyclopean view, a local image patch fusion method (based on HVS sensitivity to contrast) is incorporated to extract the local stereoscopic structural similarities. To this end, the information of the left and right channels is fused using the 3D-DCT transform. Then, we extract the local quality values using the structural similarity (SSIM) index to calculate the similarity between the reference cyclopean frame (fused left & right) and the distorted one. Moreover, the effect of depth on 3D quality of experience is taken into account through the disparity map quality component by considering the impact of the binocular vision on the perceived quality at every depth level. To this end, the variance of the disparity map and the similarity between disparity maps are incorporated to take to account the depth information in the proposed quality prediction model. HVS-based 2D metrics have been used in our design instead of MSE or MAD, as the former ones are reported to represent the perception of the human visual system more accurately [14]. A temporal pooling strategy is used to address recency and the worst section quality effects. The recency effect refers to the high influence of the last few seconds of the video on the viewer's ultimate decision on video quality. Worst section quality effect denotes the severe effect that the video segment with the worst quality has on the judgment of the viewers. The proposed metric can be tailored to different applications, as it takes into account the display size (the distance of the viewer from the display) and video resolution. The performance of our proposed method is verified through extensive subjective experiments using a large database of stereoscopic videos with various simulated 2D and 3D representative types of distortions. Moreover, the performance of our proposed scheme is compared with that of state-of-the-art 3D and 2D quality metrics in terms of efficiency as well as complexity. The main contributions of this chapter are summarized as follows: 1) Designed and formulated quality measures for the cyclopean view and depth map and proposed a 3D quality metric as a combination of these two quality measures that effectively predicts the quality of 3D content at the presence of different types of distortions, 2) Temporal quality effects, display size, and video resolution are taken into account in the design of the 3D quality metric, 3) Created a large database of stereoscopic videos containing several different representative types of distortions that may occur during the multiview video compression, transmission, and display process, and 4) Verified the performance of the proposed quality assessment method through large-scale subjective tests and provided a comprehensive comparison with the state-of-the-art quality metrics.

In Chapter 3, we investigate the effect of the frame rate on the 3D viewing experience with the objective to identify the appropriate frame rate for 3D video capturing. To this end, we analyze the relationship between frame rate, bitrate, and the 3D QoE through extensive subjective tests. We capture a database of stereoscopic 3D videos at various frame rates of 24, 30, 48, and 60 frames per second (fps) and use these videos in the experiments. In our study, the frame rates of 24 fps and 30 fps are chosen particularly because these frame rates have already been used in the 3D industry, and the frame rates of 48 fps and 60 fps were part of our experiment as there is a growing interest towards capturing 3D content with such frame rates. The findings of our study are helpful in defining bitrate adaptation guidelines for 3D video delivery over variable bitrate networks. These guidelines will allow network providers to change 3D content frame rate in order to deal with bandwidth capacity changes so that viewers' quality of experience is not significantly affected.

Chapter 4 investigates the computational modeling of visual attention of stereoscopic video and proposes an integrated saliency prediction method. Our approach utilizes both low-level attributes such as brightness, color, texture, orientation, motion, and depth as well as high-level context-dependent cues such as face, person, vehicle, animal, text, and horizon. Our model starts with a rough segmentation and quantifies several intuitive observations such as the effects of visual discomfort level, depth abruptness, motion acceleration, elements of surprise and size, compactness, and sparsity of the salient regions. To calculate local and global features describing these observations, a new fovea-based model of spatial distance between the image regions is used. Then, a random forest based algorithm is utilized to train a model of stereoscopic video saliency so that the various conspicuity maps generated by our method are efficiently fused into one single saliency map, which delivers high correlation with the eye-fixation data. The performance of the proposed saliency model is evaluated against the results of a large-scale eye-tracking experiment, which involves 24 subjects and an in-house database of 61 captured stereoscopic videos. The video saliency benchmark database used in this chapter is publicly available to the research community [125].

In Chapter 5, we propose to use 3D VAMs for quality evaluation of 3D video, for both NR and FR cases. To this end, we integrate our Learning Based Visual Saliency (LBVS-3D) prediction model for stereoscopic 3D video (along with several other VAMs) to various state-of-the-art FR and NR 3D video quality metrics. We evaluate the added value of incorporating 3D VAMs in FR and NR quality assessment of stereoscopic video using a large scale database of stereoscopic videos.

# 2 Full-Reference Human Visual System Based Quality Assessment of 3D Video

In this chapter, we propose a new full-reference quality metric for 3D content. Our method mimics HVS by fusing information of both the left and right views to construct the cyclopean view, as well as taking to account the sensitivity of HVS to contrast and the disparity of the views. In addition, a temporal pooling strategy is utilized to address the effect of temporal variations of the quality in the video. The following sections elaborate on the proposed FR quality assessment method.

## 2.1 Proposed 3D Quality Metric

As mentioned in the Introduction section, the binocular perception mechanism of the human visual system fuses the two view pictures into a single so called cyclopean view image. In addition, the perceived depth, affects the overall perceptual quality of picture. Our proposed Human-Visual-system-based 3D (HV3D) quality metric takes into account the quality of the cyclopean view and the quality of the depth information. Cyclopean view quality component evaluates the general quality of cyclopean image, while the depth map quality component measures the effect of disparity and binocular perception on the overall 3D quality. The cyclopean view is generated from the two view pictures (for the reference and distorted video) through the cyclopean view generation process. To mimic the binocular fusion of HVS, best matching blocks within the two views are found through a search process. These matching blocks are then combined in the frequency domain to generate the cyclopean view block. During the block fusion, Contrast Sensitivity Function (CSF) of HVS is taken into account through a CSF masking process. Then, the local similarities between the resulted cyclopean images are integrated into the overall cyclopean view quality component (Section 2.1.1). The depth map quality component is constructed by taking into account the quality of disparity maps as well as the impact of disparity variances (Section 2.1.2). As distortions can be perceived differently when there are different levels of depth present in a scene, disparity map variances are computed as part of the depth map quality component of our method.

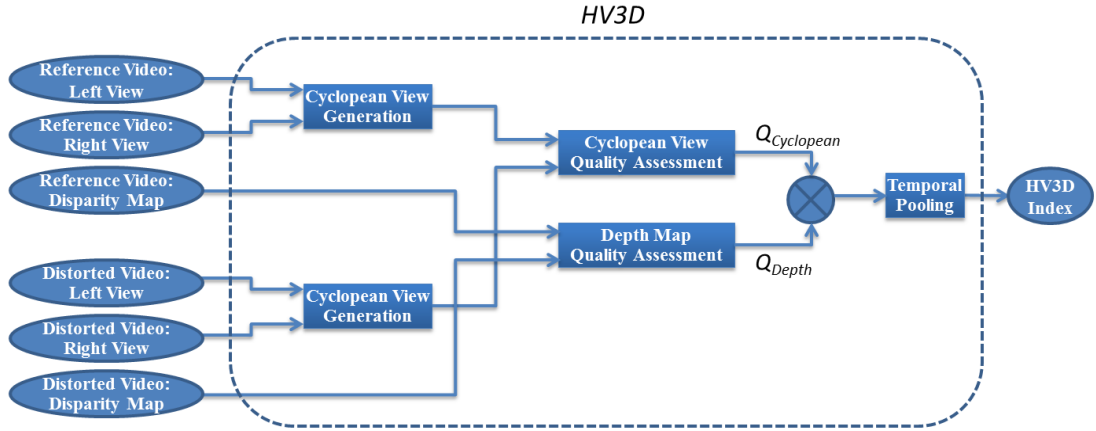

**Figure 2.1 Illustration of the proposed quality assessment technique**

The size of the blocks and geometry of the system structure in the depth map quality component are selected by considering the fovea visual focus in the HVS [126]-[130]. Fig. 2.1 illustrates the flowchart of the proposed framework.

### 2.1.1 Quality of the Cyclopean View – $Q_{Cyclopean}$

To imitate the binocular vision and form the cyclopean view image we combine the corresponding areas from the left and right views. To this end, the luma information of each view is divided into *m×m* blocks. For each block of the left view, the most similar block in the right view is found, utilizing the disparity information. In our method, the disparity map is assumed to be available for the stereo views.

The depth map is either captured using a depth capturing device (e.g., a Kinect sensor), or disparity is generated by stereo matching techniques. There are many algorithms available in the literature for disparity map generation. We use the MPEG Depth Estimation Reference Software (DERS) [131] for dense disparity map generation. Note that we extract left-to-right disparity. The use of DERS for disparity map generation is suggested only when a dense disparity map is not available as an input. Note that in our cyclopean view generation scheme we need to find the best match to each block of pixels. However a dense disparity map provides the disparity value for each pixel. To address this problem, we approximate the disparity value of each block by taking the median of the disparity values of the pixels within that block. The depth information of each block has inverse relationship with its disparity value [132]. Having disparity value per block, we can find the approximate coordinates of the corresponding block in the other view.

As Fig. 2.2 illustrates, $A_D$ is the approximate corresponding block for block $A_L$ in the right view, which has the same vertical coordinate as $A_L$ (illustrated as $j$ in Fig. 2.2), but its horizontal coordinate differs from $A_L$ by the amount of disparity (illustrated as $d$ in Fig. 2.2). Note that the $A_L$ and $A_D$ blocks are not necessarily the matching pair blocks that are fused by HVS since the position of $A_D$ is approximated based on the median of the disparity values of pixels within the block $A_L$. In the case of occlusions, the median value does not provide an accurate estimate of the block disparity, and can result to a mismatch. To find the most accurate matching block, we apply a matching block technique based on exhaustive search in a defined $M \times M$ search range around $A_D$ (see Fig. 2.2) using the Mean Square Error (MSE) cost function. In Fig. 2.2, $A_R$ is the best match for $A_L$ within the search range. Note that the block size and search range are chosen based on the display resolution. For instance, in the case of HD (High Definition) resolution video we choose $16 \times 16$ block size and the search area of $64 \times 64$, since our performance evaluations have shown that these are the best possible sizes that significantly reduce the overall complexity of our approach while allowing us to efficiently extract local structural similarities between views. The process of identifying matching blocks in the right view and the left view is done for both the reference and distorted stereo sets.

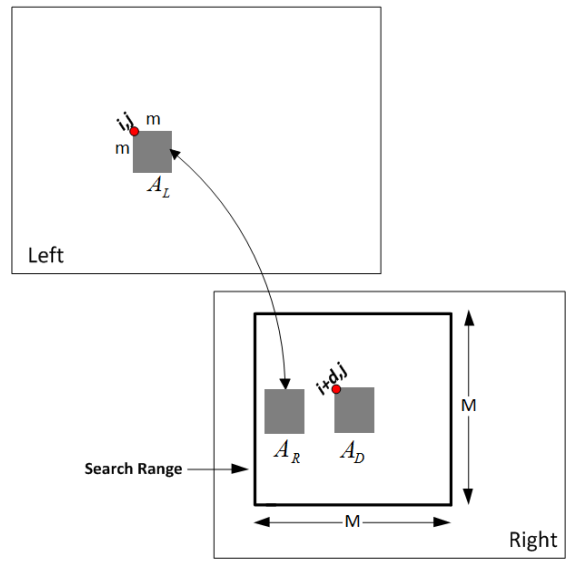

Figure 2.2 Selecting the best match: $A_D$ is the approximate corresponding block of $A_L$ in the right view + disparity; $A_R$ is the best match to $A_L$ within a search range

Note that the distortions do not have any influence on the search results, since search is only performed on the reference video frames, and then the search results (i.e., the coordination of the best matching block) are used to identify matching blocks in the reference pair as well as the distorted pair.

In order to generate the cyclopean view, once the matching blocks are detected, the information of matching blocks in the left and right views needs to be fused. Here we apply the 3D-DCT transform to each pair of matching blocks (left and right views) to generate two $m \times m$ DCT-blocks, which contain the DCT coefficients of the fused blocks. The top level $m \times m$ DCT-block includes the coefficients of lower frequencies compared to the bottom one. Since the human visual system is more sensitive to the low frequencies of the cyclopean view [9], we only keep the $m \times m$ DCT-block corresponding to the lower frequency coefficients and discard the other ones. As a next step, we consider the sensitivity of the human visual system to contrast, which also affects the perceived image quality [133]. To take into account this HVS property, we need to prioritize the frequencies that are more important to the HVS. To this end, similar to the idea presented in [134], we utilize the proposed JPEG quantization tables. The JPEG quantization tables have been obtained from a series of psychovisual experiments designed to determine the visibility thresholds for the DCT basis functions. Based on this, the DCT coefficients that represent the frequencies with higher sensitivity to the human visual system are quantized less than the other coefficients so that the more visually important content is preserved during the course of compression. In our application, instead of quantizing the DCT coefficients we decided to scale them so that bigger weights are assigned to more visually important content. To achieve this, we adopt the 8×8 JPEG quantization table and create an 8×8 Contrast Sensitivity Function (CSF) modeling mask, such that the ratio among its coefficients is inversely proportional to the ratio of the corresponding elements in the JPEG quantization table. By applying the CSF modeling mask to the 3D-DCT blocks, the frequencies that are of more importance to the human visual system are assigned bigger weights. This is illustrated as follows:

$$XC = C.\times X \qquad (2.1)$$

where $XC$ is our cyclopean-view model in the DCT domain for a pair of matching blocks in the right and left views, *X* represents the low-frequency 3D-DCT coefficients of the fused view, ".×" denotes the element-wise multiplication, and *C* is our CSF modeling mask. The elements of the CSF modeling mask are selected such that their average is equal to one. This guarantees that, in the case of uniform distortion distribution, the quality of each block within the distorted cyclopean view coincides with the average quality of the same view. Since the CSF modeling mask needs to be applied to $m \times m$ 3D-DCT blocks, for applications that require *m* to be greater than 8, cubic interpolation is used to up-sample the coefficients of the mask and create an $m \times m$ mask (in case *m* is less than 8, down-sampling will be used [128]).

Once we obtain the cyclopean-view model for all the blocks within the distorted and reference 3D views, the quality of the cyclopean view is calculated as follows:

$$Q_{Cyclopean} = \left( \sum_{i=1}^{N} \frac{SSIM(IDCT(XC_i), IDCT(XC'_i))}{N} \right)^{\beta_1} \qquad (2.2)$$

where $XC_i$ is the cyclopean-view model for the $i^{th}$ matching block pair in the reference 3D view, $XC'_i$ is the cyclopean-view model for the $i^{th}$ matching block pair in the distorted 3D view, IDCT stands for inverse 2D discrete cosine transform, *N* is the total number of blocks in each view, $\beta_1$ is a constant exponent, and SSIM is the structural similarity index [13],[14]. The value of $\beta_1$ is decided based on subjective tests presented in Section 2.2.

## 2.1.2 Quality of the Depth Map – $Q_{Depth}$

The quality of the disparity map is evaluated by: 1) fidelity measurement between the distorted and reference disparity maps and 2) taking into account variations of the depth at every different depth level. First, we elaborate on disparity map fidelity measurement and then we construct the overall depth map quality formulation.

### 2.1.2.1 Disparity Map Fidelity Measurement Using 2D Image Quality Metrics

We take into account the fidelity of the distorted disparity map in comparison to the reference disparity map. We investigate the relationship between the overall 3D video quality of experience and the depth map quality measured by perceptual-based image quality metrics, and choose the metric which produces the best correlation with the subjective tests. To this end, several particularly probable depth map artifacts are simulated and applied to the depth map sequence, and the synthesized views are generated using the original views and the distorted depth map sequence. The quality of the depth map is measured using perpetual-based image quality metrics and its relationship with the subjective quality results of the synthesized stereo videos is studied.

To study the effect that depth map quality has on the perceived 3D video quality, first we apply common probable depth map artifacts to the depth map sequence corresponding to left view of a stereo video pair. Then the right view of the stereo video pair is synthesized using the left view and its corresponding modified depth map sequence. Note that the left view utilized to synthesize the right view has original quality (it has not been distorted by any kind of processing or distortions). We measure the correlation between the quality of depth map sequences and the overall subjective 3D Quality of Experience. According to our experiment results, which will be explained in detail in Section 2.2, among the 2D image fidelity metrics, the Visual Information Fidelity (VIF) index [135] between the disparity maps achieved the highest correlation with the subjective stereo video quality. Therefore, we choose VIF to measure the fidelity of the disparity maps.

*2.1.2.2 Constructing the Disparity Map Quality Component*

Depth information plays an important role in the perceptual quality of 3D content. The quality of the depth map becomes more important if there are several different depth levels in the scene. On the contrary, in a scene with a limited number of depth levels, the quality of the depth map plays a less important role in the overall 3D quality. This suggests considering the variance of depth map in conjunction with the depth map quality to reflect the importance of the depth map quality. However the variance of depth is required to be taken into account locally in the scene, as only a portion of the scene can be fully projected onto the eye fovea when watching a 3D display from a typical viewing distance. Fig. 2.3 illustrates the relationship between the block size projected onto eye fovea and the distance of the viewer. As it can be observed, the length of a square block on the screen that can be fully projected onto the eye fovea is calculated as follows:

$$K = 2 \times d \times \tan(\alpha) \qquad (2.3)$$

where $K$ is the length of the block (in [*mm*]), $d$ is the proper viewing distance from the display (in [*mm*]), and $\alpha$ is the half of the angle of the viewer's eye at the highest visual acuity. The proper distance of a viewer from the display is decided based on the size of the display. The range of *2α* is between *0.5°* and *2°* [136]. The sharpness of vision drops off quickly beyond this range. The length of the block (*K*) can be translated in pixel units as follows:

$$k = \frac{h \times K}{H} = \frac{2 \times d \times h \times \tan(\alpha)}{H} \tag{2.4}$$

where *k* is the length of the block on the screen (in pixels), *H* is the height of the display (in [*mm*]), and *h* is the vertical resolution of the display.

The local disparity variance is calculated over a block size area that can be fully projected onto the eye fovea when watching a 3D display from a typical viewing distance. For calculating the local depth-map variance of the ith block (i.e. $\sigma_{d_i}^2$), an outer block of *k×k* pixels is considered such that the *m×m* block is located at its centre (see Fig. 2.4 for the present embodiment), and is defined as follows:

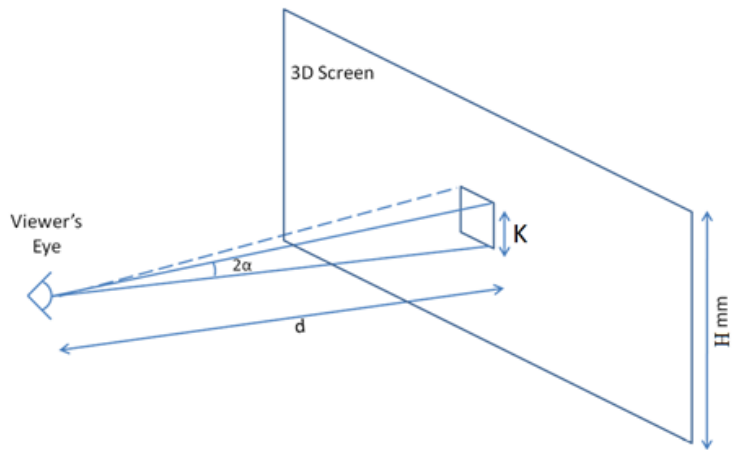

**Figure 2.3 Visual acuity of fovea, receptive field; relationship between the block size projected onto eye fovea and the distance of the viewer**

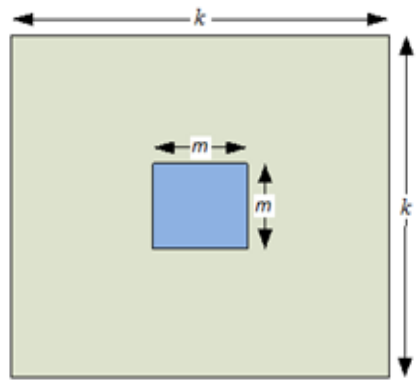

**Figure 2.4 Block structures for calculating the variance of disparities**

$$\sigma_{d_i}^2 = \frac{1}{k \times k - 1} \sum_{j,l=1}^{k} (M_d - R_{j,l})^2 \tag{2.5}$$

where $M_d$ is the mean of the depth values of each $k \times k$ block (outer block around the $i^{th}$ $m \times m$ block) in the normalized reference depth map. The reference depth map has been normalized with respect to its maximum value in each frame, so that the depth values range from 0 to 1. $R_{j,l}$ is the depth value of pixel $(j,l)$ in the outer $k \times k$ block within the normalized reference depth map.

As mentioned in the previous subsection, we verified that among the existing state-of-the-art 2D quality metrics, the Visual Information Fidelity (VIF) index of depth maps has the highest correlation with the mean opinion scores (MOS) of viewers [130]. It is also worth mentioning that the SSIM term in (2.2) compares the structure of two images only and does not take into account the effect of small geometric distortions [14]. The geometrical distortions are the source of vertical parallax, which causes severe discomfort for viewers (brain cannot properly fuse right and left view). The geometric distortions between right and left images (within the cyclopean view) are reflected in the depth map. Therefore, VIF is used to compare the quality of depth map of the distorted 3D content with respect to that of the reference one (see [14],[135] for more details on VIF) as follows:

$$Q_{Depth} = (VIF(D,D'))^{\beta_2} \left( \sum_{i=1}^{N} \frac{\sigma_{d_i}^2}{N . \max(\sigma_{d_j}^2 \mid j=1,2,...,N)} \right)^{\beta_3} \tag{2.6}$$

where $D$ is the depth map of the reference 3D view, $D'$ is the depth map of the distorted 3D view, VIF is the Visual Information Fidelity index, $\beta_2$ and $\beta_3$ are constant exponents, $N$ is the total number of blocks, and $\sigma_{d_i}^2$ is the local variance of block $i$ in the depth map of the 3D reference view. Note that the latter part in equation (2.6) performs a summation over the normalized local variances. The variance term is computed for each block according to (2.5) and normalized to the maximum possible variance value for all the blocks. The summation is then divided by $N$, the total number of blocks, to provide an average normalized local variance value in the interval of [0,1]. Normalization of the variance is performed to ensure the corresponding metric values range between 0 and 1.

### 2.1.3 Constructing the Overall HV3D Metric

Once the quality of the distorted cyclopean view and depth map is evaluated (see (2.2) and (2.6)), the final form of our HV3D quality metric is defined as follows:

$$HV3D = \left( \sum_{i=1}^{N} \frac{SSIM(IDCT(XC_i), IDCT(XC'_i))}{N} \right)^{\beta_1}$$
$$\times (VIF(D, D'))^{\beta_2} \cdot \left( \sum_{i=1}^{N} \frac{\sigma^2_{d_i}}{N \cdot \max(\sigma^2_{d_j} \mid j = 1, 2, ..., N)} \right)^{\beta_3} \quad (2.7)$$

where $XC_i$ is the cyclopean-view model for the $i^{th}$ matching block pair in the reference 3D view, $XC'_i$ is the cyclopean-view model for the $i^{th}$ matching block pair in the distorted 3D view, IDCT stands for inverse 2D discrete cosine transform, $N$ is the total number of blocks in each view, $\beta_1$, $\beta_2$, $\beta_3$ are the constant exponents, SSIM is the structural similarity index [13],[14], $D$ is the depth map of the reference 3D view, $D'$ is the depth map of the distorted 3D view, VIF is the Visual Information Fidelity index, and $\sigma^2_{d_i}$ is the local variance of block $i$ in the depth map of the 3D reference view. The exponent parameters in HV3D ($\beta_1$, $\beta_2$, and $\beta_3$) are determined in the subsection 2.1.4. It is worth noting that the choice of multiplication for the quality components can be replaced with other combining methods. Our experiments verified that the present choice outperforms linear and polynomial regression.

Since different frames of a video have different influence on the human judgment of quality, the overall quality of a video sequence is found by assigning weights to frame quality scores, according to their influence on the overall quality. Subjective tests for video quality assessment have shown that subjects' ultimate decisions on the video quality are highly influenced by the last few seconds of the video (recency effect) [22],[137][138]. Moreover, the video segment with the worst quality highly affects the judgment of the viewers [138]. This is true mainly because subjects keep the most distorted segment of a video in memory much more than segments with good or fair quality [138]. Temporal pooling algorithms have been proposed that map a series of fidelity scores associated to different frames to a single quality score that represents an entire video sequence [22]. To address the recency and worst section quality effects, a temporal pooling strategy has been used in our study, which is discussed in subsection 2.1.5. The overall approach proposed in our study to evaluate the quality of 3D content is illustrated as a flowchart in Fig. 2.1. A MATLAB implementation of our metric is available online at our web site [139].

### 2.1.4 Constant Exponents: $β_1$, $β_2$, and $β_3$

To find the constant exponents for our HV3D quality metric and validate its performance, we performed subjective tests using two different 3D databases (one set for training and one set for validation).

To estimate the exponent constants of the proposed metric as denoted in the equation (2.7), all the terms were calculated for each video in the training dataset. To determine the best values for the exponent constants $β_1$ $β_2$, and $β_3$, we need to maximize the correlation between our HV3D indices and the MOS values of the training dataset. This can be formulated as follows:

$$\max_{β_i, i=1,2,3} \{ρ(HV3D, MOS)\} \qquad (2.8)$$

where $ρ$ is the Pearson correlation coefficient. We evaluate the correlation between HV3D and MOS vectors over a wide range of $β_i$ values and select the $β_i$ values that result in the highest correlation. The accuracy and robustness of the obtained exponents is further confirmed by measuring the correlation between MOS values and the HV3D indices of the validation video set (see Section 2.4).

### 2.1.5 Temporal Pooling Strategy

In our study, to address the recency and worst section quality effects we used the exponentially weighted Minkowski summation temporal pooling mechanism [140]. As shown in [22], the exponentially weighted Minkowski summation strategy outperforms some other existing temporal pooling methods, such as the histogram-based pooling, averaging, mean value of the last frames, and local minimum value of the scores in successive frames. The exponentially weighted Minkowski summation is formulated as:

$$HV3D_{\exp Minkowski} = [\frac{1}{N_f} \sum_{i=1}^{N_f} HV3D_i^p . e^{\frac{i-N_f}{τ}}]^{1/p} \qquad (2.9)$$

where $N_f$ is the total number of frames, $p$ is the Minkowski exponent, and $τ$ is the exponential time constant that controls the strength of the recency effect. Higher values of *p* result in the overall score to be more influenced by the frame with largest degradation. In order to find the best values of *p* and *τ*, the Pearson correlation coefficient (PCC) is calculated for a wide range of these two parameters for the training video set.

Moreover, in order to adjust the proposed metric for the asymmetric video content where the overall quality of right and left view and their corresponding depth maps is not identical, the reference depth map and the base view for finding matching blocks (in the process of cyclopean view and depth map quality evaluation) are switched between the two views for every other frame. As a result, the overall 3D quality is not biased by the quality of one view or the eye dominance effect [141].

## 2.2 Experiment Setup

This section provides details on the experiment setup, video sets used in our experiments, and the parameters of the proposed metric.

### 2.2.1 Dataset

To adjust the parameters of our proposed scheme and verify its performance, we used two different video data sets (one training dataset and one validation dataset). The specifications of the training and validation video sets are summarized in Table 1 and Table 2, respectively. These sequences are selected from the test videos in [11], the 3D video database of the Digital Multimedia Lab (DML) at the University of British Columbia (publicly available [139]), and sequences provided by MPEG for standardization activities and subjective studies [142]. These datasets contain videos with fast motion, slow motion, dark and bright scenes, human and non-human subjects, and a wide range of depth effects. Note that the test sequences adopted from [11] and [139] are naturally captured stereoscopic videos, i.e., they contain only two views captured using two side-by-side cameras. The MPEG sequences, however, are originally multiview sequences with several views per each video. For each of the multiview sequences, only the two views (i.e., one stereoscopic pair) which were recommended by MPEG in their Common Test Conditions are used [142]. For each video sequence, the amount of spatial and temporal perceptual information is measured according to the ITU Recommendation P.910 [143] and results are reported in Table 2.1 and Table 2.2. For the spatial perceptual information (SI), first the edges of each video frame (luminance plane) are detected using the Sobel filter [144]. Then, the standard deviation over pixels in each Sobel-filtered frame is computed and the maximum value over all the frames is chosen to represent the spatial information content of the scene. The temporal perceptual information (TI) is based upon the motion difference between consecutive frames. To measure the TI, first the difference between the pixel values (of the luminance plane) at the same coordinates in consecutive frames is calculated. Then, the standard deviation over pixels in each frame is computed and the maximum value over all the frames is set as the measure of TI. More motion in adjacent frames will result in higher values of TI. Fig. 2.5 shows the spatial and temporal information indexes of each test sequence.

Table 2.1 Training dataset

| Sequence | Resolution | Frame Rate (fps) | Number of Frames | Spatial Complexity (Spatial Information) | Temporal Complexity (Temporal Information) | Depth Range (*cm*) |
|---|---|---|---|---|---|---|
| Poznan_Hall2 | 1920×1080 | 25 | 200 | Low (35.4658) | Low (11.1460) | High (28.93) |
| Undo_Dancer | 1920×1080 | 25 | 250 | High (81.0423) | High (26.9021) | High (30.69) |
| Kendo | 1920×1080 | 30 | 300 | Medium (47.2172) | High (26.8791) | High (21.39) |
| Balloons | 1920×1080 | 30 | 500 | Medium (48.6726) | High (21.4660) | Low (5.84) |
| Cokeground | 1920×1080 | 30 | 210 | High (86.9096) | Medium (15.9128) | Low (4.99) |
| Ball | 1920×1080 | 30 | 150 | Medium (49.7701) | Low (13.3074) | Medium (15.53) |
| Alt-Moabit | 1920×1080 | 30 | 100 | High (111.0437) | High (21.2721) | Medium (13.36) |
| Hands | 1920×1080 | 30 | 251 | High (114.6755) | High (25.2551) | Medium (15.86) |

Table 2.2 Validation dataset

| Sequence | Resolution | Frame Rate (fps) | Number of Frames | Spatial Complexity (Spatial Information) | Temporal Complexity (Temporal Complexity) | Depth Range (*cm*) |
|---|---|---|---|---|---|---|
| Poznan_Street | 1920×1080 | 25 | 250 | High (95.3103) | High (26.5562) | High (34.01) |
| GT_Fly | 1920×1080 | 25 | 250 | Medium (58.8022) | High (33.0102) | High (31.02) |
| Lovebird1 | 1920×1080 | 30 | 300 | Medium (59.2345) | Low (13.8018) | Medium (15.01) |
| Newspaper | 1920×1080 | 30 | 300 | High (65.1173) | Medium (17.1297) | Low (5.09) |
| Soccer2 | 1920×1080 | 30 | 450 | High (115.2781) | High (28.6643) | Medium (16.99) |
| Flower | 1920×1080 | 30 | 112 | Medium (43.0002) | Low (13.5305) | Low (5.86) |
| Horse | 1920×1080 | 30 | 140 | High (85.4988) | High (22.3184) | Medium (13.56) |
| Car | 1920×1080 | 30 | 235 | Medium (49.6162) | Medium (16.0197) | High (24.21) |

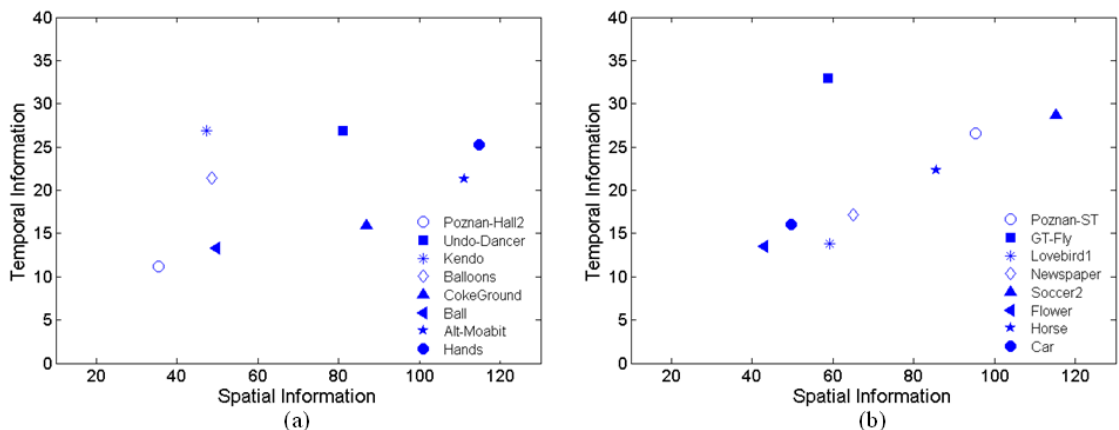

**Figure 2.5 Spatial and temporal information for the training (a) and test (b) datasets**

In addition to spatial and temporal information, for each sequence shown in Table 2.1 and Table 2.2, we also provide information about the scene's depth bracket. The depth bracket of each scene is defined as the amount of 3D space used in a shot or a sequence (i.e., a rough estimate of the difference between the distance of the closest and the farthest visually important objects from the camera in each scene) [145]. Since the information regarding the objects/camera coordinates is not available for all of the sequences, we adopt the disparity to depth conversion method introduced in [69] (which takes into account the display size and distance of the screen from the viewer) to find the depth of each object with respect to the viewer. We report the approximate averaged-over-frames of depth difference between the closest and farthest visually important objects. The visually important objects are chosen based on our 3D visual attention model [146] which takes into account various saliency attributes such as brightness intensity contrast, color, depth, motion, and texture. It is observed from Table 2.1 and Table 2.2 that the training and test videos have a similar distribution of properties (spatial and temporal complexity and depth bracket). This makes it possible to compute the required parameters in our proposed quality metric using the training video set and use the same ones for performance evaluations over the test video set.

In order to evaluate the performance of our proposed quality metric, in addition to the compression distortion which occurs during 3D video content delivery, general distortions used by 2D quality metric studies [13],[14],[135] as well as 3D metrics are also considered for comparative purposes with other methods. The suggested scheme for delivering 3D content is to transmit two or three simultaneous views of the scene and their corresponding depth maps and synthesize extra views at the receiver end to support multiview screens [147]. In this process the delivered content might be distorted due to compression of views, compression of depth maps, or view synthesizing (See Fig. 2.6 for an illustration).

Once these distortions are applied to the content, the quality of videos is evaluated both subjectively and objectively using the HV3D metric and existing state-of-the-art 2D and 3D metrics. Note that the levels of distortions applied to the 3D content are such that they lead to visible artifacts which in turn allow us to correlate subjective tests - Mean Opinion Score (MOS) - with objective results.

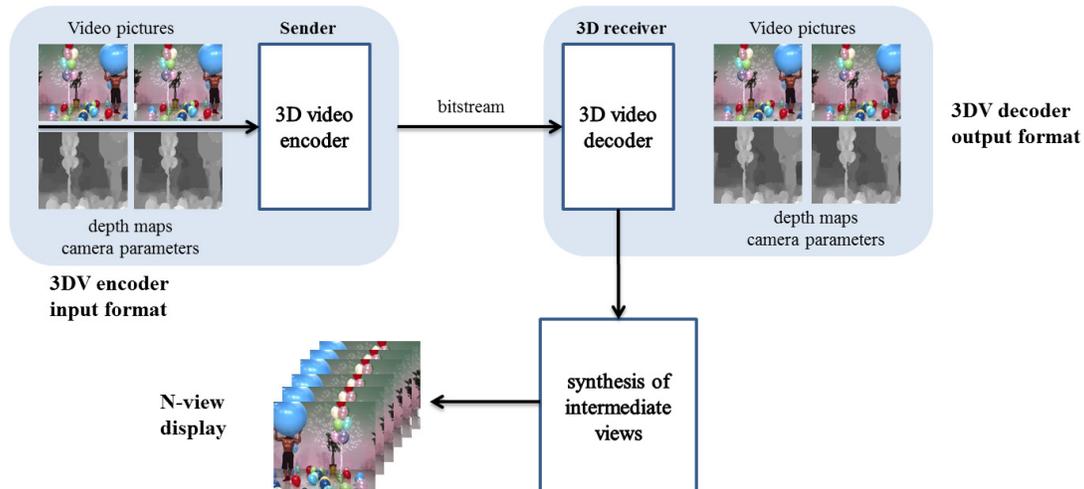

Figure 2.6 Multiview content delivery pipeline

The following types of distortions are applied to both training and validation 3D videos (seven different types of distortions applied to 16 original videos, which results in 208 distorted videos):

1) High compression of views (simulcast coding): the right and left views are simulcast coded using an HEVC-based encoder (reference software HM ver. 9.2) [148]-[150]. The low delay configuration setting with the GOP (group of pictures) size of 4 was used. The quantization parameter was set to 35 and 40 to investigate the performance of our proposed metric at two different compression-distortion levels with visible artifacts. A new depth map for the distorted stereo pair is generated using DERS [131].

2) High compression of depth map: the depth map of one view (left view in our simulations) is compressed using an HEVC-based encoder (HM reference software ver. 9.2 [148]-[150]) with Quantization Parameters (QPs) of 25 and 45, and low delay profile with GOP size of 4. Then, the right view is synthesized using the decoded depth map and the original left view. View synthesis is performed using the view synthesis reference software (VSRS 3.5) [151]. Also, a new depth map for the synthesized view is generated using DERS [131]. The quality of the stereo pair including the synthesized right view and the original left view is compared with that of the reference one.

3) High compression of 3D content (views and depth maps): the right and left view video sequences and their corresponding depth maps are encoded using an HEVC-based 3D video encoder (3D HTM reference software ver. 9 [150]) with random access high efficiency configuration. GOP size was set to 8 and QP values were set to 25, 30, 35, and 40 according to the MPEG common test conditions for 3D-HEVC [142].

4) View synthesis: using one of the views of each stereo pair and its corresponding depth map, the other view is synthesized via VSRS and a new stereo pair is generated [151]. As a result, two sets of distorted stereo pairs are formed where the synthesized view once was the left view and once the right one. Then, new depth maps are generated for these distorted stereo pairs using DERS [131].

5) White Gaussian noise: white Gaussian noise with zero mean and variance value 0.01 is applied to both right and left views. DERS is used to generate a new depth map for the distorted stereo pair [131].

6) Gaussian low pass filter: a Gaussian low pass filter with the size 4 and the standard deviation of 4 is applied to both right and left views. Then DERS generated a new depth map for the distorted stereo pair [131].

7) Shifted (increased) intensity: the brightness intensity of right and left video streams is increased by 20 (out of 255). Once more, a new depth map is generated for the distorted stereo pair using DERS [131].

In our study, capturing artifacts (such as window violation, vertical parallax, depth plane curvature, keystone distortion, or shear distortion) are not considered, as our proposed quality metric is a full-reference metric and requires a reference for comparison. In other words, it is assumed that the 3D video content and the corresponding depth map sequences are already properly captured and the goal is to evaluate the perceived quality when other processing/distortions are applied. Distortions introduced during 3D content acquisition using rangefinder sensors are not considered in our experiments. In addition, we did not take into account the effect of crosstalk, as the amount of crosstalk depends on the 3D display technology used. We assume that the display imposes the same amount of crosstalk to both reference and distorted 3D content. To design a quality metric that takes into account the effect of crosstalk, the information about the amount of crosstalk at different intensity level for different 3D display systems is required. After applying the seven above-mentioned distortions (at different levels) to the sixteen stereo videos, we obtain 208 distorted stereo videos.

*2.2.2 Subjective Test Setup*

The viewing conditions for subjective tests were set according to the ITU-R Recommendation BT.500-13 [152]. The evaluation was performed using a 46" Full HD Hyundai 3D TV (Model: S465D) with passive glasses. The peak luminance of the screen was set at 120 [$cd/m^2$] and the color temperature was set at 6500$K$ according to MPEG recommendations for the subjective evaluation of the proposals submitted in response to the 3D Video Coding Call for Proposals [153]. The wall behind the monitor was illuminated with a uniform light source (not directly hitting the viewers) with the light level less than 5 % of the monitor peak luminance.

A total of 88 subjects participated in the subjective test sessions, ranging from 21 to 32 years old. All subjects had none to marginal 3D image and video viewing experience. They were all screened for color blindness (using Ishihara chart), visual acuity (using Snellen charts), and stereovision acuity (via Randot test – graded circle test 100 seconds of arc). Subjective evaluations were performed on both training and validation data sets (see Table 2.1 & Table 2.2).

Test session started after a short training session, where subjects became familiar with video distortions, the ranking scheme, and test procedure. Test sessions were set up using the single stimulus (SS) method where videos with different qualities were shown to the subjects in random order (and in a different random sequence for each observer). Each test video was 10 seconds long and a four-second gray interval was provided between test videos to allow the viewers to rate the perceptual quality of the content and relax their eyes before watching the next video. There were 11 discrete quality levels (0-10) for ranking the videos, where score 10 indicated the highest quality and 0 indicated the lowest quality. Here, the perceptual quality reflects whether the displayed scene looks pleasant in general. In particular, subjects were asked to rate a combination of "naturalness", "depth impression" and "comfort" as suggested by [154]. After collecting the experimental results, we removed the outliers from the experiments (there were seven outliers) and then the mean opinion scores from the remaining viewers were calculated. Outlier detection was performed in accordance to ITU-R BT.500-13, Annex 2 [152].

*2.2.3 Assigning Constant Exponents and Pooling Parameters*

To find the constant exponents in equation (2.7), the two quality components were calculated for the videos in the training set. Note that in our experiment, a block size of 64×64 was chosen for measuring the variance of disparity. In this case, the 3D display has resolution of 1080×1920 (HD), its height is 573 *mm*, the appropriate viewing distance is 1830 *mm*, and the value of *2α* in equation (2.4) is roughly equal to *0.88°* which is consistent with the fovea visual focus. Then, the exponent values that result in the highest correlation between the HV3D metric and the MOS values corresponding to the training distorted video sets are selected (see equation (2.8)). The selected constant exponents are $\beta_1$=0.4, $\beta_2$=0.1, and $\beta_3$=0.29. In order to find the Minkowski exponent parameter (*p*) and the exponential time constant (*τ*) for temporal pooling (see equation (2.9)), the Pearson Correlation Coefficient (which measures the accuracy of a mapping) is calculated for a wide range of these two parameters over the training video set. In other words, these parameters are exhaustively swept over a wide range to enable us find the highest stable maximum point in the accuracy function. Extensive numerical evaluations show that the effect of slight changes in the selected pooling parameters is negligible in the overall metric performance. The same performance evaluations have shown that *p* = 9 and *τ* = 100 result in the highest correlation between the subjective tests and our metric results.

## 2.3 Results and Discussions

In this section, first we present our performance evaluations for disparity map fidelity measurement which was explained in Section 2.1.2.1. Then, we evaluate the performance of each component of the HV3D quality metric (cyclopean view and depth map quality terms) as well as its overall performance over the validation data set. The performance of HV3D is also compared with that of state-of-the-art 2D and 3D quality metrics. The performance of HV3D quality metric is discussed in the following subsections.

*2.3.1 Metrics of Performance*

In order to evaluate the correlation between our proposed HV3D quality metric and subjective MOS results collected from the validation dataset, we use the Spearman rank order correlation coefficient (SCC), the Pearson correlation coefficient (PCC), and the Root-Mean-Square Error (RMSE). While PCC and RMSE measure the accuracy of the 3D QoE prediction by quality components, SCC measures the statistical dependency between the subjective and objective results. In other words Spearman ratio assesses how well the relationship between two variables can be described using a monotonic function, i.e., it measures the monotonicity of the mapping from each quality metric to MOS. Moreover, to measure the consistency of mapping from each quality metric component to MOS, outlier ratios (*OR*) are calculated [155].

In addition to the evaluation measures mentioned above, we also use logistic fitting curves to demonstrate graphically the mapping between the proposed quality metric and the subjective scores. The logistic fitting curve is formulated as follows [156]:

$$y = \frac{a}{1 + e^{-b(x-c)}} \qquad (2.10)$$

where *x* denotes the horizontal axis (quality metric), *y* represents the vertical axis (MOS), and *a*, *b*, and *c* are the fitting parameters.

### 2.3.2 *Disparity Map Fidelity Measurement*

We measure the correlation between the objective quality of depth map sequences (using 2D image quality metrics) and the overall subjective 3D Quality of Experience. We use our 3D video dataset and perform the subjective experiments as explained earlier in Section 2.2. Note that since some of the distortions are not applied to the depth maps in practice, we only consider the depth map compression and packet loss artifacts.

To evaluate the quality of distorted depth maps in our experiment, we used the following quality metrics: PSNR, SSIM [13], MS-SSIM (Multi Scale SSIM) [157], DCT-based video quality metric (VQM) [158], and VIF [135]. Fig. 2.7 illustrates the relation between the perceptual quality of stereo views and the quality of the depth map using different quality metrics. Table 2.3 shows the statistical dependencies between the overall 3D subjective quality and various 2D image metrics applied to the depth maps. It is observed from this table that the overall perceived 3D video quality highly depends on the depth map quality. Also, by comparing the performance of the quality metrics we observe that the Visual Information Fidelity has the highest correlation with MOS. VIF and also other metrics in Table 2.3 are originally proposed as quality metrics for natural images. Our study shows that they have also a fair performance for depth map quality evaluation. This might be because depth maps roughly include some portions of the scene structure. The reported correlation values in Table 2.3 also confirm that depth map alone is not sufficient for predicting the overall 3D video quality.

### 2.3.3 *Contribution of Quality Components of HV3D Metric*

In order to investigate the contribution of the cyclopean view and depth map quality components of HV3D quality metric (see equation 2.7) in predicting the overall 3D QoE, the correlation between each quality component and the MOS values over the validating dataset is studied by calculating the SCC, PCC, RMSE, and OR. Table 2.4 shows SCC, PCC, RMSE, and OR for cyclopean view and depth map quality components of HV3D. As it is observed, the quality components of the proposed metric demonstrate high correlation with the Mean Opinion Scores. In the following subsection the overall performance of HV3D as a combination of the two quality components is analyzed.

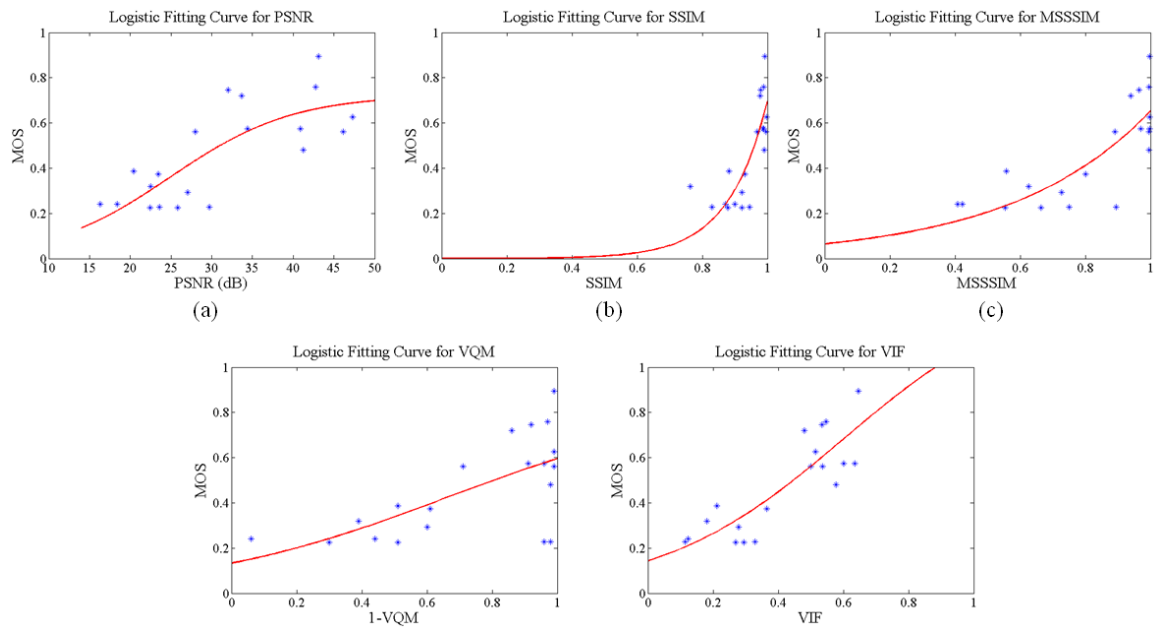

**Figure 2.7 Comparing the subjective results with objective results using PSNR (a), SSIM (b), MS-SSIM (c), VQM (d), and VIF (e) objective quality metrics**

**Table 2.3 The prediction performance of the conventional 2D quality metrics using only depth maps**

| Quality Metric | Spearman Ratio | Pearson Ratio | RMSE | Outlier Ratio |
|:---:|:---:|:---:|:---:|:---:|
| PSNR | 0.6209 | 0.6535 | 0.1324 | 0 |
| SSIM | 0.6420 | 0.6126 | 0.1247 | 0 |
| MS-SSIM | 0.6541 | 0.6623 | 0.1242 | 0 |
| VQM | 0.5748 | 0.5361 | 0.1602 | 0 |
| **VIF** | **0.6706** | **0.7451** | **0.1086** | **0** |

### 2.3.4 Overall Performance of the HV3D Metric

To prove the efficiency of the HV3D quality metric, its performance is compared with that of the state-of-the-art 3D quality metrics using the validation dataset. Note that the parameters of PHVS3D [19] and PHSD [9] are originally customized for 3D mobile applications. For a fair comparison these parameters are updated for 3D HD screen. In addition, we follow what is considered common practice in evaluating 3D quality metrics and compare the performance of the HV3D against several 2D quality metrics including PSNR, SSIM [13], VIF [135], VQM [158], and MOVIE [159]. To this end, the quality of the frames of both views is measured separately using these 2D quality metrics and then the average quality over the frames from both views is calculated. Table 2.5 provides an overview on the 3D quality metrics used in our experiments. Fig. 2.8 shows the relationship between the MOS and the resulting values from each quality metric for the entire validation set and all 7 different distortions (as described in Section 2.2). A logistic fitting curve is used for each case to clearly illustrate the correlation between subjective results (MOS) and the results derived by each metric. Fig. 2.8 shows that our HV3D objective metric demonstrates strong correlation with the MOS results.

In order to evaluate the statistical relationship between each of the quality metrics and the subjective results, PCC, SCC, RMSE, and OR are calculated for different quality metrics over the entire validation dataset and is illustrated in Table 2.6. According to this table, it appears that the performance of our HV3D in quantifying the quality of the entire validation dataset in the presence of the 7 representative types of distortions is superior to other objective metrics in terms of accuracy, monotonicity, and consistency. In particular, Pearson correlation coefficient between our metric and MOS is 90.8%, Spearman correlation ratio is 91.3%, and RMSE is 6.43. As it is observed the hybrid combination of the cyclopean view and depth map quality components has improved the correlation between the quality indices and MOS values (see Table 2.4 and Table 2.6).

**Table 2.4 Statistical performance of the cyclopean view and depth map quality components of HV3D**

| Quality Metric | Spearman Ratio | Pearson Ratio | RMSE | Outlier Ratio |
|---|---|---|---|---|
| $Q_{Cyclopean}$ | 0.8177 | 0.8660 | 7.133 | 0 |
| $Q_{Depth}$ | 0.7993 | 0.8524 | 7.398 | 0 |

**Table 2.5 Overview of different full-reference stereoscopic quality metrics**

| Quality Metric | Direct Use of the Views | Use of Cyclopean View | Use of Depth (Disparity) Map | Combine 2D and 3D Quality | Temporal Pooling | Complexity | Use of Color | Stereoscopic Dataset | Size of the Dataset |
|---|---|---|---|---|---|---|---|---|---|
| Ddl1 [16] | Yes | No | Yes | Yes | No | Low | No | Images | Medium |
| OQ [17] | Yes | No | Yes | Yes | No | Low | No | Images | Medium |
| CIQ [18] | No | Yes | No | No | No | Low | No | Images | Medium |
| PHVS3D [19] | No | Yes | No | No | No | High | No | Videos | Small |
| PHSD [9] | No | Yes | Yes | Yes | No | High | No | Videos | Medium |
| MJ3D [21] | No | Yes | No | No | No | Medium | No | Images | Medium |
| Q_Shao [20] | Yes | No | No | Yes | No | High | No | Images | Medium |
| Proposed | No | Yes | Yes | Yes | Yes | Medium | No | Videos | Large |

**Table 2.6 Statistical performance of different quality metrics over the whole validation dataset**

| Quality Metric | Spearman Ratio | Pearson Ratio | RMSE | Outlier Ratio |
|---|---|---|---|---|
| PSNR | 0.6350 | 0.6454 | 10.388 | 0.0167 |
| SSIM [13] | 0.6213 | 0.6844 | 9.852 | 0.0083 |
| VQM [158] | 0.5981 | 0.6660 | 10.095 | 0.0083 |
| VIF [135] | 0.7204 | 0.7257 | 9.166 | 0 |
| MOVIE [159] | 0.7967 | 0.7527 | 8.623 | 0 |
| Ddl1 [16] | 0.7321 | 0.7370 | 8.732 | 0 |
| OQ [17] | 0.7900 | 0.7580 | 8.610 | 0 |
| CIQ [18] | 0.7080 | 0.7200 | 9.446 | 0.0083 |
| PHVS3D [19] | 0.8233 | 0.7837 | 8.420 | 0 |
| PHSD [9] | 0.7841 | 0.7911 | 8.321 | 0 |
| MJ3D [21] | 0.8947 | 0.8640 | 7.229 | 0 |
| Q_Shao [20] | 0.7988 | 0.8348 | 7.902 | 0 |
| **HV3D** | **0.9130** | **0.9082** | **6.433** | **0** |

## 2.3.5 *Performance of HV3D in the Presence of Different Types of Distortions*

To evaluate the performance of the proposed metric in predicting the quality of the content in the presence of different types of distortions, the statistical relationship between HV3D indices and MOS values is analyzed separately per each type of distortion.

Table 2.7 shows the PCC and SCC values for various quality metrics and different types of distortions over the validation dataset. It appears that the HV3D quality metric either outperforms other quality metrics or in the worst case its performance is quite comparable in predicting the quality of distorted 3D content. The results in Table 2.7 show superior performance of our proposed quality metric specifically for 3D video coding and view synthesizing applications.

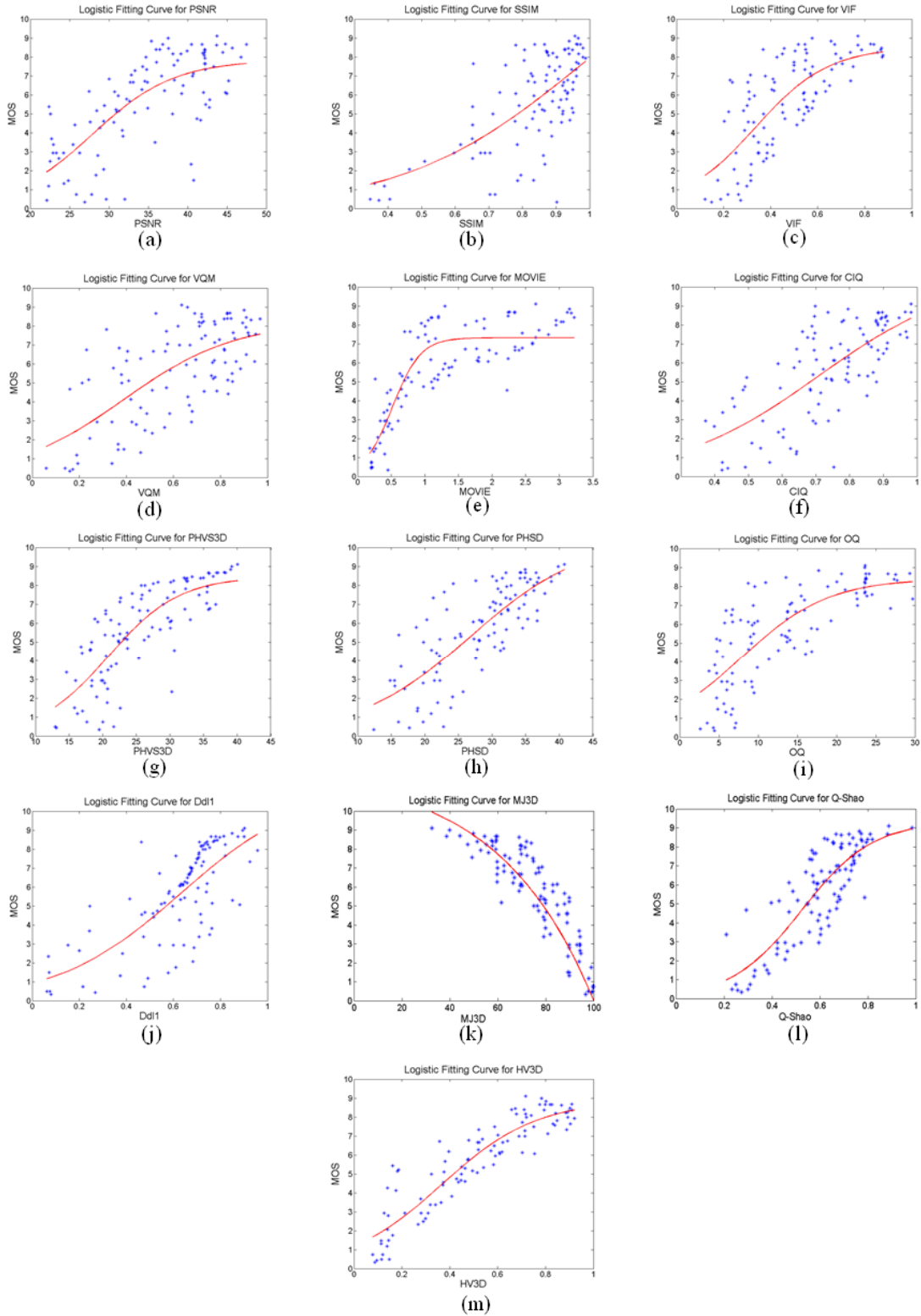

**Figure 2.8 Comparing the subjective results with objective results using PSNR (a), SSIM [13] (b), VIF [135] (c), VQM [158] (d), MOVIE [159] (e), CIQ [18] (f), PHVS3D [19] (g), PHSD [9] (h), OQ [17] (i), Ddl1 [16] (j), MJ3D [21] (k), Q_Shao [20] (l), and HV3D (m) objective quality metrics**

**Table 2.7 Statistical performance of different quality metrics for each specific type of distortion**

| Quality Metric | Distortion | Additive Gaussian Noise | | View Compression (Simulcast) | | Blurring | | Brightness Shift | | 3D Video Compression | | View Synthesis | | Depth Map Compression | |
|---|---|---|---|---|---|---|---|---|---|---|---|---|---|---|---|
| | | PCC | SCC | PCC | SCC | PCC | SCC | PCC | SCC | PCC | SCC | PCC | SCC | PCC | SCC |
| PSNR | | 0.6832 | 0.6551 | 0.7098 | 0.7211 | 0.5919 | 0.6190 | 0.4931 | 0.4097 | 0.7400 | 0.7318 | 0.6393 | 0.6238 | 0.6503 | 0.6343 |
| SSIM [13] | | 0.7716 | 0.7192 | 0.7530 | 0.7480 | 0.7159 | 0.7143 | 0.7653 | 0.7804 | 0.6998 | 0.6530 | 0.6539 | 0.6591 | 0.6700 | 0.6916 |
| VQM [158] | | 0.7367 | 0.7120 | 0.6622 | 0.7104 | 0.7496 | 0.7048 | 0.7286 | 0.7687 | 0.6842 | 0.6897 | 0.6934 | 0.6430 | 0.5739 | 0.6178 |
| VIF [135] | | 0.7967 | 0.7982 | 0.7694 | 0.7866 | 0.8232 | 0.7619 | **0.8523** | **0.8711** | 0.7307 | 0.7312 | 0.6059 | 0.6548 | 0.7535 | 0.7727 |
| MOVIE [159] | | 0.8014 | **0.8189** | 0.7497 | 0.7920 | 0.7564 | 0.7524 | 0.7098 | 0.6916 | 0.7195 | 0.7511 | 0.6877 | 0.7171 | 0.6495 | 0.6462 |
| Ddl1 [16] | | 0.6228 | 0.6796 | 0.7620 | 0.8003 | 0.7433 | 0.7190 | 0.7761 | 0.7152 | 0.7200 | 0.7311 | 0.7255 | 0.7138 | 0.8132 | 0.8197 |
| OQ [17] | | 0.7133 | 0.7347 | 0.6859 | 0.7425 | 0.7042 | 0.7714 | 0.6519 | 0.7074 | 0.7411 | 0.7805 | 0.6816 | 0.7196 | 0.7071 | 0.7696 |
| CIQ [18] | | 0.7769 | 0.7868 | 0.7741 | 0.7552 | **0.8325** | 0.8238 | 0.7240 | 0.7289 | 0.7556 | 0.7410 | 0.7123 | 0.7315 | 0.7701 | 0.8447 |
| PHVS3D [19] | | 0.6918 | 0.6791 | 0.7960 | 0.8039 | 0.7244 | 0.7286 | 0.6079 | 0.6048 | 0.8194 | 0.8430 | 0.7366 | 0.7197 | 0.7840 | 0.7711 |
| PHSD [9] | | 0.6484 | 0.6755 | 0.8522 | **0.8789** | 0.7523 | 0.7048 | 0.6249 | 0.6166 | 0.8330 | 0.8572 | 0.7532 | 0.7746 | 0.8285 | 0.8099 |
| MJ3D [21] | | **0.8277** | 0.8098 | 0.8452 | 0.8467 | 0.8123 | **0.8426** | 0.8001 | 0.8319 | 0.8009 | 0.8234 | 0.7219 | 0.7341 | 0.7018 | 0.7422 |
| Q_Shao [20] | | 0.7988 | 0.7812 | 0.8233 | 0.8122 | 0.8278 | 0.8167 | 0.7012 | 0.6911 | 0.7923 | 0.7786 | 0.7088 | 0.7121 | 0.7245 | 0.7098 |
| HV3D | | 0.7994 | 0.7823 | 0.8312 | 0.8466 | 0.8108 | 0.8001 | 0.8412 | 0.8539 | **0.8965** | **0.9010** | **0.8881** | **0.8443** | **0.8603** | **0.8554** |

Note that in the case of simulcast video compression the performance of our quality metric is slightly lower than the 3D video compression case. This is due to the low quality of depth maps generated from the compressed stereo videos in the simulcast coding scenario (in 3D video coding the coded version of reference depth maps are available).

### 2.3.6 Effect of the Temporal Pooling

The performance evaluations demonstrated in Table 2.6 presents statistical comparison between HV3D and various 2D and 3D quality metrics. Except VQM [158], MOVIE [159], and HV3D, the rest of the metrics in Table 2.6 have been originally designed for assessing the quality of images and do not take into account the temporal aspect of video content. To address this, the exponentially weighted Minkowski pooling mechanism [140] is used in conjunction with image quality metrics to convert a set of frame quality scores to a single meaningful score for a video instead of averaging the scores over the frames. Note that the parameters of the exponentially weighted Minkowski pooling are optimized for each quality metric separately (using the training set), to achieve the highest PCC with the MOS scores. The PCC and SCC values between the MOS and different quality metrics before and after applying the temporal pooling are reported in Table 2.8. As it is observed temporal pooling in general tends to improve the performance of the metrics. This improvement is more substantial in the case of the quality metrics such as PSNR and SSIM that have lower assessment performance at the presence of distortions with more temporally variant visual artifacts such as 3D compression, view synthesis, and depth map compression.

### 2.3.7 Sensitivity of Our Method to Constant Exponents

The method we presented in this chapter is based on training our quality model using a training video set and then testing it over a validation video set. For these kinds of approaches to be convincing, it is necessary to study the robustness of the algorithm against the changes in the parameters that are used in the design.

In order to evaluate the robustness of the proposed quality metric to the constant exponents, $\beta_1$, $\beta_2$, and $\beta_3$, we first choose a set of optimal parameters e.g. the ones reported in Section 2.2.3. Then, we evaluate the PCC values when the exponents are slightly changed. Over a wide range of $\beta_1$, $\beta_2$, and $\beta_3$, we observed that the rate of change (derivative) of PCC is not significant. In particular, these parameters are swept over the intervals $0.35 \leq \beta_1 \leq 0.45$, $0.05 \leq \beta_2 \leq 0.15$, and $0.25 \leq \beta_3 \leq 0.35$. Over these intervals, the PCC does not degrade significantly. Minimum PCC value for this set of intervals is 0.8533 (corresponding to SCC value of 0.8572) which occurs at $\beta_1=0.45$, $\beta_2=0.15$, and $\beta_3=0.35$.

### 2.3.8 Complexity of HV3D

Identifying matching areas over the right and left views is considered to be one of the most computationally complex procedures in the HV3D implementation. As explained in Section 2.1.1, to model the cyclopean view for each block within one view, the matching block within the other view is detected through an exhaustive search.

**Table 2.8 Statistical performance of different quality metrics with/without temporal pooling**

| Quality Metric | Averaging the frame quality scores | | With Temporal Pooling | |
|---|---|---|---|---|
| | SCC | PCC | SCC | PCC |
| **PSNR** | 0.6350 | 0.6454 | 0.6603 | 0.6661 |
| **SSIM [13]** | 0.6213 | 0.6844 | 0.6405 | 0.7019 |
| **VQM [158]** | NA | NA | 0.5981 | 0.6660 |
| **VIF [135]** | 0.7204 | 0.7257 | 0.7239 | 0.7422 |
| **MOVIE [159]** | NA | NA | 0.7967 | 0.7527 |
| **Ddl1 [16]** | 0.7321 | 0.7370 | 0.7325 | 0.7501 |
| **OQ [17]** | 0.7900 | 0.7580 | 0.7911 | 0.7790 |
| **CIQ [18]** | 0.7080 | 0.7200 | 0.7140 | 0.7216 |
| **PHVS3D [19]** | 0.8233 | 0.7837 | 0.8247 | 0.7862 |
| **PHSD [9]** | 0.7841 | 0.7911 | 0.7946 | 0.7996 |
| **MJ3D [21]** | 0.8947 | 0.8640 | 0.8988 | 0.8691 |
| **Q_Shao [20]** | 0.7988 | 0.8348 | 0.7999 | 0.8401 |
| **HV3D** | **0.9014** | **0.8959** | **0.9130** | **0.9082** |

As illustrated in Fig. 2.2, $A_R$ is the matching block for $A_L$, which is found by performing an exhaustive search around $A_D$. To reduce the complexity of HV3D, instead of performing a full search, the disparity map is used to find the matching areas within two views. To this end, for each block in the left view, we assume that the horizontal coordinate of the matching block in the right view is equal to the coordinate of the block in the left view plus the disparity of the block in the left view. This approximation as shown in Fig. 2.2 is as if $A_D$ in the right view is chosen as the matching block for $A_L$ in the left view. Using this approach, which is called Fast-HV3D, the computational complexity is reduced significantly. The experimental results show that for "Fast-HV3D" method, PCC is 0.8865, SCC is 0.8962, RMSE is 6.73, and the outlier ratio is 0, which confirms that while the complexity of the Fast-HV3D quality metric is less than that of the original HV3D quality metric, its performance is almost similar (see Table 2.6).

Note that due to the use of the disparity information for modeling the binocular fusion of the two views in Fast-HV3D, the cyclopean view quality component will be slightly correlated to the depth map quality component. However, the disparity information is used in a different way in the two quality components. More specifically, the disparity information incorporated in the cyclopean view generation is only used to find the matching blocks and the goal is to fuse the two views to create the intermediate cyclopean image. However, the depth information used in the depth map quality component is directly used to consider: 1) the effect of smooth/fast variations of depth level, and 2) the effect of distortions on the depth map quality. In addition, please note that the depth map quality component does not utilize the image intensity values at all. The cyclopean view image is a 2D intermediate image, constructed by the fusion of the two views. The depth map specifies how much each object (within the cyclopean view) is perceived outside/inside the display screen. It is observed from Table 2.4 and Table 2.6 that each quality component can predict the overall MOS to some extent. However, only the combination of the two components demonstrates very high accuracy. Note that for the Fast-HV3D method, the performance of the metric is slightly less than regular HV3D, which is due to the usage of median disparity for each block and the resulting correlation between the two quality components.

The computational complexity of different algorithms is usually expressed by the complexity degree order, which is mathematically measured. Mathematical measurement of the computational complexity of various quality metrics in terms of complexity degree order is very difficult. Thus to compare the complexity of different quality metrics, we measure the simulation time for each metric. A comparative experiment was performed on a Win7-64bit Workstation, with Intel Core i7 CPU, and 18 GBs of memory. During the experiment, it was ensured that no other program was running on the machine. Each metric was applied to a number of frames, and the total simulation time was measured. Moreover the simulation time of each metric relative to that of PSNR was calculated as follows:

$$\text{Simulation time relative to } PSNR = \frac{\text{Simulation time for each metric}}{PSNR \text{ simulation time}} \quad (2.11)$$

The average simulation times per one HD frame for different quality metrics, as well as the relative simulation time of each quality metric with respect to that of PSNR are reported in Table 2.9. As it is observed, the complexity of Fast-HV3D is 21.21% less than that of HV3D in terms of simulation time. Although HV3D is 56.5 times more complex than PSNR in terms of simulation time, its complexity is still much less than that of other quality metrics such as PHSD, PHSD-3D or MOVIE, which are 453.43 to 4972.29 times more complex than PSNR. The relative simulation time versus the PCC value for different quality metrics is illustrated in Fig. 2.9. It is observed that the proposed HV3D quality metric and its fast implementation (Fast-HV3D) perform well, while their computational complexity is moderate.

Table 2.9 Complexity of different quality metrics

| Quality Metric | Average Simulation Time per One HD Frame (seconds) | Simulation Time relative to PSNR Simulation Time |
|---|---|---|
| PSNR | 0.7 | 1 |
| SSIM [13] | 1.2 | 1.71 |
| VQM [158] | 65.3 | 93.29 |
| VIF [135] | 6.9 | 9.86 |
| MOVIE [159] | 3480.6 | 4972.29 |
| Ddl1 [16] | 1.3 | 1.86 |
| OQ [17] | 1.3 | 1.86 |
| CIQ [18] | 2.3 | 3.29 |
| PHVS3D [19] | 317.4 | 453.43 |
| PHSD [9] | 323.4 | 462 |
| MJ3D [21] | 46.1 | 65.86 |
| Q_Shao [20] | 260.33 | 371.9 |
| Fast-HV3D | 31.2 | 44.57 |
| HV3D | 39.6 | 56.57 |

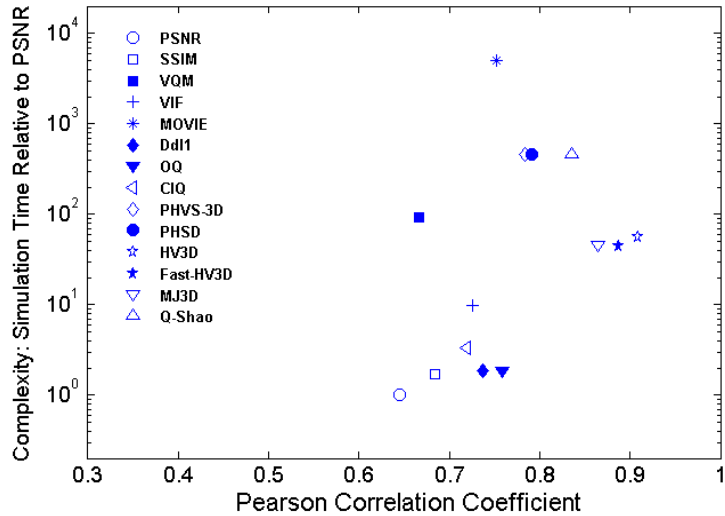
**Figure 2.9 Relative complexity of different metrics versus their PCC values**

In summary, the performance evaluations and Spearman and Pearson correlation ratio analysis showed that our 3D quality metric quantifies the degradation of quality caused by several representative types of distortions very competitively compared to the state-of-the-art quality metrics. HV3D takes into account the viewing settings i.e. distance of the viewer to display as well as the size of the screen.  An early version of HV3D was tailored for mobile 3D applications [128].

## 2.4  Conclusion

In this chapter we proposed a new full-reference quality metric called HV3D for 3D video applications. Our approach models the human stereoscopic vision by fusing the information of the left and right views through 3D-DCT transform and takes to account the sensitivity of the human visual system to contrast as well as the depth information of the scene. In addition, a temporal pooling mechanism is utilized to account for the temporal variations in the video quality. To adjust the parameters of the HV3D quality metric and evaluate its performance, we prepared a database of 16 reference and 208 distorted videos with representative types of distortions. Performance evaluations revealed that the proposed quality metric achieves an average of 90.8% correlation between HV3D and MOS, outperforming the state-of-the-art 3D quality metrics. The proposed metric can be tailored to different applications, as it takes into account the display size (the distance of the viewer from the display) and the video resolution.

# 3  The Effect of Frame Rate on 3D Video Quality and Bitrate

This chapter along with Chapter 4 are dedicated to modeling the important visual attributes in the 3D video content, which are later used for NR 3D video quality assessment. Due to its importance, we dedicate the current chapter to study the effect of motion and frame rate on 3D video quality.

Increasing the frame rate of a 3D video generally results in improved QoE. However, higher frame rates involve a higher degree of complexity in capturing, transmission, storage, and display. The question that arises here is what frame rate guarantees high viewing quality of experience given the existing/required 3D devices and technologies (3D cameras, 3D TVs, compression, transmission bandwidth, and storage capacity). This question has already been addressed for the case of 2D video, but not for 3D. The objective of this chapter is to study the relationship between 3D quality and bitrate at different frame rates. The results of our study are of particular interest to network providers for rate adaptation in variable bitrate channels.

The rest of this chapter is organized as follows: Section 3.1 explains the procedure to prepare the 3D video test set, Section 3.2 provides details on the experiment procedure, Section 3.3 contains the results and discussion, and Section 3.4 concludes the chapter.

## 3.1  Preparation of the 3D Video Dataset

This section provides details on the capturing and preparation of the video data set used in this study, including hardware configurations as well as post-processing steps.

### 3.1.1  Camera Configuration

In order to capture 3D videos for our experiments, we use four cameras of a same model, with identical firmware and camera settings. The cameras are mounted on a custom-made bar and are aligned in parallel. One camera pair is configured to capture 60 fps (two side-by-side cameras on the right side of the bar in Fig. 3.1) and the other camera pair is set to capture 48 fps (two side-by-side cameras on the left side of the bar in Fig. 3.1).

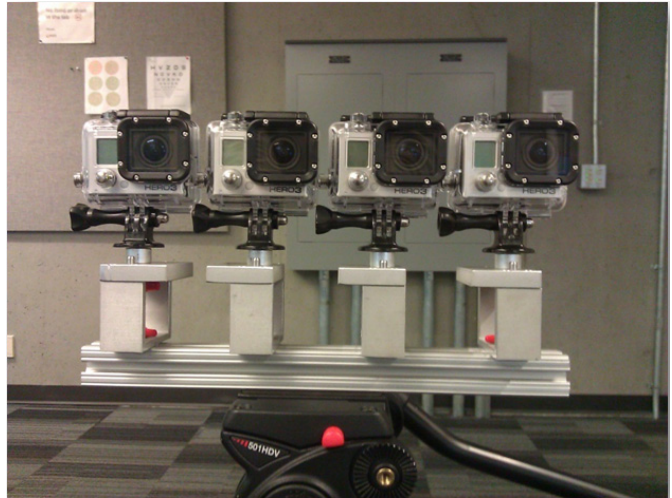
**Figure 3.1 Camera configuration**

To generate 30 fps and 24 fps stereoscopic videos, the captured 60 fps and 48 fps stereoscopic videos are then temporally down-sampled by a factor of two. This is done by starting from the first frame and dropping every other frame in each video. As a result, we obtain 3D videos at four different frame rates from the same scene. These frame rates (24, 30, 48, and 60 fps) are chosen, because they are available options in consumer cameras. Presently, theater content is shot in 24 or 30 frames per second while there is interest to know the effect of 48 fps and 60 fps for 3D.

*3.1.2   Database Capturing*

In our study, GoPro cameras are chosen for capturing the test dataset, because of their small size (which allows us to minimize the difference between the captured stereo pairs) and their capability of capturing high-resolution (HD) videos (1080×1920) at up to 60 frames per second (fps). Since the camera lenses are almost identical and have the same f-number, the camera shutter speed (exposure time) controls the amount of light that reaches the sensor. The shutter speed in these cameras is automatically set to the inverse of the video frame rate [160].

GoPro cameras come with a built-in wide-angle lens, which may cause a fisheye effect at the borders of the picture. During capturing, special attention was given to the contextually important areas to ensure they were not affected by fisheye distortions. This was further enforced by applying the 3D Visual Attention Model (3D VAM) described in [146] to identify the visually important areas of the captured videos. Videos whose visually important areas are affected by fisheye distortion, were excluded from our database. Considering that identical cameras are used for capturing the test dataset, the same amount of fisheye effect exists in all the different frame-rate versions for the same scene. This allows us to conduct a fair comparison among different frame-rate versions of the same scene and studying the effect of frame rate on 3D visual perception.

At the time of capturing, it is ensured that there is no window violation (when part of an object is popping out of the screen, which causes the brain to get confused because of two contradictory depth cues) by properly selecting the framing window.

Six indoor scenes are captured using the camera setup shown in Fig. 3.1, each 10 seconds long. The resolution of the original 3D videos is 1920×1080 (High Definition) for each view and the baseline between cameras are set at 7 [*cm*]. A snapshot of the left view of each scene is shown in Fig. 3.2. Our database is publicly available at [161]. Table 3.1 provides specifications about the captured videos. For each video sequence, the amount of spatial and temporal perceptual information is measured according to the ITU Recommendation P.910 [143] and results are reported in Table 3.1.

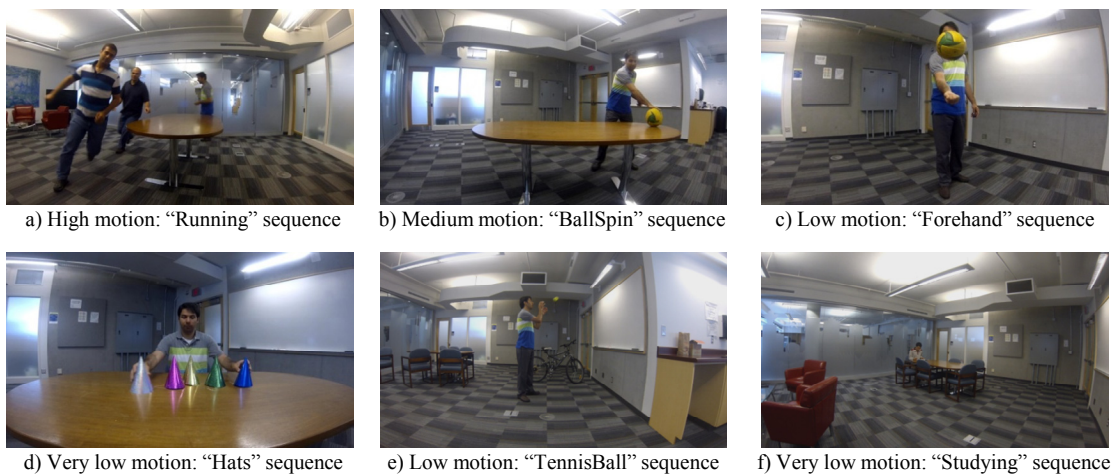

a) High motion: "Running" sequence   b) Medium motion: "BallSpin" sequence   c) Low motion: "Forehand" sequence

d) Very low motion: "Hats" sequence   e) Low motion: "TennisBall" sequence   f) Very low motion: "Studying" sequence

**Figure 3.2 Snapshots of the left views of the 3D video database**

**Table 3.1 Description of the 3D video database**

| Sequence | Resolution | Spatial Complexity (Spatial Information) | Temporal Complexity (Temporal Information) | Depth Bracket | Motion Level |
|---|---|---|---|---|---|
| **Running** | 1920×1080 | High (49.22) | High (22.19) | Wide | High |
| **BallSpin** | 1920×1080 | High (44.35) | Medium (12.05) | Medium | Medium |
| **Forehand** | 1920×1080 | Medium (34.39) | Low (5.89) | Medium | Low |
| **Hats** | 1920×1080 | Medium (35.93) | Medium (10.15) | Narrow | Very low |
| **TennisBall** | 1920×1080 | High (44.27) | Very low (3.45) | Wide | Low |
| **Studying** | 1920×1080 | High (44.17) | Very low (2.87) | Wide | Very low |

For the spatial perceptual information (SI), first the edges of each video frame (luminance plane) are detected using the Sobel filter [144]. Then, the standard deviation over pixels in each Sobel-filtered frame is computed and the maximum value over all the frames is chosen to represent the spatial information content of the scene. The temporal perceptual information (TI) is based upon the motion difference between consecutive frames. To measure the TI, first the difference between the pixel values (of the luminance plane) at the same coordinates in consecutive frames is calculated. Then, the standard deviation over pixels in each frame is computed and the maximum value over all the frames is set as the measure of TI. More motion in adjacent frames will result in higher values of TI. Note that the reported values for spatial and temporal information measures are obtained from the 60 fps version of each sequence, as this version is closer to our visual true-life perception. Fig. 3.3 shows the spatial and temporal information indices of each test sequence, as indicated in [143]. For each sequence shown in Table 3.1, we also provide information about the scene's depth bracket. The depth bracket of each scene is defined as the amount of 3D space used in a shot or a sequence (i.e., a rough estimate of the difference between the distance of the closest and the farthest visually important objects from the camera in each scene) [145]. The captured 3D streams are post-processed to ensure that they are temporally synchronized, rectified, and comfortable to watch. The following subsections elaborate on the applied post-processing schemes.

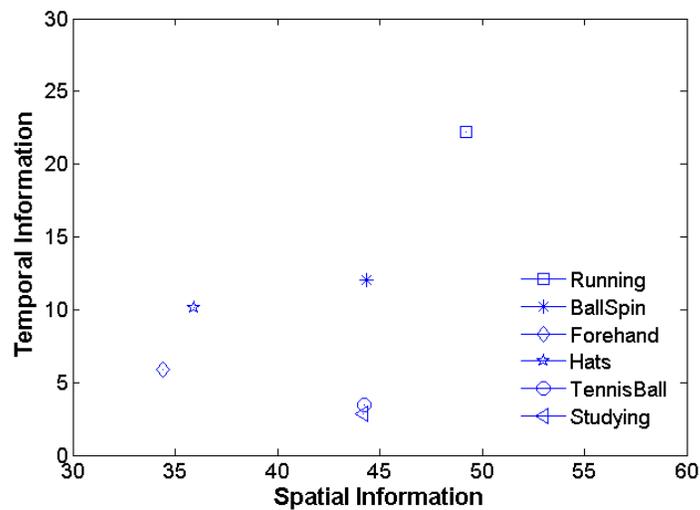

**Figure 3.3 Distribution of the spatial and temporal information over the video database**

*3.1.3 Temporal Synchronization*

To temporally synchronize the cameras, a single remote is used to control the four cameras together so that they all start and finish recording at the same time. However, in practice there are cases where, due to lack of timing accuracy between the remote and cameras, the captured videos are not completely temporally synchronized. In these cases, manual correction is applied to remove a few frames from the videos and achieve temporal synchronization. Considering that the videos are originally captured at 48 fps and 60 fps, manual correction achieves visually acceptable temporal synchronization. Note that temporal synchronization is performed before we temporally down-sample the captured videos to 24 fps and 30 fps.

*3.1.4 Alignment of the 3D Content*

Vertical parallax in stereoscopic video makes viewers uncomfortable, as fusing two views with vertical parallax is difficult for the brain. To reduce the vertical parallax, the four cameras are physically aligned by using identical screws to mount them on a horizontal bar (see Fig. 3.1). This reduces the vertical parallax to some extent, but the videos may still suffer from some vertical misalignment.

To remove the vertical parallax, the left and right views are rectified using an in-house developed software solution. Our approach first extracts the features of the first frame of the left and right views using the Scale Invariant Feature Transform (SIFT) [162]. The features of the left frame are matched to the features of the right frame. The top 10% of all matching features, whose vertical disparities are considerably different from the median disparity value of all matching features, are detected as outliers. These outlier features are removed to ensure the stability of the algorithm. The Cartesian coordinates of rest of the matching features are saved. The median of all the $y$ coordinates of matching points between the two frames, $dy$, is the amount of pixels that each original frame need to be shifted vertically. More specifically, the median vertical mismatch of the matching points gives an estimate of how much each of the views needs to be cropped so that the resulting cropped images contain rectified views without vertical parallax.

Note that since the cameras used for capturing have identical fixed focal length and no digital zoom function, the recorded views do not need zoom correction.

## 3.1.5 Disparity Correction

When 3D videos are captured by parallel stereo cameras, all the objects pop out of the screen as the cameras converge at infinity. In this case, the captured objects are known to have a negative horizontal parallax. This negative parallax occurs when the left-view of an object is located further to the right than the right-view version of the same object. Existing studies show that when objects appear to be in front of the screen for a considerable amount of time they induce visual discomfort [145].

It is a good practice to modify the disparity information (disparity correction) of the content in order to relocate the 3D effect behind the display [145]. To this end, the left frames need to be shifted towards the left and the right frames towards the right, so that the negative horizontal parallax of 3D videos is reduced [145]. To avoid black lines on the vertical edges of the frames, the content is cropped to match the aspect ratio and then it is scaled up. To determine the amount of pixels by which each original frame will be shifted horizontally (i.e., $dx$), we find the largest negative value of all the $x$ coordinates of matching points between the two frames [145]. The negative number with the largest absolute value of the $x$ coordinates represents the photographed point in space that is closest to the cameras ($d_{min}$). Once the frames are shifted according to $dx$, they are cropped and then enlarged using bicubic interpolation so that they maintain their original size before the shifting (1080×1920 pixels) [145].

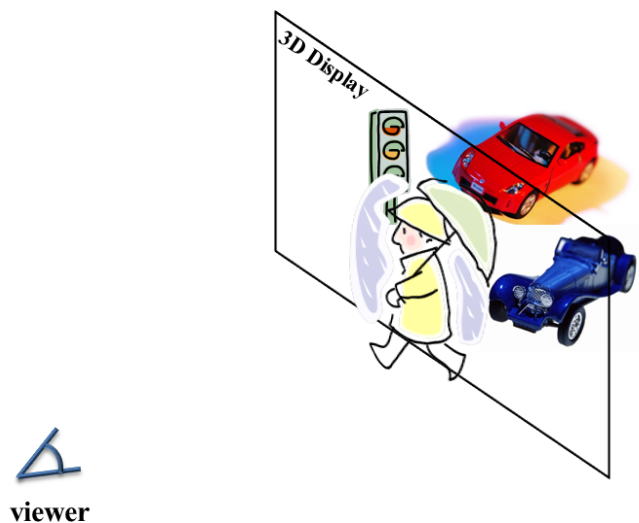

**Figure 3.4 Disparity correction mechanism: Objects are pushed to the 3D viewing comfort zone**

Considering that $d_{min}$ changes over frames in some of the scenes, the shifting parameter ($dx$) is determined based on the frame with the smallest $d_{min}$ and then the same amount of cropping is applied to the rest of the frames. This disparity correction process can improve the 3D quality of experience by 19.86% on average [145]. The effect of disparity correction is mainly reflected in reducing the 3D visual discomfort, which is caused when the eyes try to focus on the screen (accommodation), while the eyeballs try to converge on objects (vergence) that are popping out of the screen. In other words, disparity correction may only shift objects along the depth direction to push them inside the comfort zone.

## 3.2 Subjective Experiments Procedure

The effect of the frame rate on 3D QoE and bitrate was studied through two series of subjective tests using the captured 3D video database. The following subsections elaborate on our experiments.

### 3.2.1 Case Study I: Effect of the Frame Rate on the Quality of 3D Videos

In the first experiment, the goal is to study the relationship between the frame rate and quality of 3D videos and identify the appropriate frame rate for 3D video capturing. To this end, subjective tests are performed to evaluate the visual quality of the 3D test videos at different frame rates. Here, for each scene, the 3D videos with different frame rates are shown one by one and subjects are asked to rate the 3D quality of each video separately and independently from the other videos. To determine if there is a preferred frame rate for 3D viewing, no reference high frame-rate video is provided (unlike [41]). For this case study, five 3D scenes are selected from the captured database and each scene is shown at all four different frame rates (total of 20 stereoscopic videos). The selected scenes are "TennisBall", "BallSpin", "Forehand", "Running", and "Studying".

### 3.2.2 Case Study II: Effect of the Frame Rate on the Quality of the Compressed 3D Videos

In the second experiment, the objective is to study the effect of frame rate at different compression levels on the 3D quality of experience. This study allows determining at each bitrate level what frame rate results in the highest 3D quality of experience. To this end, the video scenes captured at different frame rates are compressed at a variety of bitrates, and subjective tests are performed to evaluate the quality of the compressed 3D videos.

The 3D video sequences are encoded using the emerging 3D HEVC standard (3D-HTM 8.0 reference software [147]) [148][149]. The Quantization Parameter (QP) is set according to the suggested Common Test Conditions by JCT-3V (a joint group under MPEG (ISO/IEC Moving Picture Experts Group) and VCEG (Video Coding Experts Group)) to four different levels of 25, 30, 35, and 40 [142]. The random access high efficiency configuration is used, while the GOP (Group Of Pictures) size is set to eight. Moreover, ALF (Adaptive Loop Filter), SAO (Sample Adaptive Offset), and RDOQ (Rate-Distortion Optimized Quantization) are enabled [148],[149]. In addition, in order to have a fair comparison, the encoding parameters are adjusted according to the frame rate of the 3D videos. For instance, the "intra period" parameter (number of P-frames or B-frames between every two consecutive I-frame) for 24 fps, 30 fps, 48 fps, and 60 fps videos is set at 24, 32, 48, and 64, respectively, to ensure that the size of the intra period is proportional to the frame rate and at the same time is a multiple of the GOP size (i.e., 8). In this case study, four 3D scenes are chosen from the database and for each scene, all the 3D videos with different frame rates (four frame rates) are compressed at four QP levels. As a result, the test set includes a total of 64 3D videos. The selected scenes are "Running", "BallSpin", "Forehand", and "Studying".

### 3.2.3 Test Procedure

Both experiments were conducted according to the viewing conditions specified by the ITU-R recommendation BT.500-13 [152]. Sixteen subjects participated in the first experiment and another eighteen in the second one. The subjects' age ranged from 19 to 29 years old. Before the experiments, all subjects were screened for visual acuity (using Snellen chart), color blindness (using Ishihara chart), and stereovision acuity (using Randot test) and passed the required thresholds. The 3D display used for the experiments was a 64" full HD 3D TV with circularly passive polarized glasses. The screen resolution is the same as the resolution of the videos (1080×1920, which corresponds to an area of 168.2×87.5 [*cm*] on the screen) and therefore there was no need for scaling the videos.

Test sessions were based on the Single Stimulus (SS) method, in which subjects view videos of the same scene with different frame rates in random order. Note that in both case studies, each test session included one randomly selected test video from all the scenes. Thus, the chances that subjects could become biased or exhausted watching the same scene are reduced, while the test sequences are randomized.

Grading was performed according to the Numerical Categorical Judgment (NCJ) method [152], where observers rate video quality based on a discrete range from 0 to 10 (0 representing the lowest quality and 10 representing the highest quality) [152]. As suggested by Quan et al. [154], it was explained to the subjects that the term "quality" in general means how pleasant they think a video looks. Specifically, they were asked to rate the quality based on a combination of different factors such as "naturalness" [4][154], "comfort" [5], "depth impression", "sharpness", and "temporal smoothness" [145],[154],[163]. There was a "training" session before the "test" session, so that the subjects become familiar with the videos and the test structure. During the training period participants were explained how/what to grade watching each test video. In order to minimize the effect fisheye distortions could have on the subjective evaluations, information about the fisheye effect was given to the subjects to familiarize them with this type of distortion and thus help them judge the perceptual quality of the videos without taking into account the fisheye effect. Following what is considered common practice in such tests, even though a training session was provided before each test session, a few "dummy" sequences were shown at the beginning of each test session [152]. The scores for the dummy sequences were excluded from the analysis, as their objective is to familiarize the subjects with the test procedure at the beginning of the test session.

After collecting the subjective test results, the outlier subjects were detected and their scores were removed from the analysis. Outlier detection was performed according to the ITU-R BT.500-13 recommendation, Annex 2 [152]. In the outlier detection process, the kurtosis coefficient is calculated to measure how well the distribution of the subjective scores can be represented using a normal distribution. Through this process it was found that there was no outlier in the first experiment, while there were two outliers in the second case study.

## 3.3 Results and Analysis

Once the experiment data is collected, the Mean Opinion Score (MOS) for each video is calculated as the average of the scores over the subjects set. In order to ensure the reliability of these measurements, a confidence interval of 95% is calculated [152].

*3.3.1 Case I: 3D Quality versus Frame Rate for Uncompressed 3D Sequences*

In the first case study, the quality of the original video set with different frame rates was subjectively evaluated. Fig. 3.5 shows the average perceived 3D quality at different video frame rates for the entire video database with 95% confidence interval. As it is observed, the 3D videos with frame rates of 48 fps and 60 fps are highly preferred and rated as excellent quality (MOS greater than 8). On the other hand, the 3D videos with the frame rate of 24 fps are rated as poor/fair quality (MOS between 2 and 5). Considering that the MOS of 3D videos at 60 fps with 95% confidence interval can reach 9.8, one could conclude that increasing the frames rates of 3D videos more than 60 fps may not result in visually distinguishable quality for viewers and will just increase the complexity of capturing, transmission, and display. It is also observed that there is a significant difference between the quality of 3D videos with 24 fps and the ones with 48 fps and 60 fps. In particular, average MOS-difference between videos in 60 fps and videos in 24 fps is around 5.8, indicating a high preference for these high rates.

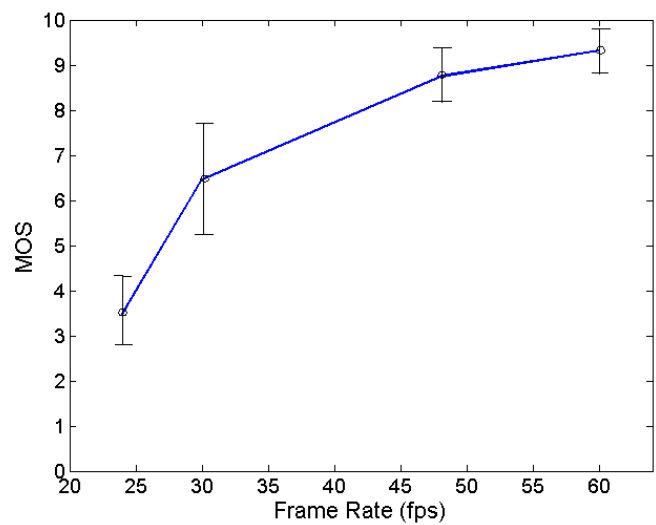

**Figure 3.5 Average perceived 3D video quality (MOS) at different frame rates**

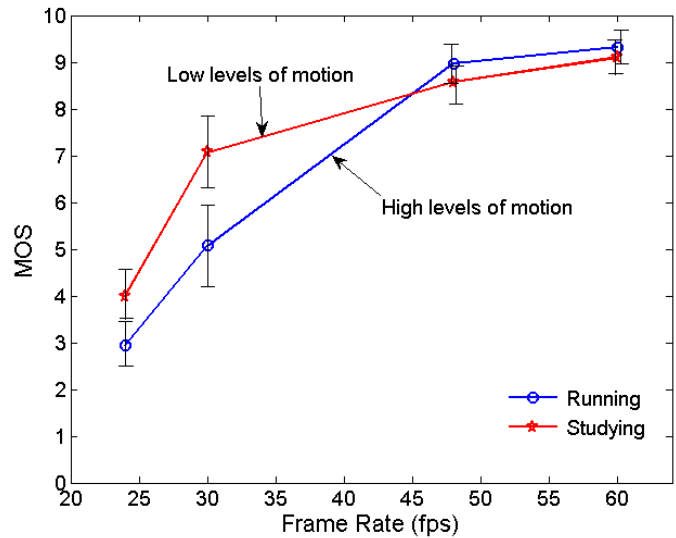

**Figure 3.6 3D video quality (MOS) at different frame rates for videos with low ("Studying" sequence) and high ("Running" sequence) levels of temporal motion**

In order to understand the effect of motion on the perceived 3D video quality, we plot 3D quality versus frame rate for two videos with low and high levels of motion (sequence "Running" for fast motion and sequence "Studying" for slow motion) in Fig. 3.6. It can be seen that the difference between the MOS values at 30 fps and 48 fps increases when the motion level is high. In particular, 3D quality drops by 3.9 in terms of MOS when the frame rate decreases from 48 fps to 30 fps for the video with fast motion, whereas the quality drops only by 1.5 in terms of MOS in the case of low-motion video. In other words, when a scene contains fast moving objects, low frame rates (in this case 24 fps and 30 fps) result in an unpleasant 3D experience. Based on this observation it is recommended to capture 3D scenes with high motion at higher frame rates than 30 fps to ensure the motion in the scene appears smooth and the 3D quality of experience is improved.

### 3.3.2 Case II: 3D Quality versus Frame Rate and Bitrate for Encoded 3D Sequences

In the second case study, the quality of compressed 3D videos with different frame rates is subjectively evaluated. After collecting the results and removing the outliers, the average MOS for each video is calculated at different frame rates and bitrates. Fig. 3.7 illustrates the relationship between 3D quality of experience and frame rate at different bitrates for 3D video sequences with variety of motion levels. By comparing the results for different video sequences, it is observed that in general 3D videos with higher bitrates and higher frame rates are more pleasant to viewers. Another useful observation derived from Fig. 3.7 is that, except for very low bitrates, subjects prefer to watch a high frame rate version of a 3D video rather than its lower frame rate version, even though the high-frame rate one is more compressed. Moreover, the illustrated results in Fig. 3.7 for test sequences with different motion levels show that the gap between the perceptual qualities of different frame-rate versions of the same 3D video becomes more significant, if the scene includes higher motion levels. In Fig. 3.7.a where the test sequence ("Running") includes fast moving objects in the scene, the MOS of the higher frame-rate versions (48 and 60 fps) are higher than those of lower frame rates (24 and 30 fps).

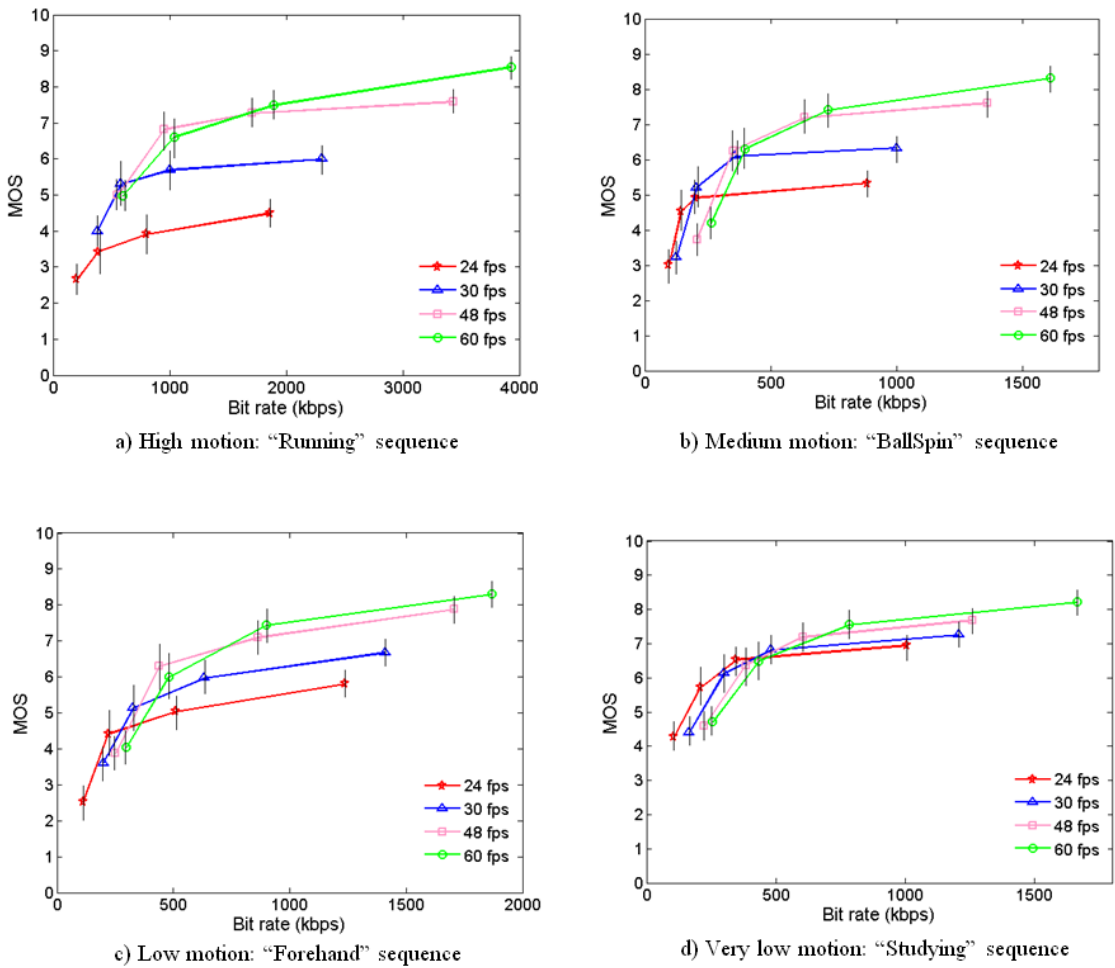

**Figure 3.7 3D quality depicted at different bitrates for various frame rates, in four different levels of motion: a) High motion, b) Medium motion, c) Low motion, and d) Very low motion**

The difference becomes quite significant at higher bit rates. Even the highly compressed 60 fps and 48 fps versions of the 3D videos (low bitrate of 1000 kbps) are preferred over the 24 fps version of the same video with slight compression (high bitrate of 2000 kbps or more). This suggests in the case of 3D video content with high motion, to transmit the high frame rate version of the content (if available) at the channel bitrate, instead of the low frame-rate version of the video.

Fig. 3.7.b and 3.7.c show the results of our experiment for 3D content with medium and low levels of motion. As it is observed the overall 3D quality of higher frame-rate versions of the sequence is still higher than that of lower frame-rate versions (except at very low bitrates, less than 500 kbps), but the MOS difference is not as high as the case where the motion level of the scene is high. In the case where the motion level in the scene is very low, as it is observed from Fig. 3.7.d, the perceptual quality of different frame-rate versions of the 3D video sequence are quite similar at the same bitrate level. In other words, frame rate is no longer a contributing factor to the 3D quality and the 3D quality is controlled by bitrate here. This is because when the motion level is low, temporal smoothness provided by frame-rate increase is no longer noticeable.

The subjective test results in Fig. 3.7 suggest that based on the amount of the available bandwidth (required bitrate), one should choose the appropriate frame rate, which provides the maximum 3D quality. More precisely, to ensure the highest possible 3D quality is achieved at high bitrate, the higher frame rate version of the 3D content shall be transmitted (if available). In case where the bandwidth drops from a very high value, then the frame rate needs to be adjusted (reduced) according to the available bandwidth. At very low bitrates, depending on the application, dropping one of the 3D streams and delivering 2D content is also suggested [164]. Following these guidelines allows network providers to deliver maximum possible 3D video quality by controlling and adjusting the frame rate at different bitrates.

We used the statistical T-test to determine if there is a significant difference between the quality scores obtained from sixteen subjects for different frame-rate versions of each sequence compressed with a specific QP setting. Table 3.2 summarizes the T-test results. The null hypothesis is if the perceptual quality of two different frame rate versions of a test sequence compressed with a specific QP setting is statistically equal. The significance level is set at 0.01. As it is observed for the "Studying" sequence, which has very low motion, there is no strong presumption against the null hypothesis in all QP levels. In other words the perceptual quality of different frame rate versions of "Studying" video sequence compressed with a specific QP setting is statistically equal (with one exception at QP of 25 and frame rate pair of (48, 60)). For the rest of the sequences with different motion levels, the null hypothesis is always rejected for low QPs of 25 and 30 and for frame rate pairs of (24, 30) and (30, 48). This implies that at high bitrates (small QPs), there is a significant difference between the quality of 24fps, 30fps, and 48fps versions of the compressed videos. By comparing the results of 48fps and 60 fps for the test sequences with low to high motion levels, it is observed that the perceptual quality of the 48fps and 60fps versions of the video sequences is equal at all QP levels, except for the case where medium or very high motion level is present and QP is very small (high bitrate). The statistical difference test results also show that in the case of medium, low, and very low motion in a scene, when the bit rate is low (high QP values of 35 and 40), no difference in the subjective quality score of (24, 30) and (30, 48) frame rate pairs is reported (with one exception). In the presence of high motion ("Running" sequence), however, distinct statistical difference is reported when the bit rate is low (QPs of 35 and 40) and frame rate pairs of (24, 30) or (30, 48) are being compared.

## 3.4 Conclusion

In this chapter, the relationship between the 3D quality of experience, bitrate, and frame rate was explored. First, a database of 3D sequences was created, involving scenes with different motion levels and frame rates of 24 fps, 30 fps, 48 fps, and 60 fps. Then, the quality of these videos was subjectively evaluated. Results of this experiment showed that subjects clearly prefer 3D videos with higher frame rates (48 and 60 fps) as there is a significant improvement in 3D quality when higher frame rates are used.

Table 3.2 Statistical difference (P-values) between the subjective scores using T-test

|  | Running (high motion) | | | Ball Spin (medium motion) | | | Forehand (low motion) | | | Studying (very low motion) | | |
| --- | --- | --- | --- | --- | --- | --- | --- | --- | --- | --- | --- | --- |
|  | $P_{60-48}$ | $P_{48-30}$ | $P_{30-24}$ | $P_{60-48}$ | $P_{48-30}$ | $P_{30-24}$ | $P_{60-48}$ | $P_{48-30}$ | $P_{30-24}$ | $P_{60-48}$ | $P_{48-30}$ | $P_{30-24}$ |
| QP 25 | 0 | 0 | 0 | 0 | 0 | 0 | 0.05 | 0 | 0 | 0.01 | 0.09 | 0.18 |
| QP 30 | 0.43 | 0 | 0 | 0.48 | 0 | 0 | 0.27 | 0 | 0 | 0.33 | 0.38 | 0.49 |
| QP 35 | 0.41 | 0 | 0 | 0.61 | 0.02 | 0.07 | 0.53 | 0.05 | 0.28 | 0.59 | 0.41 | 0.51 |
| QP 40 | 0.63 | 0 | 0 | 0.19 | 0.14 | 0.24 | 0.64 | 0.30 | 0 | 0.60 | 0.67 | 0.75 |

Moreover, the same experiment revealed that increasing the frame rate to more than 60 fps, does not noticeably improve the 3D video quality.

In the second experiment, the stereoscopic scenes with four different frame rates were encoded at four compression levels (QPs of 25, 30, 35, and 40). Subjective quality evaluations of these 3D videos showed that for scenes with fast moving objects, the effect of frame rate on the overall perceived 3D quality is more dominant than the compression effect, whereas for scenes with low motion levels the frame rate does not have a significant impact on the 3D quality. In other words, higher frame rates improve the 3D QoE significantly when there is fast motion in a scene. In addition, high frame rate 3D videos with higher compression rates are preferred over slightly compressed but low frame rate 3D videos. The subjective test results suggest that in cases where the available bandwidth for video transmission drops (variable bandwidth channel), reducing the frame rate instead of increasing the compression ratio helps achieve the maximum possible 3D quality of experience level with respect to bandwidth.

In summary, our study suggests that the best practical frame rate for 3D video capturing is 60 fps, as it delivers excellent quality of experience and producing such content is possible by using available capturing devices. In fact, going beyond this frame rate does not yield visually noticeable improvement while the required effort and resources are not justifiable.

The findings of this chapter regarding the effect of motion and frame rate on 3D video quality are combined with the other saliency attributes in the next chapter to form a 3D visual attention model, which is used in Chapter 5 for NR/FR quality assessment.

# 4  Saliency Prediction for Stereoscopic 3D Video

This chapter explores stereoscopic video saliency prediction by exploiting both low-level attributes such as brightness, color, texture, orientation, motion, and depth, as well as high-level cues such as face, person, vehicle, animal, text, and horizon. Our model starts with a rough segmentation and quantifies several intuitive observations such as the effects of visual discomfort level, depth abruptness, motion acceleration, elements of surprise, size and compactness of the salient regions, and emphasizing only a few salient objects in a scene. A new fovea-based model of spatial distance between the image regions is adopted for considering local and global feature calculations. To efficiently fuse the conspicuity maps generated by our method to one single saliency map that is highly correlated with the eye-fixation data, a random forest based algorithm is utilized. The performance of the proposed saliency model is evaluated against the results of an eye-tracking experiment, which involved 24 subjects and an in-house database of 61 captured stereoscopic videos. Our video saliency benchmark database is publicly available with this thesis.

The rest of this chapter is organized as follows: Section 4.1 explains the proposed saliency prediction method, Section 4.2 elaborates on the database creation, and subjective tests, results and discussions are provided in Section 4.3, and Section 4.4 concludes the chapter.

## 4.1  Proposed Saliency Prediction Method

As mentioned in Chapter 1, the existing monocular saliency models are not able to accurately predict the attentive regions when applied to 3D image/video content, as they do not incorporate depth information. Fig. 4.1 demonstrates an example where 2D VAMs fail to predict the stereo saliency. Also, a brief overview of the state-of-the-art 3D VAMs and eye-tracking datasets are provided in Table 4.1 and Table 4.2, respectively.

Our proposed visual attention model takes into account various low-level saliency attributes as well as high-level context-dependent cues. In addition, several intuitive observations are quantified and considered in the design of our VAM. Once the feature maps are extracted, a random-forest-based algorithm is adopted to train a model of saliency prediction. The flowchart of the proposed method is illustrated in Fig. 4.2. The following subsections elaborate on the details of our model.

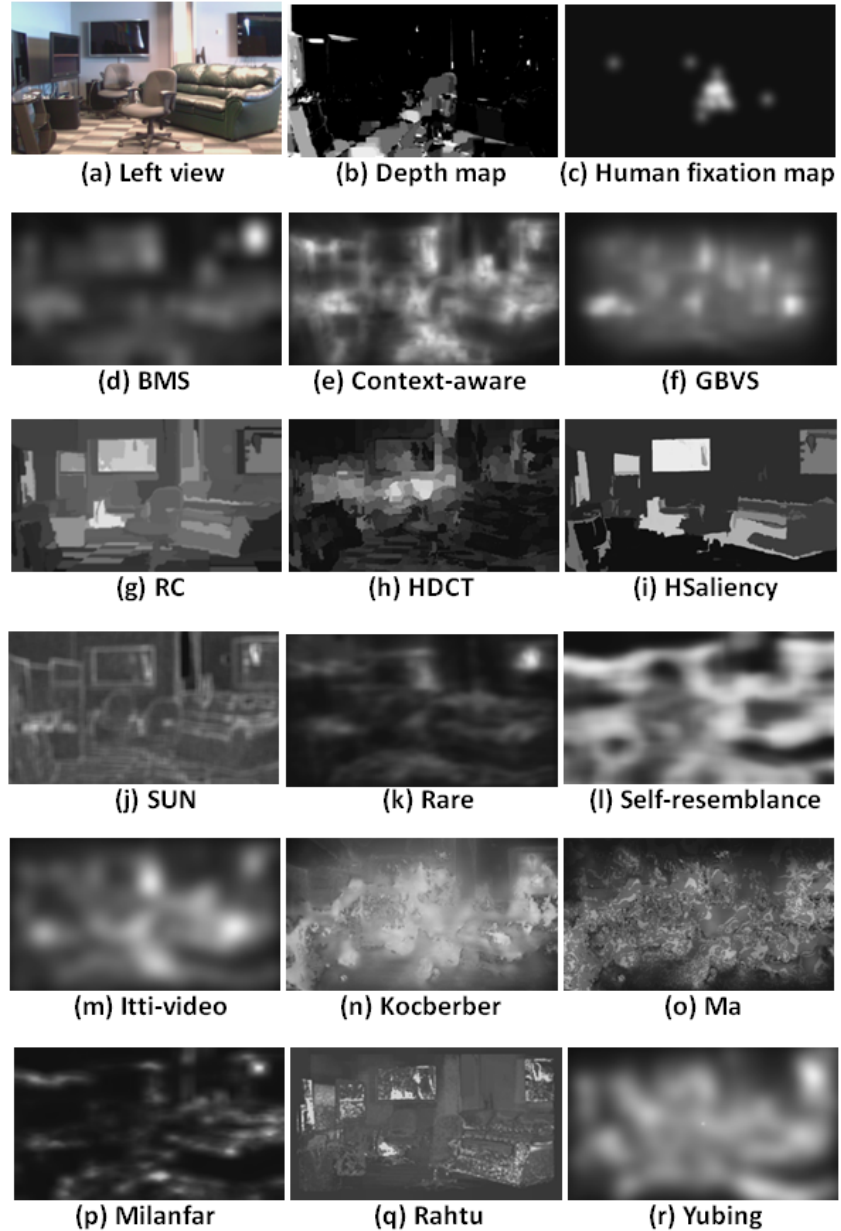

Figure 4.1 An example showing the inaccuracy of 2D saliency prediction methods when applied to a stereoscopic image pair: (a) The left view of the image, (b) depth map, (c) human fixation map from subjective tests, and generated saliency maps using different methods: (d) BMS [52], (e) context aware [64], (f) GBVS [56], (g) RC [58], (h) HDCT [165], (i) HSaliency [166], (j) SUN [67], (k) Rare [62], (l) self-resemblance static [65], (m) Itti-video [57], (n) Kocberber [167], (o) Ma [168], (p) self-resemblance dynamic [65], (q) Rahtu [169], (r) Yubing [170]. The chair is the most salient object as it stands out in 3D due to its depth

Table 4.1 Existing stereoscopic 3D visual attention models at a glance

| 3D-VAM | Type | Year | Description | Features | Feature fusion method | Validation dataset |
|---|---|---|---|---|---|---|
| Maki [72] | 3D Image | 1996 | Direct use of depth map as a weighting factor in conjunction with an existing saliency detection method | Motion and disparity | Pursuit and saccade modes (AND operator) | Qualitative evaluation |
| Ouerhani [75] | 3D Image | 2000 | Use of the model presented in [57] for three depth-based proposed features and combine them | Depth, surface curvature, depth gradient, and color | Weighted average similar to [57] | Qualitative evaluation |
| Potapova [171] | 3D Image | 2011 | Three depth-based features are proposed and combined with 2D saliency features | Surface height, orientation, edges, color, and intensity | Average and multiplication | Object labeling accuracy |
| Park [172] | 3D Image | 2012 | Combine the 2D VAM model of Itti [57] with a map of skin detection and visual-comfort-weighted disparity map | Intensity, color, orientation, disparity, and faces | Multiplication | Qualitative evaluation |
| Niu [70] | 3D Image | 2012 | Two disparity-based conspicuity maps (one from disparity contrast and another one from 3D comfort zone) combined | Disparity | multiplication | Object detection |
| Lang [76] | 3D Image | 2012 | Measure saliency probability at every depth Level to combine the resulted depth saliency with existing 2D VAMs | Depth | Average or multiplication | Eye-tracking: 600 scenes and 80 users |
| Wang [69] | 3D Image | 2013 | Generate a depth saliency map using a Bayesian approach to be combined with existing 2D VAMs | Disparity | Average | Eye-tracking test: 18 scenes and 35 users |
| Iatsun [173] | 3D Image | 2014 | Generate artificial depth from a 2D view and combine it with an existing 2D VAM | Luminance, color, orientation, and depth | Logarithmic multiplication | Eye-tracking test: 27 scenes and 15 users |
| Jiang [122] | 3D Image | 2014 | Create a depth saliency map (based on foreground and background extraction) and combine it with VAM of [123] | The ones in VAM of [123] along with disparity | Weighted average | Subjective quality assessment |
| Ju [174] | 3D Image | 2014 | Measure disparity gradient in 8 directions to generate a Anisotropic Center-Surround Difference (ACSD) map purely from disparity | Disparity | No fusion needed | Object labeling accuracy |
| Fang [68] | 3D Image | 2014 | Create feature maps form DCT coefficients of 8×8 patches | Color, luminance, texture, and depth | Weighted sum according to compactness | Eye-tracking test: 18 scenes and 35 users |
| Fan [175] | 3D Image | 2014 | Depth brightness, color contrast, and spatial compactness are measured, and combined to generate the region-level saliency map | Luminance, color, and disparity | Summation | Object labeling accuracy |
| Chamaret [74] | 3D Video | 2010 | ROIs are detected using the 2DVAM of [106] and refined by the disparity map | Luminance, color, motion, and disparity | Multiplication | ROI detection accuracy |
| Zhang [73] | 3D Video | 2010 | Combine depth map with an existing 2D VAM of [57] and motion saliency map | Features of [57] along with motion and disparity | Weighted average | Qualitative evaluation |
| Coria [146] | 3D Video | 2012 | Linear combination of the disparity map (as a conspicuity map) with 2D saliency features | Texture, disparity, motion, color, intensity | Average | Reframing performance |
| Kim [71] | 3D Video | 2014 | Combine "saliency strength" for different features taking into account size, compactness, and visual discomfort | Luminance, color, disparity, and motion | Weighted sum | Eye-tracking: 5 scene types and 20 users |
| Proposed: LBVS-3D [124] | 3D Video | 2015 | Learn the important saliency features by training a Random Forest model for saliency prediction | Luminance, color, texture, motion, depth, face, person, vehicle, text, animal, horizon | Random forest learning method | Eye-tracking: 61 scene types and 24 users |

Table 4.2 Description of different eye-tracking datasets

| Dataset | Scenes | Type | Year | Resolution | Subjects | Sampling freq. (Hz) | Viewing dist. (*cm*) | Screen diag. (in) | Screen type | Length (sec) |
|---|---|---|---|---|---|---|---|---|---|---|
| **IRCCyN Image 1 [106]** | 27 | 2D Image | 2006 | ~768×512 | 40 | 50 | NA | NA | CRT | 15 |
| **TUD Image 1 [93]** | 29 | 2D Image | 2009 | varying | 20 | 50 | 70 | 19 | CRT | 10 |
| **VAIQ [96]** | 42 | 2D Image | 2009 | varying | 15 | NA | 60 | 19 | LCD | 12 |
| **TUD Interactions [95]** | 54 | 2D Image | 2011 | 768×512 | 14 | 50/60 | 70 | 17 | CRT | NA |
| **GazeCom Image [88]** | 63 | 2D Image | 2010 | 1280×720 | 11 | 250 | 45 | 22 | CRT | 2 |
| **IRCCyN Image 2 [89]** | 80 | 2D Image | 2010 | 481×321 | 18 | 50 | 40 | 17 | LCD | 15 |
| **KTH [90]** | 99 | 2D Image | 2011 | 1024×768 | 31 | NA | 70 | 18 | CRT | 5 |
| **LIVE DOVES [91]** | 101 | 2D Image | 2009 | 1024×768 | 29 | 200 | 134 | 21 | CRT | 5 |
| **Toronto [63],[92]** | 120 | 2D Image | 2006 | 681×511 | 20 | NA | 75 | 21 | CRT | 4 |
| **TUD Image 2 [94]** | 160 | 2D Image | 2011 | 600×600 | 40 | 50 | 60 | 17 | CRT | 8 |
| **McGill ImgSal [86]** | 235 | 2D Image | 2013 | 640×480 | 21 | 60 | 70 | 17 | LCD | NA |
| **FiFa [87]** | 250 | 2D Image | 2007 | 1024×768 | 7 | 1000 | 80 | NA | CRT | 2 |
| **MIT Benchmark [83]** | 300 | 2D Image | 2012 | ~1024×768 | 39 | 240 | 61 | 19 | NA | 3 |
| **NUSEF [78]** | 758 | 2D Image | 2010 | 1024×860 | 13 | 30 | 76 | 17 | LCD | 5 |
| **MIT CVCL [79]** | 912 | 2D Image | 2009 | 800×600 | 14 | 240 | 75 | 21 | CRT | NA |
| **MIT CSAIL [53]** | 1003 | 2D Image | 2009 | ~1024×768 | 15 | NA | 61 | 19 | NA | 3 |
| **MIT LowRes [85]** | 1544 | 2D Image | 2011 | 1024×860 | 8 | 240 | 61 | 19 | NA | 3 |
| **SFU [80]** | 12 | 2D Video | 2012 | 352×288 | 15 | 30 | 80 | 19 | LCD | 3-10 |
| **GazeCom Video [88]** | 18 | 2D Video | 2010 | 1280×720 | 54 | 250 | 45 | 22 | CRT | 20 |
| **ASCMN [107]** | 24 | 2D Video | 2012 | VGA-SD | 13 | NA | NA | NA | NA | 2-76 |
| **TUD Task [100]** | 50 | 2D Video | 2012 | 1280×720 | 12 | 250 | 60 | 17 | CRT | 20 |
| **USC CRCNS Orig. [101]** | 50 | 2D Video | 2004 | 640×480 | 8 | 240 | 80 | 22 | CRT | 6-90 |
| **USC VAGBA [102]** | 50 | 2D Video | 2011 | 1920×1080 | 14 | 240 | 98 | 46 | LCD | 10 |
| **IRCCyN Video 1 [98]** | 51 | 2D Video | 2009 | 720×576 | 37 | 50 | 276 | 37 | LCD | 8-10 |
| **DIEM [97]** | 85 | 2D Video | 2011 | SD-HD | 42 | 1000 | 90 | 21 | NA | 27-217 |
| **IRCCyN Video 2 [99]** | 100 | 2D Video | 2010 | 720×576 | 30 | NA | 150 | 40 | LCD | 10 |
| **USC CRCNS MTV [82]** | 523 | 2D Video | 2006 | 640×480 | 16 | 240 | 80 | 22 | CRT | 1-3 |
| **Actions [81]** | 1857 | 2D Video | 2012 | 640×480 | 16 | 500 | 60 | 22 | LCD | <60 |
| **3DGaze database [69]** | 18 | 3D Image | 2013 | varying | 35 | 500 | 93 | 26 | LCD | 15 |
| **HVEI2013 [103]** | 54 | 3D Image | 2013 | 1920×1080 | 15 | 50 | 234 | 42 | LCD | 20 |
| **NUS3D [76]** | 600 | 3D Image | 2012 | 640×480 | 80 | NA | 80 | NA | LCD | 6 |
| **EyeC3D [104]** | 8 | 3D Video | 2014 | 1920×1080 | 21 | 60 | 180 | 46 | LCD | 8-10 |
| **IRCCyN 3D Video [105]** | 47 | 3D Video | 2014 | 1920×1080 | 40 | 60 | 93 | 26 | LCD | NA |
| **Proposed [176]** | 61 | 3D Video | 2015 | 1920×1080 | 24 | 250 | 183 | 46 | LCD | 8-12 |

### *4.1.1 Bottom-up Saliency Features*

The proposed model includes luminance, color, texture, motion, and depth as low-level saliency features. Each of them is explain in the following subsections. Note that since the position of objects is slightly different between the left and right views, with the exception of depth features, the rest of the features (the ones exploited from brightness, color, texture, and motion information) are extracted from the view of the video for which the depth map is available. In our experiments, for each video both the left and right views are initially available. We calculate the left-to-right disparity (which corresponds to disparity map of the right view) using the MPEG DERS software [131]. Therefore, the right view is used for computing the 2D saliency attributes. The motivation behind the selection of the right view is that humans are mostly right-eye dominant (approximately 70%) [177].

*4.1.1.1 Segmentation*

During the pre-attentive stage of HVS, the visual information of each scene is partitioned to different regions. Computational resources are then allocated to each region based on its relative importance. Similarly, we perform a rough segmentation on the right view picture and assign bottom-up saliency attributes to each segment separately, by averaging the specified pixel-wise saliency values over each segment. In our implementation, we use the Edge Detection and Image Segmentation (EDISON) System proposed by Comanicu et al. [178].

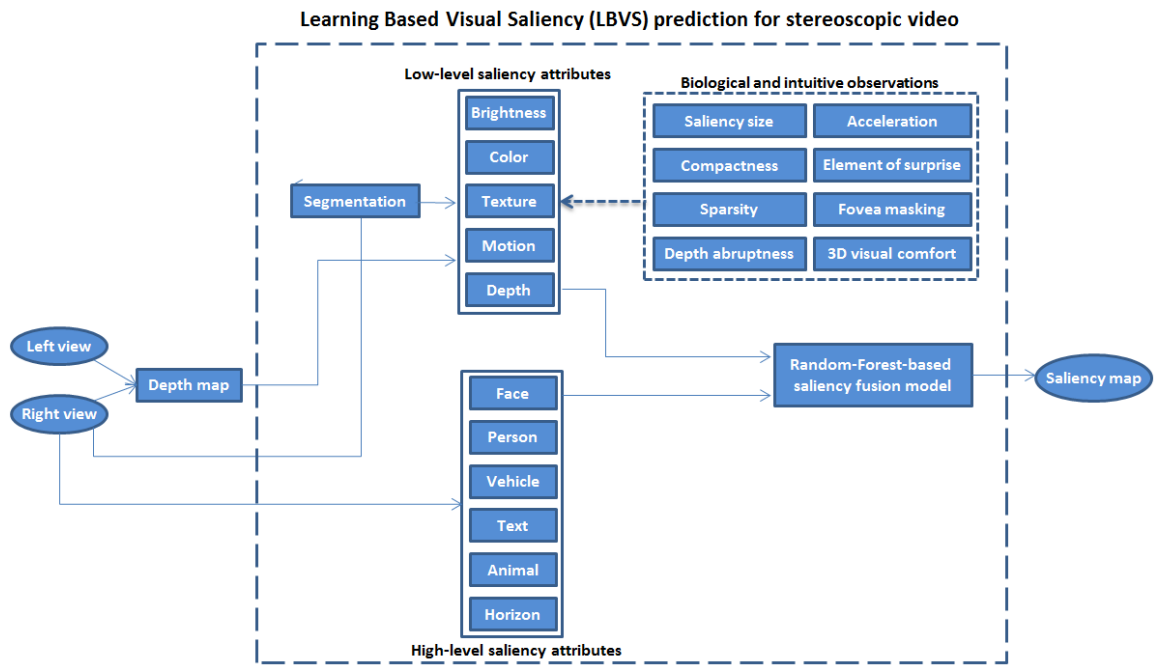

**Figure 4.2 Proposed computational model for visual saliency prediction for stereoscopic video**

*4.1.1.2 Brightness Map*

Studies have shown that human attention is directed towards areas with higher brightness variations in a scene [51],[53]. In our model, we include a map of local brightness variances. To this end, each frame of the right view is transformed to the YUV color space first. Then, local variances are calculated in a circular neighborhood around each pixel using the following formula:

$$\sigma_x^2 = \frac{1}{size_{\{NHOOD\}} - 1} \sum_{i \in NHOOD} (Y_x - Y_i)^2 \qquad (4.1)$$

where $Y_x$ and $Y_i$ represent the brightness intensity values of the center pixel and surrounding pixels respectively, *NHOOD* is a circular neighborhood, *x* denotes the current pixel, and *i* is pixel index over the outer area. The size and shape of the outer area used for variance calculation are chosen based on the fovea photoreceptor concentration, which is explained in Section 4.1.1.7. The variance map is normalized to [0-1].

The resulting brightness variance map contains local brightness variances. However, attention is directed towards certain areas of an image based on both local and global scene properties. Therefore, we adjust the value of the brightness variance map for each pixel as follows. For each pixel, the weighted average difference between its brightness variance and the brightness variance of the surrounding pixels is calculated, resulting in a brightness variance contrast map. This process is known as "center-surround operation", and the Differences-of-Gaussians (DoG) are common approaches for computing differences [51]. Here, instead of the Gaussians, we use a circular mask, which is designed based on fovea photoreceptor concentration (see 4.1.1.7). Hereafter, we refer to this mask as "fovea mask"). The fovea mask assigns different weights to different pixels based on their distance from the pixel located at the center of the mask. The center-surround operation for globalizing the variance map is performed as follows:

$$S_{Brightness}^{K} = \frac{1}{n_K} \sum_{i \in R_K} \sum_{\substack{j \in fovea\ mask \\ j \neq i}} \left( |Var_{diff}(i,j)| \times Fovea_{mask}(j) \right) \qquad (4.2)$$

where $S_{Brightness}^{K}$ is the saliency value assigned to the $K^{th}$ region ($R_K$), $n_K$ is the number of pixels in the $K^{th}$ region, *i* denotes the $i^{th}$ pixel in the current segment, *j* is the $j^{th}$ pixel in the fovea mask centered at *i*, and $Var_{diff}$ is the difference between $i^{th}$ and $j^{th}$ pixels in the brightness variance map (obtained from Eq. 4.1). The center-surround operand results in a brightness variance contrast map in which every segment (or roughly every object) is assigned with one saliency probability.

*4.1.1.3 Color Maps*

Color is one of the most important channels among human senses, accounting for 80% of the visual experience [179]. For the proposed model, the following feature maps are extracted from the color information of each scene:

1) <u>Color histogram map</u>: Naturally, humans tend to look at the objects with colors that stand out in a scene, i.e., rare colors tend to attract human attention. To account for the effect of color rarity, we compute a color histogram for each picture, based on the occurrence probability of each of the three color channels in the RGB space (each channel represented by 8 bits). Suppose *p(R=r, G=g, B=b)=P*. We define the values of $e^{\frac{-P}{\bar{P}}}$ as a saliency map related to color rarity and call it "histogram saliency map" ($\bar{P}$ is the average of *P*). The exponential function is particularly chosen to project the probability values in the interval [0-1] and expose a local maximum where a color is rare.

2,3) <u>Color variance contrast maps</u>: HVS is highly sensitive to color contrast. Similar to the brightness variance contrast map generated before (see 4.1.1.2), two new color saliency maps are created for *a\** and *b\** color components in the CIE *L\*a\*b\** domain. This color space is particularly chosen due to its uniform chromaticity properties [180].

4,5) <u>Warmth and saturation color maps</u>: Experiments showed that warm and saturated colors are generally salient to HVS [181],[182]. Warm colors dominate their surroundings regardless of the existence of color contrast in the background [182]. Highly bright and saturated colors are salient regardless of their associated hue value [182]. The reason is partially due to the eye sensitivity to these types of colors [180], and partially due to human natural instincts, which interpret warm and saturated colors (e.g., saturated red) as a potential threat. To account for the effect of warm colors, the color temperature of the right view picture is calculated first. In general, warm colors correspond to low temperatures and vice versa. The color warmth map is defined as the inverse of the color temperature map. To account for saliency based on saturation of colors, we follow the saturation formula of Lübbe [183]:

$$\text{Saturation } map = \frac{C^*_{ab}}{\sqrt{C^{*2}_{ab} + L^{*2}}} \qquad (4.3)$$

where $L^*$ is the lightness (brightness) and $C^*ab$ denotes the chroma of the color calculated as:

$$C^*_{ab} = \sqrt{a^{*2} + b^{*2}} \qquad (4.4)$$

6) <u>HVS color sensitivity map</u>: Generally, human eyes have different perception sensitivity at different light wavelengths. We use the CIE 1978 spectral sensitivity function values [180] at different wavelengths to locate the image regions for which eyes are more sensitive to the light. To this end, the dominant wavelength for a table of spectral colors (monochromatic colors) is computed first [184]. We assume that the image colors are monochrome (note that conversion from RGB values to light wavelength is not possible for non-monochrome colors, as each color can be represented by many combinations of $R$, $G$, and $B$ values at different wavelengths). Then, for each pixel of the right view image, the closest spectral color and thus its associated dominant wavelength are selected from the available look-up table. Finally, the eye sensitivity at each wavelength is depicted as a map of sensitivity probability.

7) <u>Empirical color saliency map</u>: Several subjective studies have been carried out to test the visual attention saliency of colors using eye-tracking information [181],[185]. In these studies, shapes of different colors are shown to the viewers and based on the eye fixation statistics conclusions are made on the saliency of various colors. Gelasca et al. [185] sorted 12 different colors based on their received visual attention. We use the results of their experiment to build a look-up table for these colors. Then, for each pixel within the right view picture, a saliency probability is assigned based on the closest numerical distance of the RGB values of that pixel and the table entries (mean squared error of the $R$, $G$, and $B$ values).

*4.1.1.4 Texture Map*

Texture and orientation of picture elements are among the most important saliency attributes [50]. To generate a texture saliency map, we utilize the Gabor filters (which are widely used for texture extraction) to create a Gabor energy map at 4 different scales and 8 orientations. The *L2* norm of the Gabor coefficients results in a Gabor energy map. Since image texture is perceived locally at each instance of time, we apply our fovea mask to the Gabor energy map to generate a texture map that contains the edges and texture structure of the image. However, not every edge or image structure is salient. To emphasize salient edges and de-emphasize non-salient texture, we create an edginess map and multiply (element-wise) it by the current texture map. Edginess per unit area is defined by:

$$Edginess_K = \frac{1}{n_K} \sum_{R_K} Edgemap_K \qquad (4.5)$$

where $R_K$ denotes the region $K$, $n_K$ is number of pixels in the $K^{th}$ region, and *Edgemap* for the $K^{th}$ region contains the edges for this segment. Note that edge maps are available as a part of the segmentation algorithm explained previously. Our choice of *Edgemap* ensures that areas with dense edges are assigned with higher saliency probabilities compared to areas with sparse edges.

### *4.1.1.5 Motion Maps*

Due to humans' biological instincts, moving objects always attract human attention. In 3D, there exist three different motion directions: horizontal (*dx*), vertical (*dy*), and perpendicular to the screen (*dz*). Moreover, since it is not clear which one of these directions or which combination of them implies higher impact on the visual saliency, we generate one motion vector for each direction, keep them as separate motion maps, and examine the importance of each attribute using our random forest learning algorithm (Section 4.1.3).

To extract horizontal and vertical motion maps (*Dx* and *Dy* respectively) for right view frames, we incorporate the correlation flow algorithm by Drulea and Nedevschi [186], as this method has shown promising performance on various datasets and is publicly available.

Motion along the *Z* direction particularly exists for 3D video and does not appear in the 2D case. To extract the motion vector along the *Z* direction, we utilize the available depth information (more details on the availability of depth data is presented in Section 4.1.1.6) as follows:

$$Dz = \text{Ref}_{\text{Depth}}(i_x, i_y) - \text{Current}_{\text{Depth}}(i_x + dx, i_y + dy) \tag{4.6}$$

where $Ref_{Depth}(i_x,i_y)$ is the depth value of the ith pixel of a reference frame (or previous frame) with horizontal location of *x* and vertical location of *y*, and $Current_{Depth}(i_x+dx,i_y+dy)$ is the depth value of the ith pixel in the current frame with horizontal and vertical coordinates of $i_x+dx$ and $i_y+dy$, respectively. Note that *dx* and *dy* are calculated using the optical flow algorithm of [186] and $(i_x+dx,i_y+dy)$ in the current frame is the approximate location of $(i_x,i_y)$ in the previous frame.

Once the *Dx*, *Dy*, and *Dz* motion maps are generated, they are normalized to the interval of [0-1]. Then, for each segment of the right view picture, the average motion value is assigned to that segment under the assumption that object motions are homogeneous. In addition to these three maps, several more motion maps are created from the velocity vectors and used in our scheme as follows:

1-4) <u>Velocities</u>: Velocity in different directions is defined as:

$$V_x = (fr-1)dx \; , \; V_y = (fr-1)dy \; , \; V_z = (fr-1)dz \tag{4.7}$$

where $f_r$ is the frame rate of the stereoscopic video. Note that according to our experiments in Chapter 3, frame rate is of particular importance in 3D as motion highly affects the 3D video QoE [41],[187]. The velocity vector magnitude is evaluated as:

$$V = \sqrt{V_x^2 + V_y^2 + V_z^2} \tag{4.8}$$

5) <u>Velocity with emphasize on $V_z$</u>: Due to humans' survival instincts, objects that are on a collision path towards them are treated as a possible threat. Therefore, attention is directed towards them [188]. Inspired by this fact, we modify the velocity vector, emphasizing the velocity in the *Z* direction. Thus, *Dz* is re-defined as:

$$Dz_1 = e^{dz} - 1 \tag{4.9}$$

Using the new *Dz* value, the velocity vector in *Z* direction and then the velocity vector for each segment are re-calculated. The resulting map is referred to as $V_{z\text{-}emphasized}$.

6) <u>Acceleration</u>: Objects with relatively high acceleration are generally considered to be salient. We add a map of relative acceleration to our set of motion saliency maps by computing the acceleration using the following formula:

$$A = \frac{\Delta V}{\Delta t} = (fr-1)[V_{Current \atop frame} - V_{Reference \atop frame}] \qquad (4.10)$$

where velocity is calculated using (4.8) for the current and reference frames.

7) <u>Element of surprise</u>: One of the differences between saliency prediction in images and videos is the possible introduction of an element of surprise in the video, which turns the attention towards itself. To account for the effect of an unusual motion, we emphasize on the saliency of segments with small motion vector occurrence probabilities. To this end, we first find the joint probability density function $p(\underline{d})$ of the motion vectors in three directions. Then, we define a motion histogram map in the range of [0-1] as:

$$M_{probability} = e^{\frac{-p(\underline{d})}{\bar{p}}} \qquad (4.11)$$

*4.1.1.6 Depth Saliency Map*

As mentioned previously, in our method we extract the left-to-right disparity map using the DERS [131] software. However, the same disparity values in a disparity map could correspond to different perceived depths depending on the viewing conditions. Therefore, we use the disparity map to generate a depth map for the right view picture. Fig.4.3 illustrates two similar triangles when an observer is watching a 3D video. Writing down the trigonometry equations gives:

$$\begin{cases} \dfrac{disparity\ (W/R_W)}{L_{eyes}} = \dfrac{x}{Depth} \\ x = Z_{observer} - Depth \end{cases} \Rightarrow Depth = \dfrac{Z_{observer}}{1 + \dfrac{disparity \times W}{L_{eyes} \times R_W}} \qquad (4.12)$$

where $Z_{observer}$ represents the distance of the viewer's eyes to 3D screen (183 [*cm*] in our experiments), $L_{eyes}$ is the inter-ocular distance between the two eyes (on average 6.3 [*cm*] for humans), and $W$ and $R_W$ are the horizontal width (in [*cm*]) and resolution (in pixels) of the display screen, respectively. Note that this method of disparity to depth conversion results in read-world depth values (in [*cm*]) and has been similarly used by other colleagues [68],[69]. Also, note that our saliency prediction mechanism requires a disparity which does not necessarily have to be generated using DERS. Any disparity detection algorithm can be used for disparity map generation.

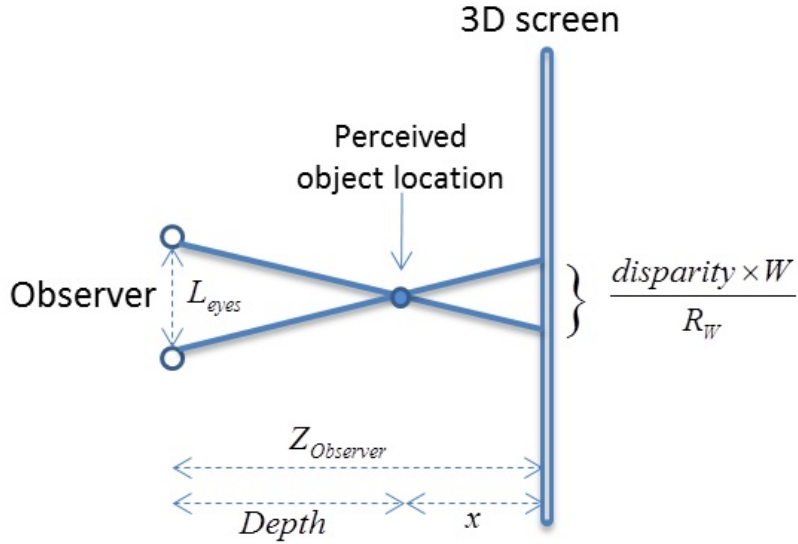
**Figure 4.3 Disparity to depth conversion scheme viewed from above**

A depth map can be incorporated to create a depth saliency map in which closer objects are assigned higher saliency values. To create such a map, the inverse of depth values is linearly mapped to [0-1]. The resulting map, however, does not demonstrate the saliency perfectly since not every close object is salient (see Fig. 4.4 for a counter example). An object that stands out due to its depth is likely to have depth values that abruptly change compared to its neighborhood. The depth value of the object itself is often smooth and its depth contrast is low. Visual attention is directed to objects with lower inner disparity contrast [189] but vary abruptly outside their depth value [70]. To account for this fact, a depth abruptness mask is created in which the mask contains high values when depth changes abruptly (and contains values close to zero in case depth is changing smoothly).

To create the depth abruptness mask, four points are selected around each segment (See Fig. 4.5 for an illustration). We observe that $P_1$ and $P_3$ have the same horizontal coordinates as the center of mass (centroid) of the segment, while $P_2$ and $P_4$ share the same vertical coordinate with the centroid. Suppose $d_1$, $d_2$, $d_3$, and $d_4$ are the highest horizontal and vertical distances from the center of mass of the segment to any point on the segment perimeter towards up, right, down, and left side directions, respectively. The four selected points are located at $P_1$:($\bar{C}_x, \bar{C}_y - d_1 - 0.01 \times H$), $P_2$:($\bar{C}_x + d_2 + 0.01 \times W, \bar{C}_y$), $P_3$:($\bar{C}_x, \bar{C}_y + d_3 + 0.01 \times H$), and $P_4$:($\bar{C}_x - d_4 - 0.01 \times W, \bar{C}_y$), where $H$ and $W$ are the height and the width of the display screen. These four points are likely to fall within neighbor segments and therefore, can be used to evaluate the depth change rate. Note that studies have shown that objects' sizes affect their visual saliency [190].

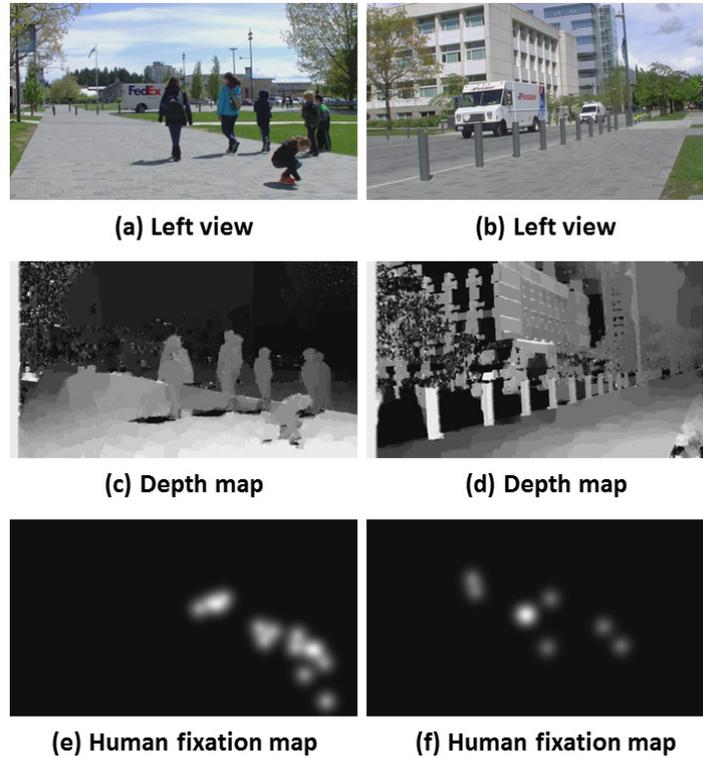

(a) Left view          (b) Left view

(c) Depth map         (d) Depth map

(e) Human fixation map     (f) Human fixation map

**Figure 4.4 Closer objects are not necessarily salient. "Ground" is not considered salient in these examples**

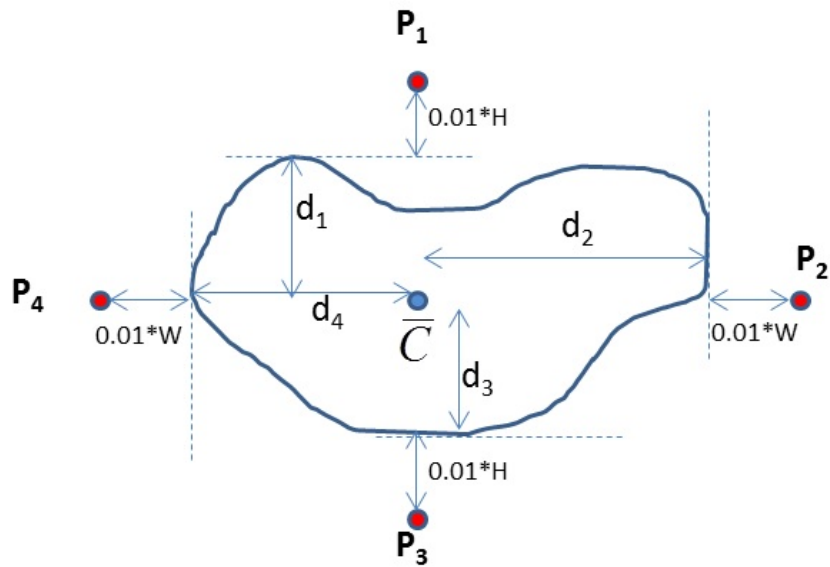

**Figure 4.5 Creating a depth abruptness mask: An example segment with four selected points around it**

Walther et al. used a threshold of 5% for the minimum suitable area of a salient region. Our experiment settings (are explained in details in 4.2) suggest using an area threshold of 1% due to the utilized display size and viewers' distance. As a result, we add 1% of the display height to $d_1$ and $d_3$ and 1% of the display width to $d_2$ and $d_4$ to reach the four points. Next, for each segment and its associated set of selected points we find the weighted average of depth differences between the points and the centroid, and assign the resulting average value to that segment. Continuing this process for all of the segments generates a depth difference map, $Diff_{Depth}$. The weights in finding the average difference are equal to the inverse of the distances between each $P_i$ point and the centroid. The depth abruptness for each segment is defined as follows:

$$Abruptness_K = \begin{cases} Max(Diff_{Depth}) & if \quad Diff^K_{Depth} > \overline{Diff_{Depth}} \\ Diff^K_{Depth} & if \quad Diff^K_{Depth} \leq \overline{Diff_{Depth}} \end{cases} \qquad (4.13)$$

where $Diff^K_{Depth}$ represents the depth difference associated to the $K^{th}$ segment. In our formulation, we interpret a local depth difference of higher than average as an abrupt change and a depth difference of lower than average is referred to as a relatively smooth change. The final depth abruptness mask is normalized to the interval of [0-1]. Note that since synthetic depth maps can always contain artifacts, in our implementation we choose multiple (three in the present embodiment) close points instead of only one point at each direction to increase the robustness of the process against the depth map artifacts. These multiple points are selected very close (a few pixels away) to each $P_i$ point. Also, note that instead of four neighboring points, one can choose eight neighboring points (with $45^O$ rotations), or a scanning line in each of the four directions to measure the maximum depth differences. Our simulation experiments, however, verified that the additional accuracy is negligible compared to the added computational complexity.

We perform a slight smoothing (using a simple Gaussian filter) on the depth image to prevent the imperfections that can be caused by depth map artifacts. To create the final depth saliency map, we multiply the depth abruptness mask (element-wise) by the current segmented depth map. The depth saliency map is linearly normalized to [0-1] at the end.

*4.1.1.7   Fovea Masking*

When focusing on a specific region of an image, HVS perceives the neighborhood around that region very sharply but as the distance from the center of attention is increased the rest of the picture seems blurry to the human eye. This is due to the photoreceptor concentration density in fovea, which decreases from the center of fovea rapidly [191]. Cone concentration is about 150 000 cells/mm2 at the foveal center and decreases to 6000 *cells/mm$^2$* at a distance of 1.5 *mm* to the fovea [191]. Rod density peaks at 150 000 *cells/mm$^2$* at a ring-like area at a distance of about 3-5 *mm* from the foveola. It decreases towards the retinal periphery to about 30 000 *cells/mm$^2$* [191]. Fig. 4.6 demonstrates an example of this photoreceptor distribution phenomenon. As mentioned previously in this section, we incorporate a fovea-masking-based center-surround operation in generating some of the proposed feature maps. Inspired by the photoreceptor concentration in the fovea, in our implementation, this mask is a circular disk in which the value of each element is proportional to the photoreceptor density of the corresponding location. Radius of the mask is defined based on the size of the display and distance of the viewer from the display. Suppose *α* is half of the angle of the viewer's eye at the highest visual acuity. The range of *2α* is between 0.5° and 2° [136]. Sharpness of vision drops off quickly beyond this range. The mask radius is approximately defined by:

$$L = Z_{observer} \tan(\alpha) \; [cm] = \frac{Z_{observer} \tan(\alpha) R_H}{H} \; [pixel] \qquad (4.14)$$

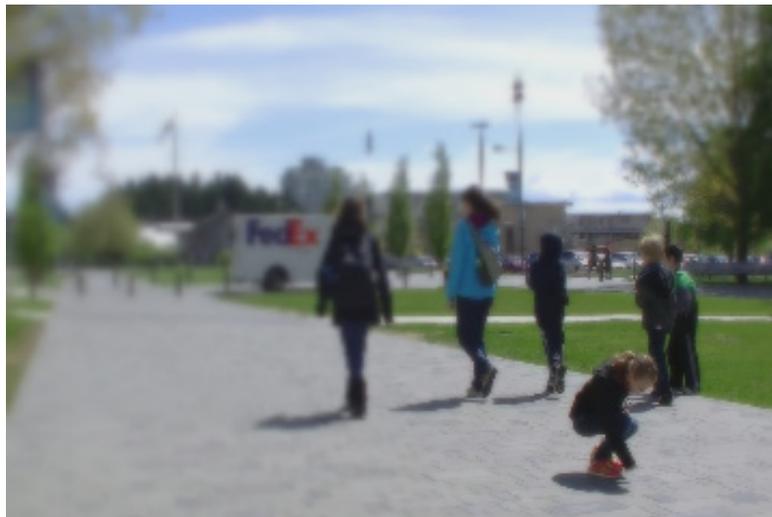

**Figure 4.6 Foveation effect: due to photoreceptor concentration of fovea, sharpness of vision drops quickly as the distance increases from the center of focus**

where $H$ and $R_H$ are the vertical height and resolution of the display, and $Z_{observer}$ is the distance of the viewer to display. In our implementation (explained in 4.2), we choose an angle of $\alpha=1^o$, HD resolution video at 1080×1920, viewing distance of 183 [$cm$], and display height of 57.3 [$cm$]. The resulting mask radius for this setup is 60 pixels.

*4.1.1.8 Size, Compactness, and Sparsity of Salient Regions*

As already mentioned, psycho-visual experiments have revealed that object size affects the saliency [190]. In our method, segments with size of less than 1% of the resolution in each direction are not considered as salient.

In addition to size of the regions, compactness of each region affects its visual saliency. Due to the nature of the eye-tracking studies (and similarly in reality), human attention is directed towards compact objects [71]. This makes more sense considering the fact in video saliency prediction there are only a few fixations per frame, which mostly are associated with compact objects. To account for the effect of object compactness, the moment of inertia for each segment is chosen as a compactness measure [192]. A map of compactness is then created in which each segment is associated with its compactness value. This map will be applied as multiplicative mask to all feature maps.

Another important factor regarding the saliency of objects is that there are only very few salient objects in each scene. Given the allocated time for human viewers to view each frame of a video and hardware capability to record their eye fixations data, there are only a couple of fixations per frame. Therefore, a proper saliency prediction algorithm should extract only a few highly probable salient regions. To account for the sparsity of the salient regions, we propose a mechanism that puts emphasis on local maxima points in the available feature maps. Assuming a feature map is scaled to [0-1], we seek a convex function that projects this feature map to another map with the desired properties. One candidate for such operation is:

$$\Phi(F) = e^F - m_F \tag{4.15}$$

where $F$ denotes a feature map and $m_F$ represents the average of $F$. Since $F$ values are within [0-1] then the values of $\Phi$ fall within [$1-m_F, e-m_F$]. For the $F$ values equal to $m_F$, $\Phi$ would approximately become 1 (normally, there are only a few salient regions in each frame. Therefore, for a sparse saliency map, $m_F$ is very small since it is equal to the average of saliency probabilities). The *F* values above the average are subject to higher increments compared to *F* values below the average. Once the resulting *Φ-map* is rescaled to [0-1], the local maxima points are relatively more amplified than the points with feature values below average. Rescaling is performed linearly as follows:

$$\Psi = \frac{\Phi - \Phi_{min}}{\Phi_{max} - \Phi_{min}} \quad (4.16)$$

*4.1.1.9 3D Visual Discomfort*

When watching stereoscopic content, several reasons may degrade the 3D quality of experience. Assuming that the 3D display does not have any crosstalk and the data is captured properly without any unintended parallax, the main source of discomfort is caused by the vergence-accommodation conflict [193]. When viewing stereoscopic 3D content, there is a comfort zone for the content within which the objects should appear. Any region perceived outside of the comfort zone degrades the 3D QoE, as eye muscles try to focus on the display screen to perceive a sharper image while they also try to converge outside of the display screen to avoid seeing a double image. This decoupling between vergence and accommodation results in fatigue and degrades the QoE. Studies suggest using a maximum threshold for 3D content disparity. In particular as a rule-of-thumb, it is widely accepted to use maximum allowed of $1^o$ disparity [193]. Beyond this threshold QoE drops rapidly.

To account for 3D discomfort level, we create a discomfort penalizing filter and apply (element-wise multiplication) it to our depth feature map as it represents the 3D saliency attributes. The discomfort penalizing mask elements are assigned to 1 when a segment falls within the comfort zone and to a penalty value when the segment is outside of the comfort zone. Several quantitative measures of 3D visual discomfort are proposed so far. We don't limit our visual attention model to any particular discomfort measurement as it is not yet clear how exactly discomfort can be measured. The choice of the discomfort metric does not affect our proposed penalizing scheme. However, for the sake of illustration, we choose a simple discomfort metric, which is based on subjective visual experiments presented in [193]. By averaging the 3D QoE values of [193] for various types of content at different resolutions and disparity ranges, we derive the following rough estimate for the penalizing mask:

$$\begin{array}{c} Discomfort \\ Mask \end{array} = \begin{cases} 1 & d \leq 60^{\text{minutes arc}} \\ 1.36 - 0.006 \times disparity & d > 60^{\text{minutes arc}} \end{cases} \quad (4.17)$$

In our experiment setup, $1^\text{o}$ disparity corresponds to 60 pixels.

### 4.1.2 High-Level Saliency Features

In addition to low-level bottom-up saliency features, we add several high-level top-down features to our model. The use of high-level features helps to improve the accuracy of saliency prediction. The following high level features are considered in our method: face, person, vehicle, animal, text, and horizon. For each feature, a saliency map (using a bounding box around the detected salient region) is created and used in the training of the proposed visual attention model.

When a human appears in a video shot, the observer's attention is naturally drawn to the person. In order to detect faces and appearance of people in a scene, we use the Viola-Jones algorithm [194] and Felzenszwalb's method [195] (trained on the PASCAL VOC 2008 dataset [196]), respectively. Felzenszwalb's method [195] is also used to detect the presence of bicycles, motorbikes, airplanes, boats, buses, cars, and/or trains. The same method is incorporated to detect animals including birds, cats, cows, dogs, horses, and/or sheep.

Image areas containing text also attract the human attention. To account for the appearance of text we use the Tesseract OCR (Optical Character Recognition) engine [197]. In addition, Gist descriptor is used to detect a horizon in the scene [198].

Generation of the bottom-up and top-down feature maps results in 24 maps which are then fused to a single saliency map using random forests regression. Fig. 4.7 demonstrates an example of generated feature maps using our method.

### 4.1.3 Feature Map Fusion based on Random Forests

Since it is not clear how map fusion happens in the brain, computational models of visual attention use different approaches. On one hand, some approaches integrate the features internally and do not produce several separate feature maps [54]-[56],[65],[67].

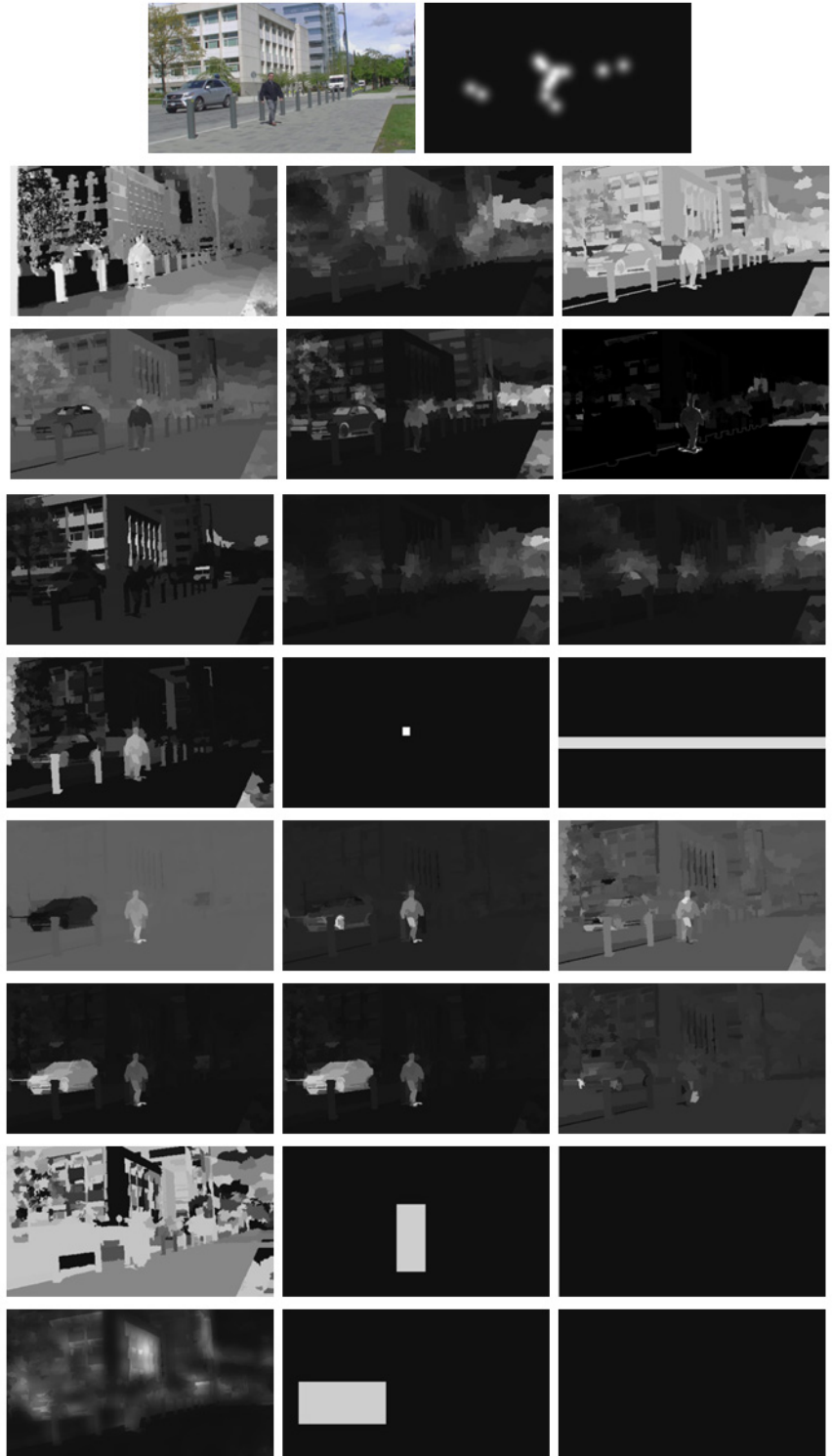

**Figure 4.7 Generated feature maps using the proposed method: Images from left to right in a raster-scan order are: Right view picture, human fixation map, disparity map, brightness feature, color histogram feature, color contrast feature of a$^*$, color contrast feature of b$^*$, color warmth, color saturation, HVS color sensitivity feature, empirical color saliency feature, depth feature, face, horizon, motion *dx*, motion *dy*, motion *dz*, velocity magnitude, velocity magnitude with emphasize on *Z* direction, acceleration, element of surprise, person, text, texture, vehicle, animals**

One the other hand, other methods incorporate linear [51],[52],[60][62],[64], SVM (support vector machine) [53], or deep-networks-based map fusion schemes [199]. In the case of 3D visual attention modeling, map fusion is mostly performed by linear combination [69]-[71].

In our approach, we utilize random forest regression for fusing the different feature maps generated by our method. The motivation behind this tool selection is explained below. Random Forest (RF) methods construct a collection of decision trees using random selection of input features samples [200]. Generally, each decision tree might not perform well over unseen data. However, the ensemble of the trees usually generalizes well for test data. This follows the structure of HVS (and also the proposed feature extraction process) in incorporating several visual features.

Each feature on its own may not predict the saliency well, but integrating various features provides a much more accurate prediction. One of advantages of RF regression techniques is that they do not require extensive parameter tuning since they intuitively divide data over the trees based on how well they classify the samples. Moreover, the importance of each individual feature can be evaluated using the out-of-bag-errors once a model is trained. This makes it possible for our proposed method to find an assessment of how important each feature is in the saliency prediction, so that an appropriate decision can be made on the trade-off between number of features (computational complexity) and prediction accuracy. For a set of training videos (details in 4.2), we extract different feature maps and use them to train a RF regression model. Bagging is applied to tree learners to construct the decision trees. The resulting RF model is later used to generate a saliency map for unseen test features. In addition, the importance of each feature helps to decide what number of features to use.

## 4.2 Benchmark 3D Eye-Tracking Dataset for Visual Saliency Prediction on Stereoscopic 3D Video

It is widely acceptable that the existence of a large-scale eye-tracking dataset for stereoscopic 3D videos will accelerate the development of precise 3D VAMs. This section introduces our publicly available eye-tracking dataset of stereoscopic videos, which is used to verify the performance of our proposed 3D VAM and also aims at facilitating the research community's efforts to compare and validate new 3D saliency prediction algorithms. This section provides information about our eye-tracking dataset and the subjective experiments.

*4.2.1 Capturing the Stereoscopic 3D Videos*

Sixty-one (61) indoor and outdoor scenes are selected to produce our 3D video dataset for the eye-tracking experiment. A stereoscopic 3D camera (JVC Everio) was used to capture the sequences. The distance between the two lenses of the camera is 6.5 [*cm*], which is the same as the average inter-ocular distance for humans. The camera was fixed on a tripod at the time of capturing so there were no camera movements. No zooming was used for the camera lenses. Special attention was put during the capturing on avoiding possible window violations. Note that window violation is a common artifact for 3D video acquisition and occurs when parts of an object fall outside the capturing frame. As a consequence, part of the object is perceived inside and screen and other parts appear outside the 3D screen. The two contradictory depth cues result in 3D visual discomfort and reduce the 3D QoE.

Each view is captured in full HD resolution of 1920×1080 at the frame rate of 30 frames per second (fps). Moreover, the length of each video is approximately 10 seconds. The video sequences contain a wide range of intensity, motion, depth, and texture density. Scenes are selected in a way that the captured dataset covers almost all different possible combinations of these parameters. In other words, there are multiple video sequences captured for each combination of four parameters, intensity, motion, depth, and texture density, with two levels of each (low and high), resulting in 16 different scenarios. As a result, our dataset is not biased towards a specific scenario. Fig. 4.8 provides a statistical overview of the video sequences. In this figure, in order to measure the motion and texture density, we utilize the temporal and spatial information (TI and SI, respectively) metrics. We follow the recommendation of ITU [143] for the calculation of SI and TI measures. In addition, we provide the histograms of the average intensity values for the video dataset as well as the average disparity bracket of the scenes. Disparity bracket is defined as the range of the disparity (in pixels) for each stereopair (i.e., horizontal coordinate of the closest point subtracted by the horizontal coordinate of the farthest point). Note that since some of the VAMs in the literature utilize machine learning techniques for saliency prediction, we split the dataset to the training and validation sets so that the training part can be used for training different VAMs and validation is used for validation of the models.

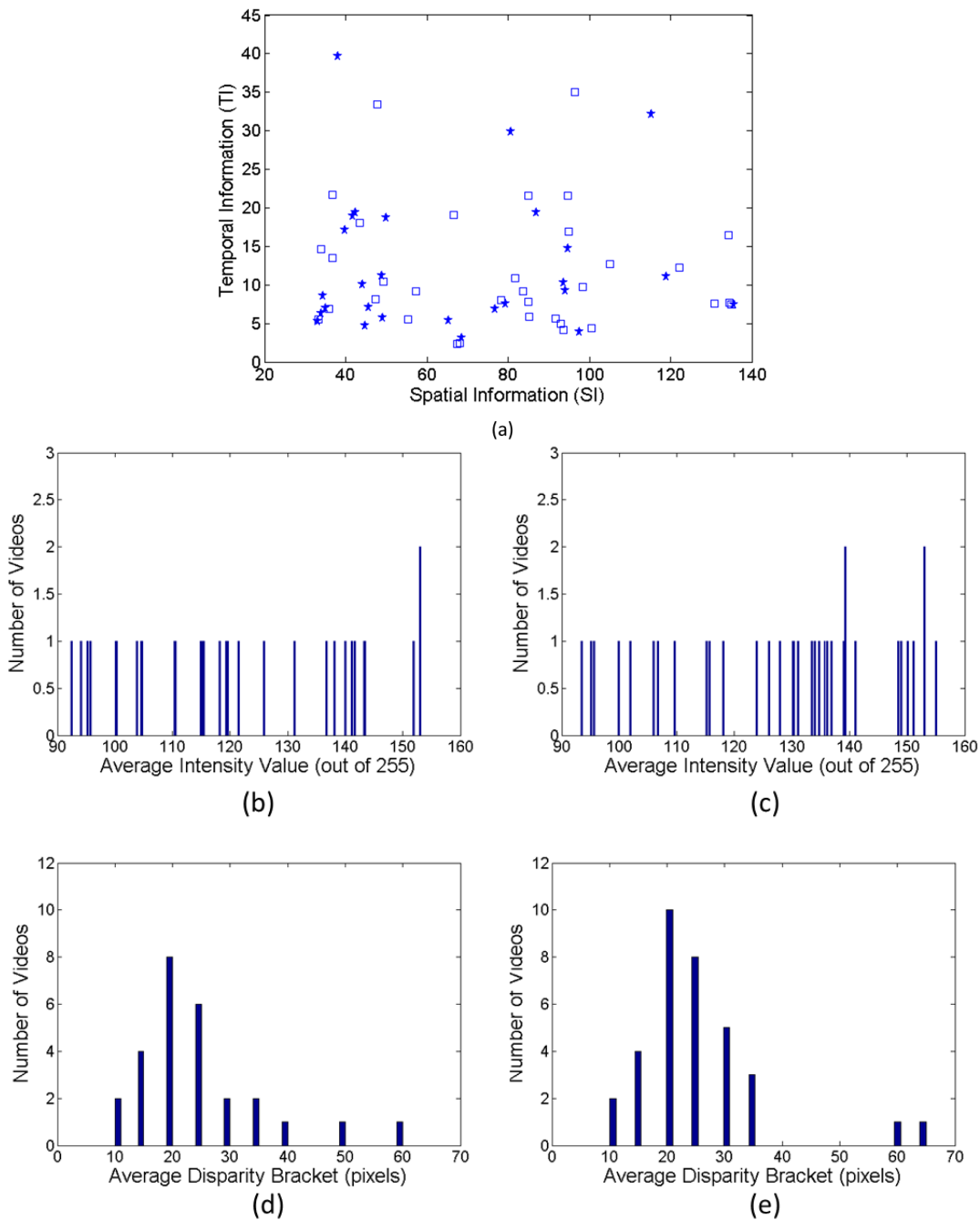

**Figure 4.8 Statistics of the stereoscopic video dataset: Temporal and spatial information measures (a) stars correspond to the training videos and squares refer to validation videos, histograms of the average intensity values for the train set (b) and validation set (c), and histograms of the average disparity bracket for the train set (d) and validation set (e)**

We chose 24 videos for training (~ 40%) and 37 sequences for validation (~ 60%). We tried to distribute the videos in the two sets fairly, according to their TI values, SI values, intensity histograms, and depth bracket histograms. Statistical properties are provided for both the training and validation sets in Fig. 4.8. It is observed from this figure that the training and validation sets roughly demonstrate similar statistical properties and, therefore, an unbiased generalization is provided for learning methods.

Moreover, in order to have a balance with respect to the high-level saliency attributes, videos are captured such that humans appear in roughly half of the scenes while in the other half there is no human appearance. Similarly, it is ensured that approximately half of the scenes contain vehicles. A snapshot of the right view of some of the videos in our database is demonstrated in Fig. 4.9. Since the videos are captured in two left and right views, disparity estimation algorithms should be incorporated to extract disparity.

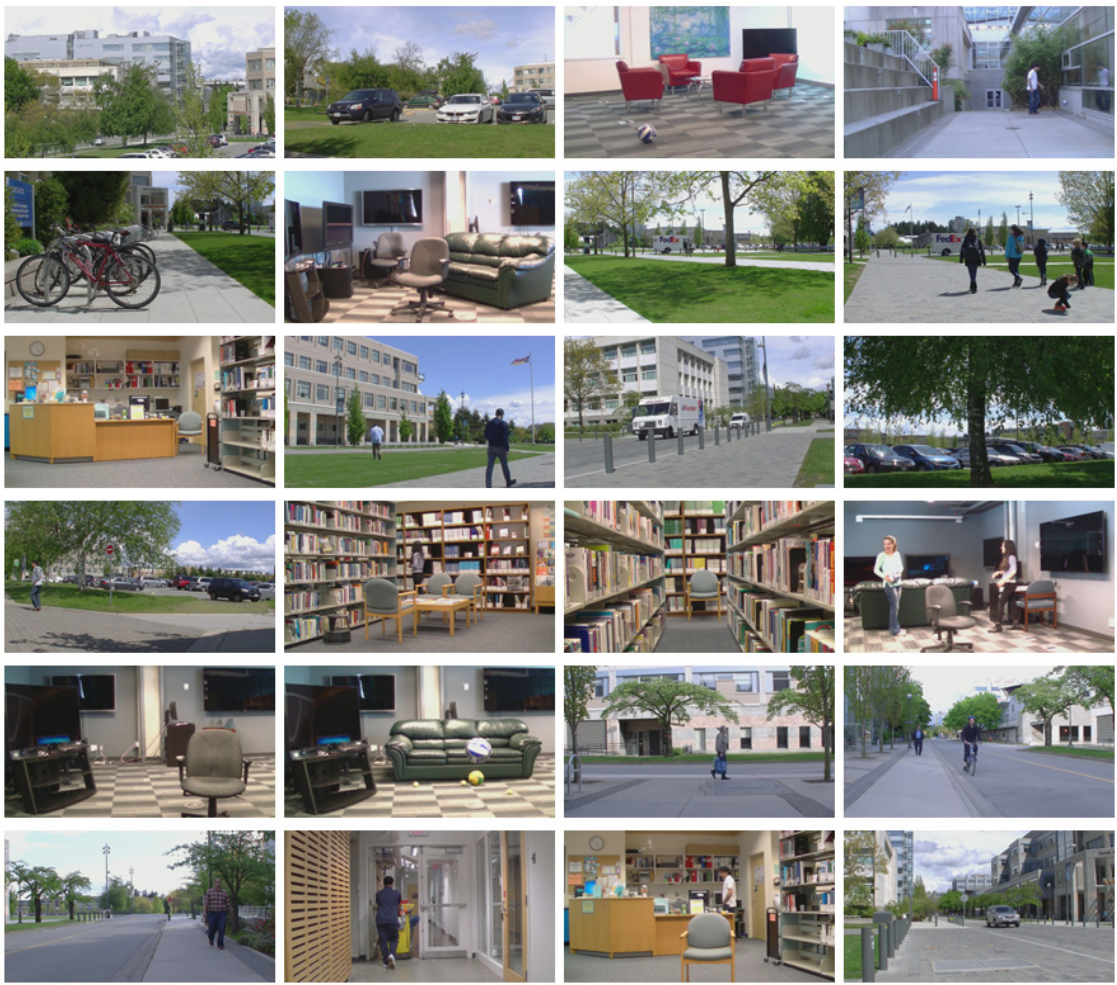

**Figure 4.9 Snapshots of the videos in our database**

*4.2.2 Post-Processing the Captured Videos*

Using a 3D camera with parallel lenses results in negative parallax, which corresponds to objects popping out of the 3D display screen. Experimental studies have shown that viewers distinctly prefer to perceive the objects of the interest on the screen and other objects inside/outside of the screen [145]. At the time of capturing, we used the manual mode of disparity adjustment so that the camera doesn't automatically change the disparity. Therefore, it is required to bring the objects to the 3D comfort zone to provide the viewers with a high 3D quality of experience. To this end, we perform disparity correction to the captured sequences. Disparity correction is achieved by cropping the left side of the left view and the right side of the right view. The amount of cropping is selected based on the disparity of the object of interest in a way that the disparity corrected stereopair places the object of interest on the screen. For each video sequence the object of the interest is identified through a subjective user study. More details regarding this disparity correction methodology can be found in [187].

*4.2.3 Eye-Tracking Experiments*

In order to record the eye movements of the viewers we use a SensoMotoric Instrument (SMI) iView X RED system [201]. The sampling rate of the eye-tracker is 250 Hz, i.e., it has the capability of tracing the eye movements 250 times in each second. Moreover, the accuracy is up to $0.04 \pm 0.03^{o}$ of visual angle. A 46" Hyundai (S465D) 3D TV was used for the presentation of the test material. The TV utilizes passive glasses in the 3D mode. The resolution of the display is full HD (1920×1080) in 2D mode and, thus, there is no need for up/down sampling the video data. The video sequences were displayed in interlaced format so that the eye-tracker grabs displayed frames through an HDMI cable. This results in losing half of vertical resolution (which is unavoidable when using a 3D TV). However, it avoids biasing towards one of the views and presents the actual perceived scene. Peak luminance of the LCD screen is 120 cd/m$^2$. Color temperature was fixed at 6500K, which is recommended by MPEG for subjective evaluation of the proposals submitted in response to the 3D Video Coding Call for Proposals [153]. Subjective experiments were performed in a room that is specifically designed for visual experimental studies. The wall behind the display was illuminated using a uniform light source, with the light level of 5% of the screen peak luminance. In order for the eye-tracker to accurately follow the eye movements, it was placed in between the viewers and the 3D display. The distance and height of the eye-tracker is based on the recommendations of the SMI software. Fig. 4.10 sketches the test setup and placements of the devices. Fig. 4.11 shows the real test environment.

Twenty-four participants attended our test sessions (13 male and 11 female). Prior to the actual eye-tracking experiments, all the subjects were screened by pre-tests to assess their visual acuity using the Snellen chart, color blindness using the Ishihara graphs, and stereovision using the Randot test.

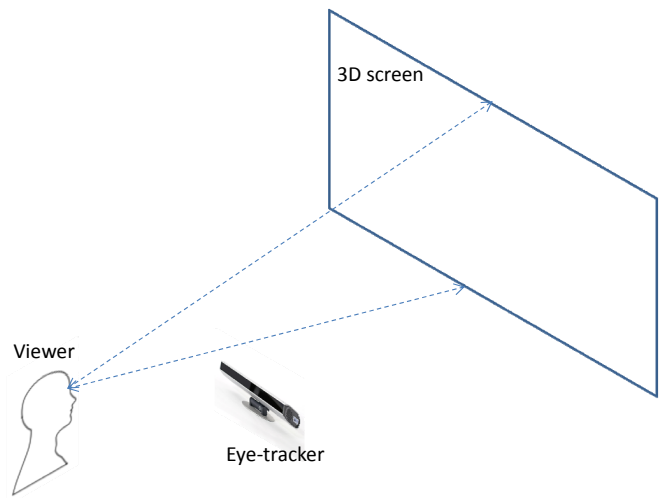

**Figure 4.10 Eye-tracking experiment setup**

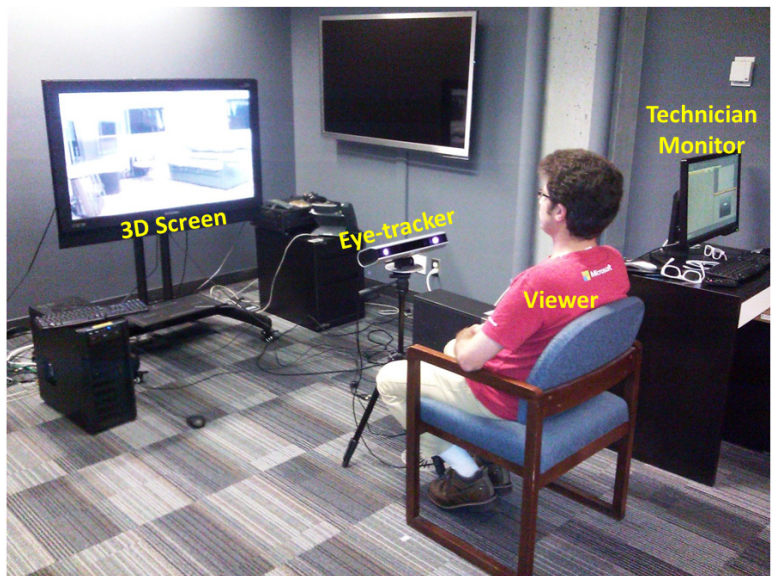

**Figure 4.11 Test environment**

Users who didn't pass any of the three pre-tests were not entered to the eye-tracking experiment. Users were selected among university students who were naïve to the purpose of the test and it was ensured that they do not have prior knowledge about the research topic. Subjects participated in the eye-tracking experiment one at a time. The test involved with a free-viewing task in which viewers freely watched the sequences and their eye movements were recorded. In order to study the statistical differences between the 2D and 3D fixations and saccades, for each participant we performed a test with the 3D sequences and another one with only one of the views (right view). To ensure that the subjects were not biased towards the order of the presentation material, the order of the 2D and 3D videos was switched for every subject so that counter balance was achieved. Note that prior to each test session a calibration was performed for each user to ensure the accurate recording of the gaze points. The calibration was performed a few times at the beginning and throughout the experiment so that the eye-tracker doesn't lose track of the eye movements. For calibration, the internal SMI software was used to show a circular dot on the screen and viewer was asked to follow its movement. The circular dot moved in random fashion to the corners of the screen to avoid the center-bias phenomenon for the fixations [202]. It is also worth noting that the presence of passive glasses does not affect the performance of the eye-tracker as infrared technology is used. That being said, we also performed pre-tests to ensure that the glasses do not interfere with tracking.

*4.2.4 Data Collection and Fixation Map Generation*

After performing the eye-tracking experiments, the gaze points are collected and are ready to be processed. In the literature, the gaze points are usually converted to a fixation density map (or sometimes heat map) to provide a 2D representation of most salient points. This map basically shows the likelihood of an object to draw attention.

Due to the possible inaccuracies of the eye-trackers and to account for the drop in vision sharpness as the distance increases from the fovea center point, the gaze points are usually filtered using a Gaussian kernel to create the fixation density maps [53][68][69],[76],[203],[204]. Here, instead of using a Gaussian kernel for creating the fixation density maps, we incorporate a "fovea masking" kernel, which is created based on the photoreceptor concentration density in the fovea. Due to photoreceptor distribution in the human eyes, objects at the center of focus are projected to the center of fovea and therefore appear to be the sharpest. With increasing the eccentricity from the fovea, vision sharpness drops rapidly [136],[191],[205],[206]. To mimic the foveation effect, we create a circular shape mask with maximum value (of 1) at the center and lower values as the radius increases. We adopt the cell count distributions measured in physiological studies [191],[205],[206] and resize this kernel to the desired size. The radius size $L$ of the mask is selected based on the 3D display size and the distance of the viewer from the display (Fig. 4.12) according to equation (4.14). In our setup, the distance of the viewer from the display, $Z_{observer}$, is set at 183 [$cm$] and the display height, $H$, is 57.3 [$cm$], which corresponds to a radius of 60 pixels for the resulting fovea masking kernel.

After applying the fovea-masking kernel to the gazed points, a fixation density map is generated for each frame of the videos. A similar process is repeated for both 2D and 3D videos to generate the fixation maps for each user. User fixation maps are then averaged to create a fixation map video for each sequence. Fig. 4.13 shows the average of the fixation maps for the 2D and 3D sequences and all users. It is observed from Fig. 4.13 that the fixations are distributed all around the frames and therefore there is no significant center-bias observed. This shows that the scenes are captured in a way that viewers are not directed only towards the center of the screen. It also makes the comparison of different VAMs fairer.

The created fixation maps are used as ground truth saliency maps for validating the performances of visual attention models. Our eye-tracking dataset of stereoscopic videos is made publicly available and can be accessed at: http://ece.ubc.ca/~dehkordi/saliency.html. Moreover, we created an online benchmark dataset and evaluated the performance of the existing 2D and 3D VAMs on our 3D eye-tracking dataset. In our online platform, we provide the comparison of the performances of many saliency detection algorithms and make it possible to add new models. We provide the ground truth fixation maps for the training part of the dataset and ask interested contributors to send us the implementation of their VAM or their generated saliency maps so that we conduct performance evaluations using their input and our validation part of the dataset. Then, the results will be added to the benchmark.

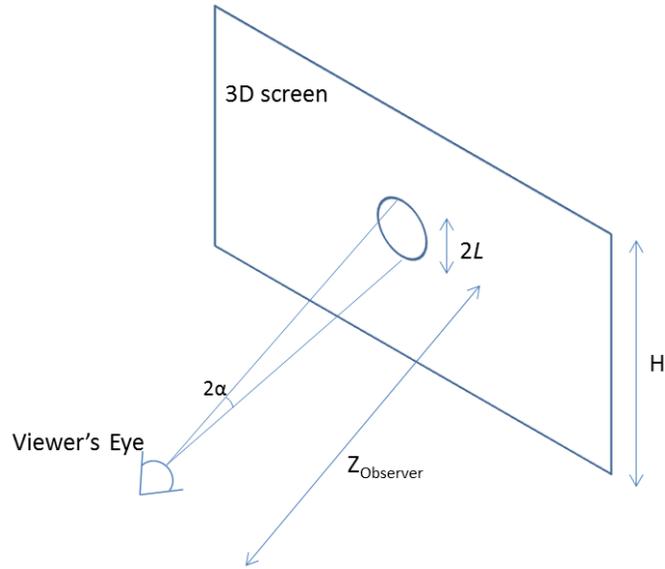

**Figure 4.12 Fovea masking scheme**

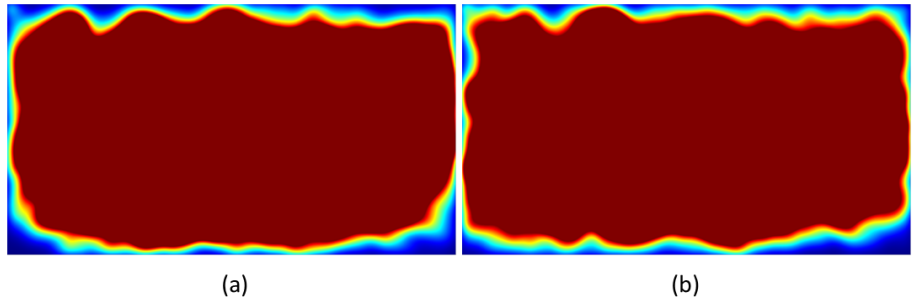

(a)                  (b)
**Figure 4.13 Average fixation maps: 2D videos (a) and 3D videos (b)**

## 4.3 Performance Evaluations

This section elaborates on the results of our experiments and compares the performance of the proposed saliency prediction method with that of the state-of-the-art.

### 4.3.1 Metrics of Performance

The Area Under the Receiver Operating Characteristics (ROC) Curve (commonly referred to as AUC) is a basic measure of performance in our evaluations [207]. In addition to AUC, several other metrics are incorporated for evaluating different VAMs. In particular, the shuffled AUC [202],[208], Normalized Scanpath Saliency (NSS) [202], Pearson correlation ratio (PCC) [50], Earth Mover's Distance (EMD) [209],[210], Kullback–Leibler Divergence (KLD) [211], and a Similarity score proposed by Judd et al. [83] were used in our evaluations. The values of the above metrics are calculated for each frame of our validation video dataset and then averaged over all the frames to provide one metric score for each video. Each of these metrics is explained in this subsection.

1) <u>AUC</u>: AUC has been frequently used for evaluating the performance of saliency prediction models [207],[212][215]. AUC is computed between a saliency map and a ground truth fixation map and quantifies the ability of a saliency map in classifying the fixated and non-fixated locations. The ROC curve creates a binary map of fixations by setting a threshold on the saliency map. Then, each point is assigned to one of two categories: positive set (fixation points) and negative set (a random draw from the non-fixated points). A point is labeled as True Positive (TP) if it belongs to the positive set and is detected as salient in the thresholded saliency map. Similarly, False Positives (FPs) are defined as the points which belong to the negative set but were detected as salient in the thresholded saliency map. Consequently, True Positive Rate (TPR) and False Positive Rate (FPR) are defined as the number of the TP and FP points over the entire number of pixels within each set. The ROC curve is sketched as TPR against the FPR for different values of the threshold in [0-1]. AUC achieves values between 0.5 and 1. A higher value of AUC denotes a more accurate saliency map. A value of unity indicates 100% accuracy in saliency prediction and a value of 0.5 indicates that the saliency map predicts the saliency points no better than by chance.

2) <u>Shuffled AUC</u>: Due to the presence of center-bias when watching the videos, it is expected that a high density of the fixation points are located around the center of each frame. Experimental studies also verify this argument [216]. In the case of 3D video, the center-bias effect is expected to be even more significant, since if the 3D content contains window violations or any other artifact that results in visual discomfort around the borders, viewers tend not to look at the edges of the display and focus around its center. The way that AUC calculates the accuracy of a saliency map is affected by the center-bias phenomenon [67][202],[216]. To address this disadvantage, the definition of the "negative set" in classic AUC is modified to account for the center-bias problem. Negative points for the current frame are randomly selected from the fixated points of other scenes in the same dataset (as opposed to regular AUC which chooses a random draw of non-fixated points of the same frame). This definition of AUC is known as Shuffled AUC (sAUC) [67],[202],[216]. sAUC achieves higher reliability compared to classic AUC and reduces the effect of the center bias. We use a publicly available implementation of sAUC provided by Zhang and Sclaroff [52].

3) <u>NSS</u>: Normalized Scanpath Saliency (NSS) is a measure of how well a saliency map can predict the fixations along the scanpath of the fixations. It is measured as the average of saliency map values along the viewer's scanpath [202]. A saliency value of greater than unity indicates that the saliency map detects saliency much better along the scanpath compared to the other locations. A negative NSS value, however, indicates that the saliency predictions are no better than identifying the salient locations just by chance.

4) <u>PCC</u>: The Pearson Correlation Coefficient (PCC or PLCC) measures the correlation between a saliency map and a fixation map by treating them as two random variables [50]. It is a linear measure of correlation that ranges from -1 to 1, with values close to -1 and 1 indicating a perfect linear correlation (thus a more accurate saliency map) and values near 0 indicating no correlation between a saliency map and a ground truth fixation map (thus very low accuracy).

5) <u>EMD</u>: It is argued that the AUC metric does not take into account the distance between the points in a saliency map and the corresponding points in the fixation map. The fact that a point in the saliency map is a hit/miss (TP or FP) is the only information taken into account for AUC calculations. However, the distance between these points can also be a measure of how close a saliency map is to a fixation map [217]. The Earth Mover's Distance (EMD) [209] is often used for taking into account the distance between two probability distribution functions. In the context of visual attention modeling, EMD measures the cost of converting a saliency map to its corresponding human fixation map. Here, the cost for each window means the difference in the saliency probabilities weighted by the distance between the windows. EMD is a distance measure, so an EMD value of 0 means 100% accuracy of saliency detection, while higher EMD values correspond to lower accuracies. A fast implementation of EMD prepared by Pele and Werman [210] is used in our saliency benchmark.

6) <u>KLD</u>: The Kullback-Leibler Divergence (KLD) (a.k.a information divergence) evaluates the saliency prediction accuracy in an information theoretic context. It models the saliency and fixation maps as two probability distributions and measures how much information is lost when one is used to approximate the other [211]. Similar to EMD, KLD is also a distance measure and achieves non-negative values. Lower KLD values indicate better saliency prediction accuracies.

7) SIM_measure: This Similarity Metric (SIM) was originally used by Judd et al. [83] to evaluate the similarity between a saliency map and its corresponding fixation map. SIM measures the summation of the minimum values of two probability distribution functions (PDFs) evaluated at different points of the distributions. The distributions are scaled so that they sum up to unity. Therefore, SIM takes values in the interval of 0 and 1, 1 indicating that a saliency map perfectly matches the corresponding fixation map, while 0 showing no similarity between the two.

*4.3.2 Baselines*

When comparing the performance of different visual attention models over our stereoscopic video dataset, we use four different baselines as reference models to be compared against each VAM:

1) Chance: a map of random values between 0 and 1 is created to indicate the chance map.

2) Center: This model is particularly chosen to measure how much center-bias exists in the eye-tracking data. A Gaussian kernel with the size of the frame (i.e., 1920×1080 for our dataset) and the optimum standard deviation (*std*) is chosen to represent the center map (the kernel is normalized to be 1 at the center and lower values around the center). The optimal value of the standard deviation is chosen by measuring the AUC between the resulting center map and the fixation maps of the training set for many different *std* values. The value of *std* is exhaustively swept between 0.1 to 1000 (using MATLAB's fspecial function); the optimal value of *std*=300 resulted in AUC of 0.6064 over the validation set.

3) One human's opinion: since people look at different parts of videos, the fixation distributions are different for different viewers. As a result, the fixation map of one viewer can only partially reflect the true average human fixation map. As a baseline saliency model, we measure how well one human observer can predict the fixation map of the other observers. In other words, the fixation map of a random participant is used as a saliency map and its performance in predicting the fixation maps of other participants is evaluated. Similar to [83], we repeat the random selection of the observer 10 times and measure their average saliency prediction performance against the others. This provides robustness against outlier subjects.

4) <u>Infinite human observers</u>: ideally, finding the average of the fixation maps of a large number observers results in the most reliable and accurate representation of the human fixation map. In practice, there would be only a limited number of participants in a test. Therefore, we find an estimate of the performance for infinite number of humans and use that as an upper bound for the performance of visual attention models. To this end, we find the performance of *i* humans in predicting the fixations of *N-i* humans. Note that *N* is the total number of participants (which is 24) and *i* can vary from 1 to 12 (12 observers predicting the fixations of the other 12 observers). For each value of *i*, we select a random group of *i* participants and repeat the process 10 times to ensure more robustness (less bias towards particular subjects). Fig. 4.14 shows the performance of the observers in predicting fixations of other observers. To find an estimate of the infinity-vs-infinity case, we follow a same method used in [83] by fitting a curve with the following form to this graph:

$$AUC(x) = ax^b + c \qquad (4.18)$$

where *a*, *b*, and *c* are constant coefficients and x is the number of observers. The AUC for an infinite number of humans is extrapolated from Fig. 4.14. In other words, the limits of the AUC function when the number of observers approaches infinity is an approximation of the upper bound. The same procedure is repeated for other metrics to find the bounds for each metric. The values of the parameters in (4.18) for different metrics along with their corresponding 95% confidence intervals (CIs) are reported in Table 4.3.

Table 4.3 Curve fitting parameters for different metrics

| Metric | a (CI) | b (CI) | c (CI) | One human | Limit at infinite number of humans |
|---|---|---|---|---|---|
| AUC | -0.28 (-0.30,-0.25) | -0.42 (-0.46,-0.40) | 0.99 (0.98,1) | 0.7173 | 0.99 |
| sAUC | -0.17 (-0.18,-0.16) | -0.76 (-0.81,-0.74) | 0.98 (0.97,1) | 0.8136 | 0.98 |
| NSS | -2 (-2.3,-1.8) | -0.50 (-0.51,-0.48) | 4.27 (4.26,4.28) | 2.3304 | 4.27 |
| PCC | -0.5 (-0.54,-0.48) | -0.39 (-0.42,-0.37) | 1 (0.99,1) | 0.5125 | 1 |
| SIM | -0.49 (-0.52,-0.45) | -0.28 (-0.30,-0.27) | 0.94 (0.93,0.95) | 0.455 | 0.94 |
| KLD | 1.72 (1.7,1.74) | -0.43 (-0.45,-0.41) | 0 (0,0.01) | 1.4164 | 0 |
| EMD | 0.26 (0.25,0.28) | -0.43 (-0.46,-0.41) | 0.03 (0.02,0.04) | 0.2931 | 0.03 |

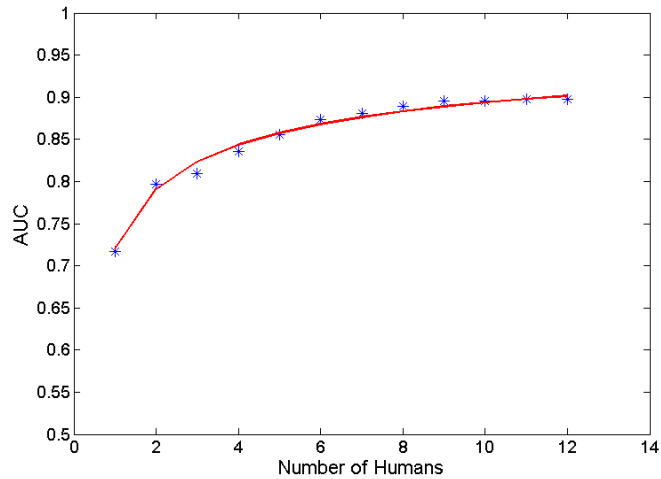

**Figure 4.14 AUC appears to converge to 0.99 for an infinite number of human observers**

*4.3.3 Contribution of Each Proposed Feature Map*

A total of 27 sequences were selected for training the random forest model and the rest (37 videos) were used for performance validation. We categorized the videos in a way that both training and validation sets contain videos with a wide variety of possible scenarios of depth range, brightness, motion, etc. The training set is used to train a random forest model with all 24 features. For a fast implementation, we chose 40 trees and around 7% of the training data (540 frames) for training the model. The impact of learning parameters is investigated in the next subsections. The resulting model achieves AUC=0.7085, sAUC=0.7565, EMD=0.5021, SIM=0.2846, PCC=0.2599, KLD=0.1446, and NSS=1.3067.

The use of random forest ensemble learning makes it possible to extract the relative importance of each feature compared to other ones by comparing their corresponding "out of bag" error values. Table 4.4 shows the relative feature importance values. We can observe that among the high-level features, the presence of humans is of the highest importance. Among the low-level features, motion, brightness, and color are the most important in saliency prediction. Note that since disparity is corrected in these videos and there are no animals in the scenes, the importance values for these two features are equal to zero. Due to the flexibility of the incorporated learning method, any of the existing features can be removed from the model, or new features can be added. Table 4.5 shows the performance evaluation results using different saliency metrics and the validation video set, when only one feature is used. We also study the performance of our method when the first $i$ ($i=1,2,...,24$) important features are used.

Fig. 4.15 shows the performance metrics as functions of the number of features used. The results in this figure verify that higher performance is achieved by using a larger number of features. However, higher number of features increases the complexity for both training and validation. An analysis of complexity is provided in 4.3.6. It is worth mentioning that according to Fig. 4.15, using only the first 12 important features (half of the features) results in 95% percent (in terms of AUC) of the highest algorithm performance (when the entire feature set is used).

Table 4.4 Relative Feature Importance (RFI)

| Feature | RFI | Feature | RFI |
|---|---|---|---|
| Motion2 (Dy) | 1 | Color6 (Contrast-b′) | 0.47 |
| Person | 0.88 | Motion4 (V) | 0.45 |
| Brightness | 0.84 | Color3 (Saturation) | 0.43 |
| Color4 (HVS sensitivity) | 0.74 | Motion7 (Surprise element) | 0.41 |
| Depth | 0.72 | Vehicle | 0.40 |
| Motion1 (Dx) | 0.71 | Text | 0.35 |
| Color7 (Contrast-a′) | 0.63 | Color5 (Empirical) | 0.34 |
| Color2 (Warmth) | 0.58 | Motion5 (Z-emphasis) | 0.33 |
| Motion6 (A) | 0.55 | Horizon | 0.29 |
| Motion3 (Dz) | 0.51 | Face | 0.05 |
| Texture | 0.49 | Animals | 0 |
| Color1 (histogram) | 0.48 | Discomfort | 0 |

Table 4.5 Individual performance of different features using the validation video set

| Feature (alphabetical order) | AUC | sAUC | EMD | SIM | PCC | KLD | NSS |
|---|---|---|---|---|---|---|---|
| Animals | 0.5204 | 0.5000 | 0.9911 | 0.1420 | 0.0897 | 1.9620 | 0.1123 |
| brightness | 0.6023 | 0.6066 | 0.8155 | 0.1703 | 0.1077 | 0.1650 | 0.5777 |
| color1 (histogram) | 0.5673 | 0.5559 | 1.1464 | 0.1512 | 0.0611 | 0.1414 | 0.3238 |
| color2 (Warmth) | 0.5520 | 0.5377 | 1.2118 | 0.1466 | 0.0495 | 0.1184 | 0.3009 |
| color3 (Saturation) | 0.5533 | 0.5303 | 1.0723 | 0.1484 | 0.0501 | 0.1444 | 0.2393 |
| color4 (HVS sensitivity) | 0.5448 | 0.5253 | 1.2688 | 0.1446 | 0.0371 | 0.1279 | 0.1564 |
| color5 (Empirical saliency) | 0.5256 | 0.5087 | 1.1246 | 0.1356 | 0.0136 | 0.1045 | 0.0916 |
| color6 (Contrast-U) | 0.5911 | 0.5880 | 0.9768 | 0.1559 | 0.0747 | 0.1907 | 0.4214 |
| color7 (Contrast-V) | 0.5927 | 0.5865 | 0.9246 | 0.1553 | 0.0703 | 0.1950 | 0.4009 |
| depth | 0.5583 | 0.5459 | 1.3388 | 0.1221 | 0.0623 | 0.1324 | 0.3346 |
| discomfort | 0.5000 | 0.5000 | 0.9912 | 0.1420 | 0.0201 | 1.9620 | 0.0910 |
| face | 0.5221 | 0.5039 | 0.9593 | 0.1430 | 0.0346 | 1.9534 | 0.0445 |
| horizon | 0.5737 | 0.5460 | 0.8016 | 0.1534 | 0.0647 | 1.6581 | 0.2642 |
| motion1 (Dx) | 0.5784 | 0.5960 | 1.0395 | 0.1717 | 0.1448 | 0.0916 | 1.1262 |
| motion2 (Dy) | 0.599 | 0.6129 | 0.7397 | 0.2118 | 0.1633 | 0.2323 | 0.7206 |
| motion3 (Dz) | 0.5072 | 0.5002 | 1.3667 | 0.1246 | -0.011 | 0.0937 | -0.0369 |
| motion4 (V) | 0.5890 | 0.6139 | 1.0458 | 0.1697 | 0.1283 | 0.1147 | 0.9394 |
| motion5 (V-emphasize on z) | 0.5885 | 0.6133 | 1.0503 | 0.1691 | 0.1264 | 0.1164 | 0.9251 |
| motion6 (A) | 0.5541 | 0.5549 | 1.1481 | 0.1501 | 0.0703 | 0.0888 | 0.4828 |
| motion7 (Surprise element) | 0.5302 | 0.5118 | 1.2975 | 0.1394 | 0.0090 | 0.0337 | 0.0313 |
| person | 0.5863 | 0.5923 | 0.7995 | 0.1582 | 0.1203 | 1.5536 | 0.9235 |
| text | 0.5189 | 0.5089 | 1.0135 | 0.1349 | 0.1365 | 1.2851 | 0.2975 |
| texture | 0.5706 | 0.5700 | 1.0573 | 0.1452 | 0.0558 | 0.1673 | 0.3511 |
| vehicle | 0.5302 | 0.5119 | 0.9714 | 0.1451 | 0.0975 | 1.9144 | 0.2138 |

We also evaluated the saliency detection performance of only the low-level features, i.e., the ones extracted from brightness, color, texture, motion, and depth. This helps us understand the influence of low-level and high-level features in visual attention. To this end, a model was trained using the low-level features and was tested using the validation set. The resulting metric values are 0.6879, 0.7355, 0.601, 0.2599, 0.2414, 0.1696, and 1.2028 for AUC, sAUC, EMD, SIM, PCC, KLD, and NSS, respectively. The resulting values are close to the ones corresponding to the proposed model with the entire feature set. This shows that our model does not rely on the high-level features.

*4.3.4 Tuning the Training Parameters*

In this subsection we study the impact of each parameter in the model performance.

1) Size of the training data: Using a different number of frames from each training video may change the learning performance. The training video dataset consists of 27 stereoscopic videos. In the present implementation, we pick the first 20 frames of each video (540 frames in total). To investigate the impact of the size of the training data, we select different number of frames from each sequence to train new models. Performance evaluations show that, in general, even using a small portion of the training dataset results in an acceptable accuracy and that the saliency prediction accuracy is not highly sensitive to the size of the training dataset. In particular, when the size of the training set varies from 27 frames to 1620 frames, the AUC changes between 0.66 and 0.71, sAUC between 0.69 and 0.77, PCC between 0.21 and 0.27, NSS between 1 and 1.4, SIM between 0.24 and 0.29, KLD between 0.19 and 0.10, and EMD varies between 0.61 and 0.49. Using very large training datasets results in higher accuracies, but at the same time increases the overall computational complexity. Fig. 4.16 shows the variations of different metrics with respect to the size of the training dataset. A suitable point of trade-off may be selected depending on required accuracy and complexity.

2) Random forest parameters: In the ensemble random forest learning method, we used boot strapping with sample ratio of 1/3. The minimum number of observations per tree leaf is set to 10. In the current implementation, the number of trees is set to 40 for a fast performance. To study the impact of the number of trees on the saliency prediction, we train additional models with different number of trees, and evaluate their performances over the validation video set. We observe that choosing different number of trees, between 1 and 100, results in very smooth variations in the performance metrics, with slight improvement as the number of trees increases. However, choosing a very large number of trees possibly results in over-fitting and may degrade the overall performance. Fig. 4.17 shows the performance metrics variations against the number of training trees.

### 4.3.5 Comparison with Different Map Fusion Approaches

It is common practice to fuse various conspicuity maps into a final saliency map. The proposed random forest approach combines the individual maps efficiently and according to their relative importance. To demonstrate the strength of random forests in map fusion, we provide a comparison between the performance of our model and different fusion schemes adopted in the state-of-the-art methods. In particular, we compare our method against: 1) Averaging (finding the average of different conspicuity/feature maps), 2) Multiplication, 3) Maximum, 4) Sum plus Product (SpP), 5) Global Non-Linear Normalization followed by Summation (GNLNS) [57], 6) Least Mean Squares Weighted Average (LMSWA), and 7) Standard Deviation Weight (SDW). Table 4.6 and Table 4.7 show the result of these fusion methods and our random forest approach, clearly indicating that our fusion outperforms the other types of map fusion.

### 4.3.6 Number of Features versus Complexity

We provide an estimated complexity of our proposed VAM in terms of simulation running time, since it is not feasible to measure the complexity of visual attention models mathematically as they usually involve very complex methods.

As mentioned before, the cost of the higher accuracy is higher complexity in feature extraction, training, and validation. Therefore, we want to determine a trade-off point where the accuracy is high while the complexity stays at an acceptable level. Fig. 4.18 shows the simulation time with respect to the number of features used (features are used according to their importance reported in Table 4.4).

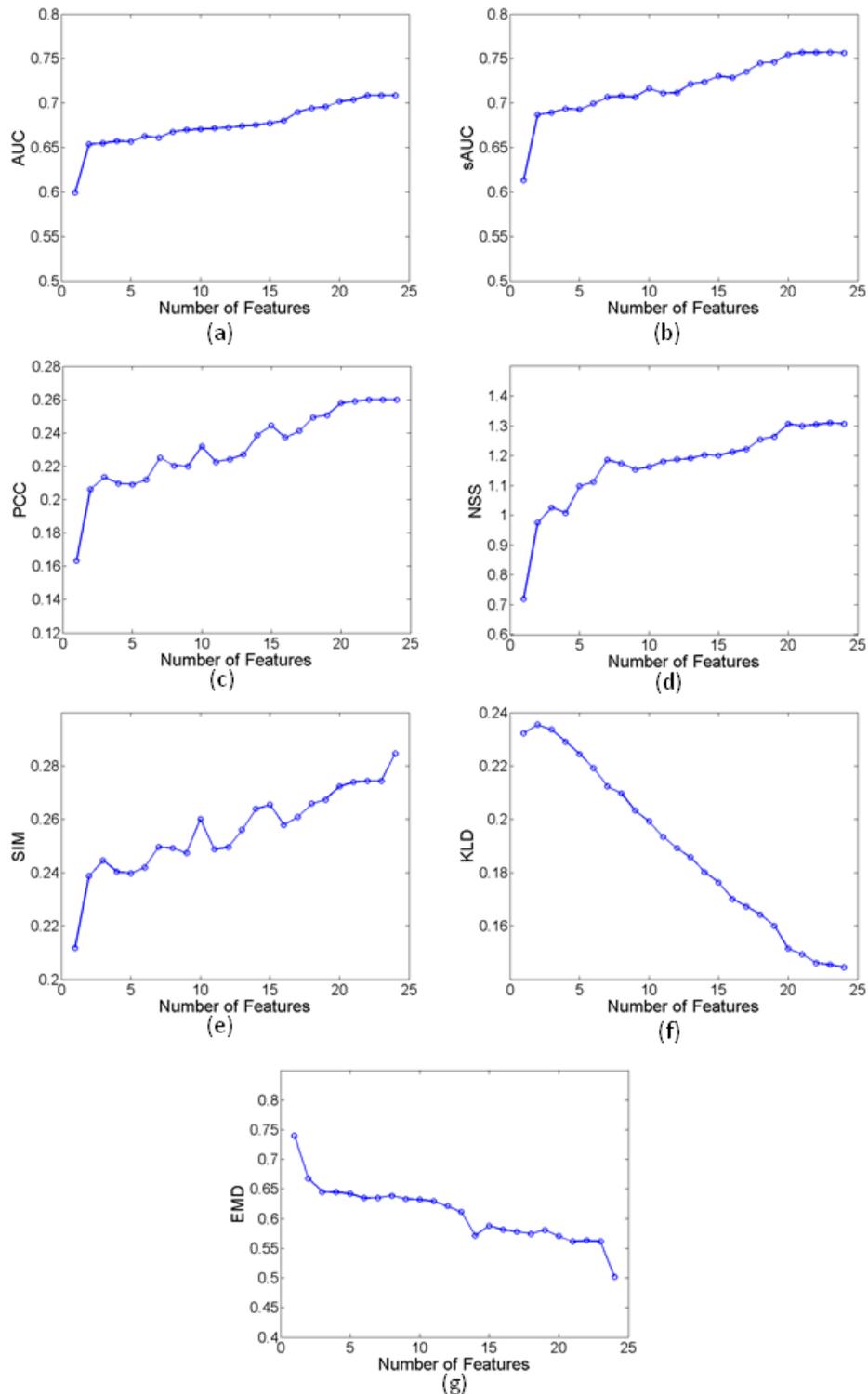

**Figure 4.15 Performance metrics when different number of features are used: AUC (a), sAUC (b), PCC (c), NSS (d), SIM (e), KLD (f), EMD (g)**

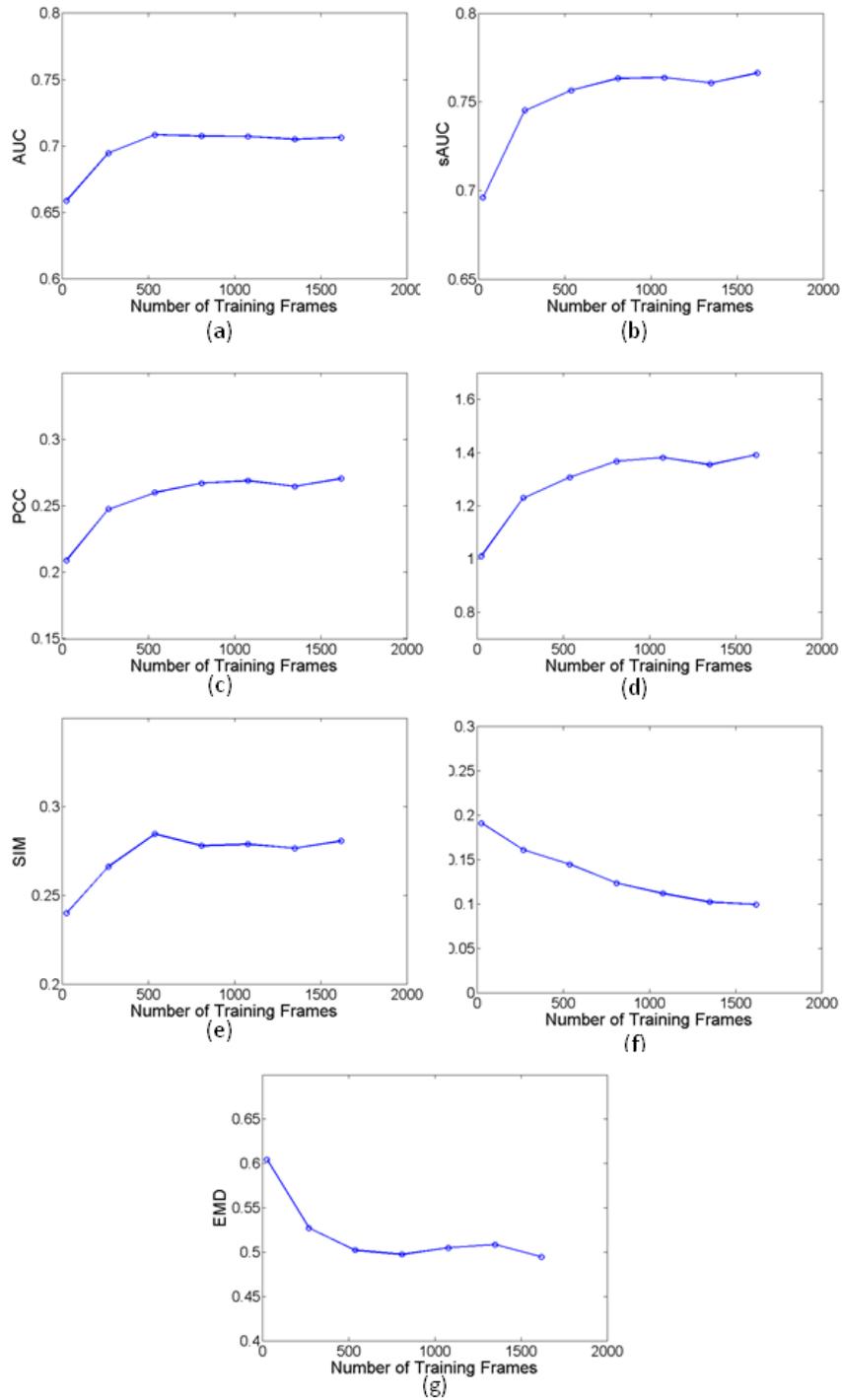

**Figure 4.16 Performance variations when different number of training frames are used: AUC (a), sAUC (b), PCC (c), NSS (d), SIM (e), KLD (f), EMD (g)**

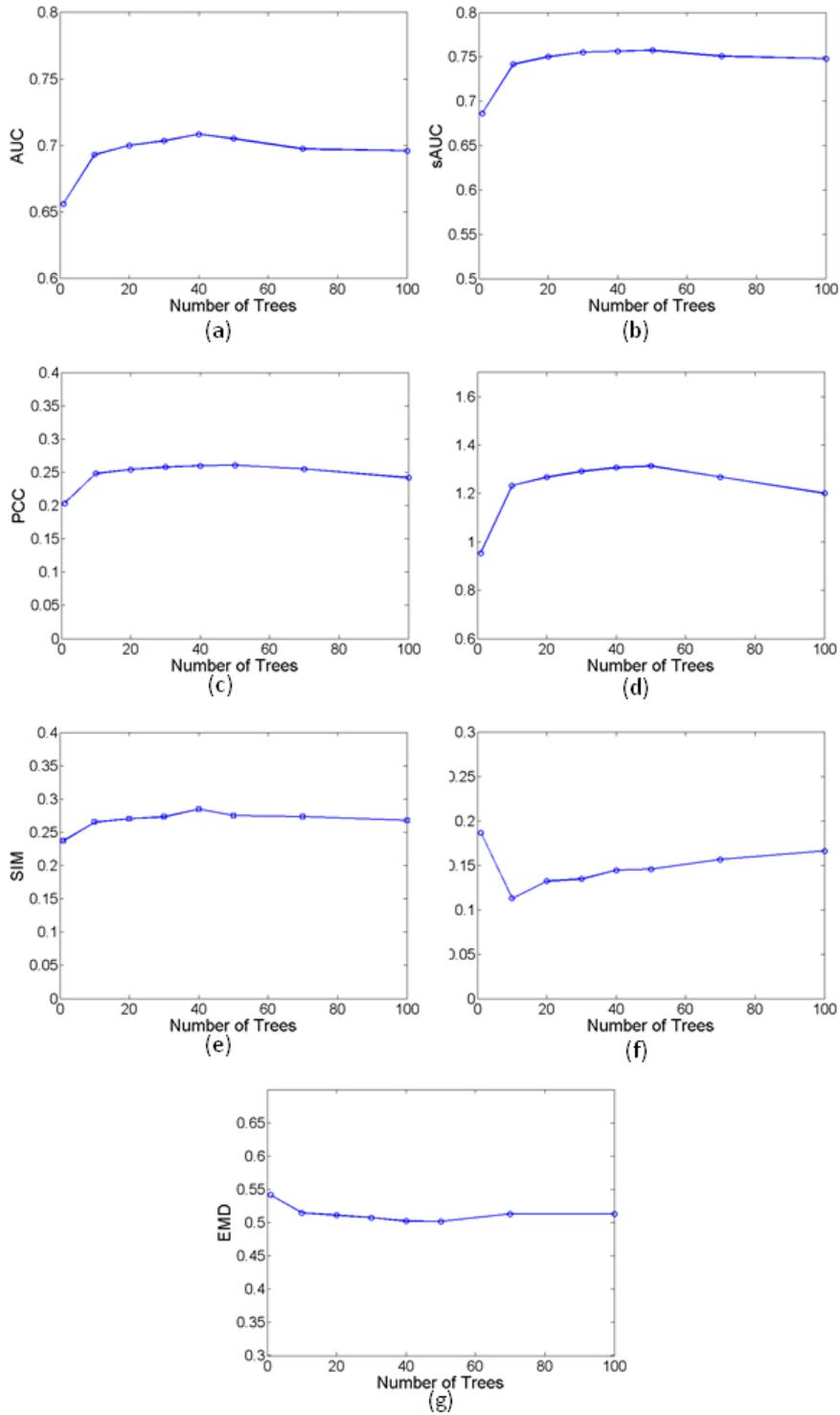

**Figure 4.17 Performance metrics variations when different number of trees are used: AUC (a), sAUC (b), PCC (c), NSS (d), SIM (e), KLD (f), EMD (g)**

**Table 4.6 Comparison of different feature fusion methods**

| Fusion Method | Sample Video 01 | Sample Video 02 |
|---|---|---|
| Video Frame | 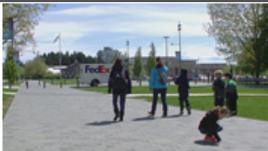 | 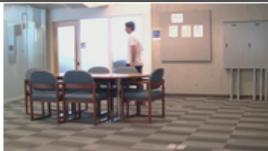 |
| Average | 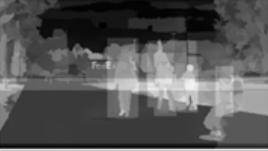 | 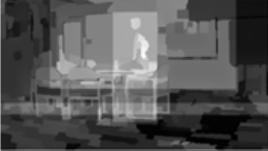 |
| Multiplication | 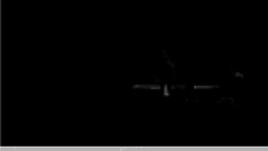 | 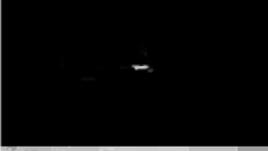 |
| Maximum | 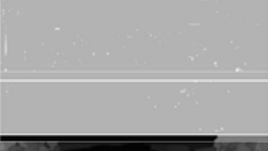 | 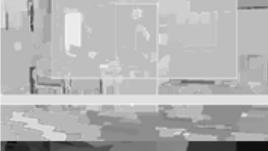 |
| Sum plus product | 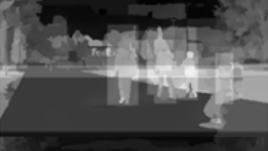 | 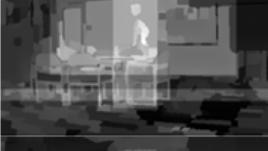 |
| GNLNS | 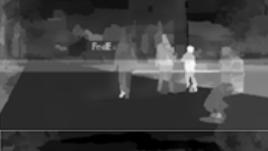 | 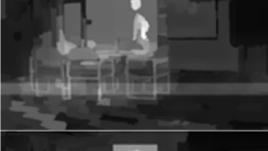 |
| Least Mean Squares Weighted Average | 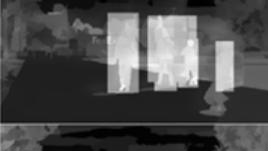 | 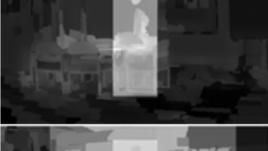 |
| Weighting according to STD of each map | 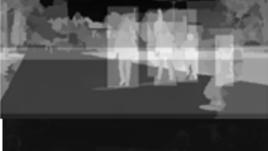 | 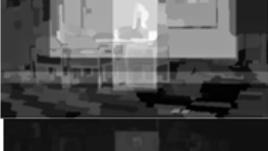 |
| **Random Forest** | 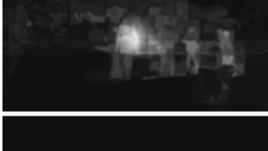 | 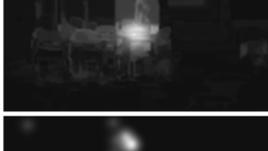 |
| Ground Truth | 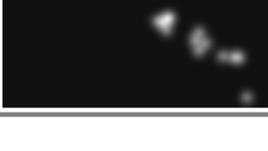 | 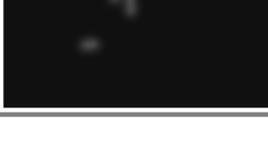 |

**Table 4.7 Evaluation of different feature fusion methods**

| Fusion Method | AUC | sAUC | EMD | SIM | PCC | KLD | NSS |
|---|---|---|---|---|---|---|---|
| Average | 0.642 | 0.689 | 0.969 | 0.217 | 0.196 | 0.220 | 1.064 |
| Multiplication | 0.530 | 0.530 | 0.943 | 0.172 | 0.153 | 1.992 | 0.693 |
| Maximum | 0.554 | 0.556 | 1.066 | 0.175 | 0.171 | 0.848 | 0.799 |
| SpP | 0.642 | 0.688 | 0.969 | 0.217 | 0.196 | 0.218 | 1.064 |
| GNLNS [57] | 0.655 | 0.666 | 1.042 | 0.196 | 0.238 | 1.381 | 1.141 |
| LMSWA | 0.657 | 0.712 | 0.797 | 0.234 | 0.221 | 0.193 | 1.258 |
| SDW | 0.634 | 0.666 | 0.966 | 0.214 | 0.176 | 0.208 | 0.867 |
| **Random Forest** | **0.709** | **0.757** | **0.502** | **0.285** | **0.260** | **0.145** | **1.307** |

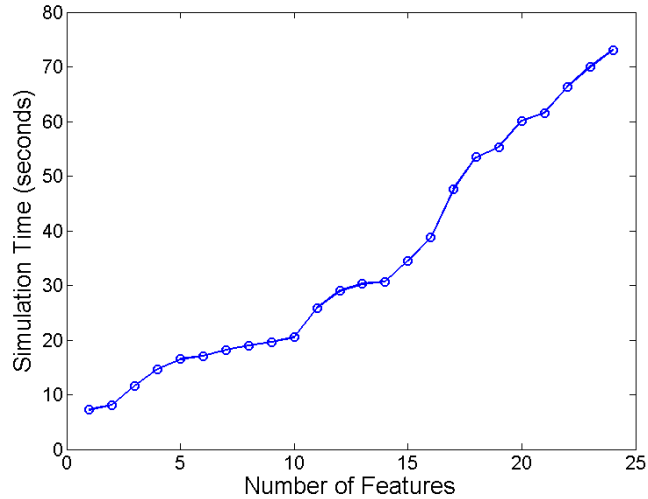

**Figure 4.18 Complexity versus the number of features**

It is worth mentioning that in our method, the resulting model uses random forest regression to generate the saliency maps for different test frames. The complexity values in Fig. 4.18 are calculated by summing the feature extraction time with the saliency map generation time per each frame. Moreover, most of the time is spent for feature extraction. The testing takes around 5% of the whole simulation time (our workstation has an i7 CPU with 18 GBs of memory), and the rest is spent on feature extraction.

*4.3.7 Comparison with the State-of-the-Art VAMs*

This section provides a comparison between the performance of the existing 2D and 3D visual attention models over our stereoscopic video dataset. Currently, we have considered 50 different models. Any new model can be submitted via our online platform and will be added to the benchmark [125]. The selected models were originally designed for 2D/3D image/video saliency detection. The following VAMs are considered in our comparisons (in alphabetical order): AIM (Attention based on Information

Maximization) [63], AIR (Saliency Detection in the Compressed Domain for Adaptive Image Retargeting) [218], Bayesian (Saliency via Low and Mid-Level Cues) [219], BMS (Boolean Map based Saliency) [52], Chamaret [74], Context-Aware saliency [64], CovSal (Visual saliency estimation by nonlinearly integrating features using region covariances) [54], DCST (Video Saliency Detection via Dynamic Consistent Spatio-Temporal Attention Modelling) [220], LBVS-3D (Learning based Visual Saliency prediction for 3D video) [124], DSM (Depth Saliency Map) [69], Fan [175], Fang [68], FT (Frequency-tuned salient region detection) [59], GBVS (Graph-Based Visual Saliency) [56], HC (Global Contrast based Salient Region Detection) [58], HDCT (Salient Region Detection via High-dimensional Color Transform) [165], HSaliency (Hierarchical Saliency Detection) [166], HVSAS (Bottom-up Saliency Detection Model Based on Human Visual Sensitivity and Amplitude Spectrum) [221], ImgManipulate (Saliency for Image Manipulation) [61], Itti (without motion features) [57], Itti video (including the motion features) [57], Jiang [122], Ju [174], Judd [53], Kocberber [167], LC (when applied to images only) [60], LC video (Visual attention detection in video sequences using spatiotemporal cues) [60], Ma [168], Manifold Ranking [222], Self-resemblance video [65], Niu [70], Ouerhani [75], Park [172], PCA [223], Rahtu [169], Rare (rarity-based saliency detection) [62], RC (Saliency using Region Contrast) [58], RCSS (saliency map based on sampling an image into random rectangular regions of interest) [224], FES (Fast and Efficient Saliency) [55], SDSP (saliency detection method by combining simple priors) [225], Self-resemblance (for static scene only) [65], SIM-saliency (Saliency Estimation using a non-parametric vision model) [226], Spectral Saliency [227], SR (spectral residual approach) [123], SUN (Saliency Using Natural statistics) [67], SWD (Visual Saliency Detection by Spatially Weighted Dissimilarity) [228], Torralba [229], Coria [146], Yubing [170], and Zhang [73]. Note that for the 2D image/video models, the saliency is computed only for one of the views (right view). Moreover, we compare the performance of the different visual attention models with those of the four baseline models mentioned in subsection 4.3.2, as well as the chance and center models.

It is recommended (and is common practice) to perform histogram matching between a saliency map and the corresponding human fixation map [83] so that the saliency evaluation metrics attain more meaningful and fair comparisons. Note that histogram matching forces saliency maps to place majority of salient locations at fixated regions.

In addition to histogram matching, blurring and adding center bias is a common practice for saliency prediction models as they slightly increase the performance of different models [83],[216],[219][230]. To account for the center-bias, a Gaussian disk located at the center of the coordinate axes (with the same size as the saliency map) is added to each saliency map as follows:

$$S' = w.S + (1-w).S_{center} \tag{4.19}$$

where $w$ specifies the weight of the center-bias. The values of the weight of center-bias ($w$) are swept between 0 and 1, while the values of the standard deviation are swept between 0.1 and 500 (pixels). The values resulting in the highest average AUC are selected for each model. To account for the blurring effect, each saliency map is convolved with a Gaussian (of the same size). The standard deviation of the blurring kernel is separately optimized for each VAM by sweeping over a wide range of values. Histogram matching, blurring, and adding a center-bias generally can increase the AUC performance by several percentage points.

Table 4.8 shows the performance of different VAMs with respect to several saliency prediction metrics. In order to compare the accuracy using all of the metrics, we assign a rank to each model using each metric independently and then sorting the resulting ranks. In other words, assuming that various metrics are of the same importance, average of the ranks demonstrates how well each VAM performs with respect to all of the metrics.

### 4.3.8 Complexity of Different VAMs

As mentioned earlier, the mathematical definition of the complexity of an algorithm involves calculating the number of operations. Due to the complex structure of most of the visual attention models, it is not possible to calculate their mathematical complexity. Instead, the only feasible solution to compare the complexity of different algorithms is to compare their simulation times. To this end, we used a workstation with i7 CPU and 18 GBs of memory to perform complexity measurements. The results of complexity measurements are reported in Table 4.8. Fig. 4.19 shows the performance (average rank) versus the complexity for different VAMs used in our benchmark. Complexity is reported as the average simulation time for saliency prediction of one frame.

Table 4.8 Performance evaluation of different VAMs using our eye-tracking dataset of stereoscopic videos

| Model | AUC | sAUC | EMD | SIM | PCC | KLD | NSS | Simulation Time (sec) | Average Rank | Type |
|---|---|---|---|---|---|---|---|---|---|---|
| **Infinite humans** | 0.99 | 0.98 | 0.03 | 0.94 | 1 | 0 | 4.27 | 0 | 1 | |
| **LBVS-3D [124]** | 0.7085 | 0.7565 | 0.5021 | 0.2846 | 0.2599 | 0.1446 | 1.3067 | 73.11 | 4.85 | 3D video |
| **Judd [53]** | 0.676 | 0.7162 | 0.603 | 0.2466 | 0.2398 | 0.218 | 1.3691 | 58 | 8.42 | 2D image |
| **CovSal [54]** | 0.6577 | 0.6757 | 0.605 | 0.2439 | 0.216 | 0.2664 | 1.1867 | 17.9 | 11.57 | 2D image |
| **SWD [228]** | 0.6492 | 0.6809 | 0.6246 | 0.2243 | 0.1956 | 0.249 | 1.1187 | 11.37 | 12.28 | 2D image |
| **Rare [162]** | 0.6307 | 0.6798 | 0.6337 | 0.2078 | 0.1685 | 0.212 | 1.0091 | 1.52 | 13 | 2D image |
| **PCA [223]** | 0.6505 | 0.6759 | 0.6501 | 0.2262 | 0.1929 | 0.2607 | 1.0092 | 17.88 | 13.28 | 2D image |
| **BMS [52]** | 0.6445 | 0.6999 | 0.7719 | 0.2215 | 0.1888 | 0.2422 | 1.1702 | 10 | 14.71 | 2D image |
| **Spectral saliency [227]** | 0.6118 | 0.6526 | 0.7599 | 0.2033 | 0.1726 | 0.1929 | 1.1279 | 0.34 | 15.28 | 2D image |
| **One human** | 0.7173 | 0.8136 | 1.4164 | 0.455 | 0.5125 | 0.2931 | 2.3304 | 0 | 16.28 | |
| **AIR [218]** | 0.6275 | 0.6461 | 0.6491 | 0.202 | 0.1547 | 0.2349 | 0.7697 | 7.58 | 17.57 | 2D image |
| **Fang [68]** | 0.635 | 0.651 | 0.6666 | 0.2091 | 0.1607 | 0.2649 | 0.781 | 3.25 | 17.57 | 3D image |
| **GBVS [56]** | 0.6187 | 0.6527 | 0.656 | 0.189 | 0.1403 | 0.2824 | 0.8246 | 2 | 19.42 | 2D image |
| **ImgManipulate [61]** | 0.6298 | 0.6486 | 0.736 | 0.2014 | 0.1531 | 0.2511 | 0.773 | 12 | 19.71 | 2D image |
| **Itti video [57]** | 0.6099 | 0.6536 | 0.7788 | 0.1851 | 0.14 | 0.2109 | 0.8373 | 4.05 | 19.71 | 2D video |
| **Kocberber [167]** | 0.6143 | 0.6398 | 0.6606 | 0.184 | 0.1394 | 0.2187 | 0.739 | 122 | 21 | 2D video |
| **RC [58]** | 0.6125 | 0.6293 | 0.6578 | 0.1861 | 0.1474 | 0.2252 | 0.763 | 2.2 | 21 | 2D image |
| **FES [55]** | 0.6011 | 0.6278 | 0.6614 | 0.1892 | 0.1364 | 0.2164 | 0.8119 | 0.82 | 21.71 | 2D image |
| **Yubing [170]** | 0.6209 | 0.6448 | 0.6857 | 0.1891 | 0.1354 | 0.2289 | 0.6627 | 40.1 | 21.85 | 2D video |
| **AIM [63]** | 0.5915 | 0.6416 | 0.6071 | 0.1727 | 0.1231 | 0.2134 | 0.8468 | 25.88 | 22 | 2D image |
| **HDCT [165]** | 0.6039 | 0.6294 | 0.7723 | 0.179 | 0.1363 | 0.2087 | 0.7454 | 153.24 | 23.42 | 2D image |
| **Context Aware [64]** | 0.6024 | 0.6392 | 0.8472 | 0.1831 | 0.1311 | 0.2088 | 0.765 | 24.16 | 23.71 | 2D image |
| **Center** | 0.6064 | 0.6071 | 0.6936 | 0.2219 | 0.1177 | 0.2758 | 0.8827 | 0.06 | 23.85 | |
| **Manifold Ranking [222]** | 0.6142 | 0.6116 | 0.6776 | 0.1912 | 0.1582 | 0.9531 | 0.704 | 12.94 | 23.85 | 2D image |
| **HSaliency [166]** | 0.603 | 0.6122 | 0.6945 | 0.1768 | 0.1308 | 0.2112 | 0.6276 | 14.28 | 26.57 | 2D image |
| **Self-resemblance static [65]** | 0.6043 | 0.6321 | 0.8776 | 0.1741 | 0.1115 | 0.1994 | 0.5948 | 2.12 | 27 | 2D image |
| **Park [172]** | 0.5962 | 0.632 | 0.808 | 0.17 | 0.1139 | 0.2082 | 0.6863 | 1.68 | 27.14 | 3D image |
| **RCSS [224]** | 0.5974 | 0.6137 | 0.6158 | 0.1787 | 0.1265 | 1.7298 | 0.6312 | 15.86 | 27.57 | 2D image |
| **Itti [57]** | 0.5963 | 0.6284 | 0.8102 | 0.1692 | 0.114 | 0.2103 | 0.6311 | 1.8 | 28.42 | 2D image |
| **SDSP [225]** | 0.5961 | 0.6037 | 0.6205 | 0.1808 | 0.1309 | 1.4002 | 0.6173 | 0.19 | 28.42 | 2D image |
| **DCST [220]** | 0.5602 | 0.6181 | 1.286 | 0.1453 | 0.1208 | 0.1247 | 0.9526 | 146.7 | 29.42 | 2D video |
| **Rahtu [169]** | 0.5732 | 0.5688 | 0.6994 | 0.168 | 0.099 | 0.1559 | 0.45 | 4.84 | 30.14 | 2D video |
| **SIM [226]** | 0.5795 | 0.6218 | 1.1066 | 0.1687 | 0.1115 | 0.1542 | 0.7215 | 12.2 | 30.42 | 2D image |
| **Coria [146]** | 0.5382 | 0.5687 | 0.8989 | 0.154 | 0.0901 | 0.1379 | 0.698 | 3.03 | 32.14 | 3D video |
| **Self-resemblance dynamic [65]** | 0.5631 | 0.5867 | 0.958 | 0.1554 | 0.0944 | 0.1709 | 0.5623 | 1.29 | 33.14 | 2D video |
| **Bayesian [219]** | 0.5614 | 0.5661 | 0.8872 | 0.1588 | 0.0793 | 0.1797 | 0.4035 | 149.24 | 34.14 | 2D image |
| **SR [123]** | 0.5529 | 0.5848 | 1.255 | 0.1523 | 0.0838 | 0.1365 | 0.589 | 1.9 | 34.85 | 2D image |
| **FT [59]** | 0.553 | 0.5618 | 1.0611 | 0.1565 | 0.0774 | 0.1489 | 0.4133 | 2 | 35.28 | 2D image |
| **DSM [69]** | 0.5461 | 0.5545 | 0.8881 | 0.1373 | 0.0933 | 0.1057 | 0.1391 | 39.44 | 35.57 | 3D image |
| **LC [60]** | 0.5397 | 0.551 | 1.0526 | 0.1531 | 0.0729 | 0.1148 | 0.4225 | 2.3 | 36.14 | 2D image |
| **Torralba [229]** | 0.5405 | 0.5625 | 0.9806 | 0.1499 | 0.0606 | 0.1391 | 0.4099 | 3.38 | 36.57 | 2D image |
| **Ouerhani [75]** | 0.5337 | 0.5504 | 0.9481 | 0.147 | 0.0537 | 0.1405 | 0.3145 | 7.21 | 39.14 | 3D image |
| **Fan [175]** | 0.5454 | 0.5407 | 0.9104 | 0.1504 | 0.0499 | 0.1464 | 0.183 | 128.98 | 39.28 | 3D image |
| **HC [58]** | 0.5497 | 0.553 | 1.1065 | 0.1528 | 0.0621 | 0.1636 | 0.2919 | 2.1 | 39.42 | 2D image |
| **Niu [106]** | 0.5411 | 0.545 | 1.0013 | 0.1471 | 0.0445 | 0.1502 | 0.2161 | 165.82 | 40.57 | 3D image |
| **HVSAS [221]** | 0.5303 | 0.5599 | 0.906 | 0.1297 | 0.0333 | 0.1434 | 0.2255 | 24.6 | 40.85 | 2D image |
| **LC video [60]** | 0.5317 | 0.548 | 0.9062 | 0.1384 | 0.0206 | 0.1139 | 0.1645 | 1.96 | 40.85 | 2D video |
| **Ma [168]** | 0.5349 | 0.5383 | 1.0671 | 0.1286 | 0.0486 | 0.0953 | -0.0619 | 39.12 | 42 | 2D video |
| **SUN [67]** | 0.5324 | 0.5552 | 1.0808 | 0.1324 | 0.0428 | 0.1551 | 0.3086 | 3.07 | 42.57 | 2D image |
| **Jiang [122]** | 0.5458 | 0.5343 | 1.0761 | 0.1168 | 0.0332 | 0.1085 | -0.1375 | 1.25 | 43.14 | 3D image |
| **Ju [174]** | 0.5384 | 0.5395 | 1.1968 | 0.1299 | 0.0244 | 0.0997 | -0.0723 | 2.05 | 43.42 | 3D image |
| **Chamaret [74]** | 0.5347 | 0.5458 | 1.0852 | 0.1326 | 0.0853 | 0.2405 | 0.3305 | 64.62 | 44.28 | 3D video |
| **Zhang [73]** | 0.525 | 0.5374 | 1.0624 | 0.1144 | 0.0255 | 0.0993 | -0.0892 | 0.73 | 44.28 | 3D image |
| **Chance** | 0.5 | 0.5 | 1.4662 | 0.136 | 0 | 0.0716 | -0.0019 | 0.072 | 45 | |

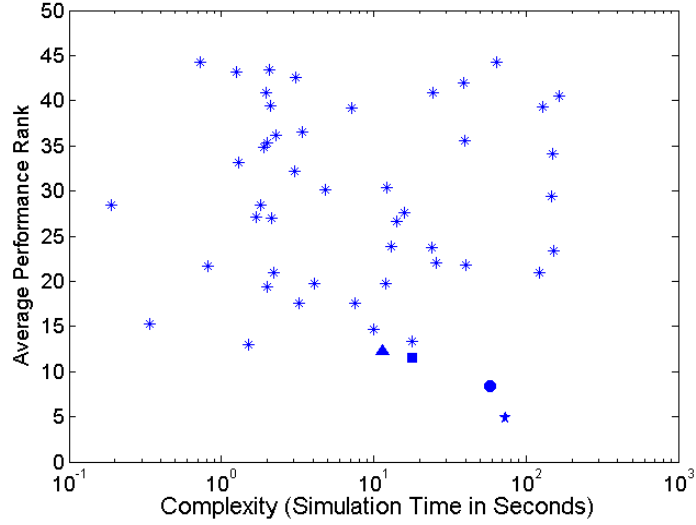

**Figure 4.19 Average performance ranking depicted in terms of algorithm complexity: The top four models in Table 4.8 are labeled as Star, Circle, Square, and Triangle, respectively**

*4.3.9  Derivation of Image Moments Used in Section 4.1*

In this subsection we provide the definitions for the image moments used in the design of our proposed 3D VAM in Section 4.1.

Suppose *I(x,y)* is a two dimensional grayscale image. Then, the raw image moments are defined by:

$$M_{ij} = \sum_x \sum_y x^i y^j I(x,y) \qquad (4.20)$$

The center of mass, or centroid, is calculated as:

$$\bar{C} = (\bar{x}, \bar{y}) = \left(\frac{M_{10}}{M_{00}}, \frac{M_{01}}{M_{00}}\right) \qquad (4.21)$$

The central image moments are defined as:

$$\mu_{pq} = \sum_x \sum_y (x-\bar{x})^p (y-\bar{y})^q I(x,y) \qquad (4.22)$$

Based on the central moments, scale and translation invariant moments can be defined using the following formula:

$$\eta_{ij} = \frac{\mu_{ij}}{\mu_{00}^{\left(1+\frac{i+j}{2}\right)}} \qquad (4.23)$$

Using these moments, it is possible to define moments which are invariant to scale, translation, and rotation. The moment of inertia is one such moment which belongs to the Hu set of invariant moments [231]:

$$I_1 = \eta_{20} + \eta_{02} \tag{4.24}$$

## 4.4 Conclusion

This chapter introduces a new computational visual attention model for stereoscopic 3D video. Both low and high level features are incorporated in the design of our model. Several intuitive biological observations are quantified and adopted in our method. A random forest learning algorithm is utilized to train a saliency prediction model and efficiently fuse various feature maps generated by the proposed approach. Our method is flexible in that it allows new features to be added without changing the structure of the model. To verify the performance of the proposed VAM, we capture a dataset of stereoscopic videos and collect their eye-tracking results. Performance evaluations demonstrated the high performance of our visual attention model.

# 5 Saliency Inspired Quality Assessment of Stereoscopic 3D Video

This chapter investigates the added value of incorporating LBVS-3D VAM into Full-Reference and No-Reference quality assessment metrics for stereoscopic 3D video. To this end, our proposed 3D VAM introduced in Chapter 4 is integrated to quality assessment pipeline of various existing FR and NR stereoscopic video quality metrics. We compare the performance of the metrics before and after using the 3D saliency information. The rest of this chapter is organized as follows: Section 5.1 describes our saliency integration methodology, Section 5.2 contains details regarding our experiments, Section 5.3 provides the results, and Section 5.4 concludes the chapter.

## 5.1 Methodology

Video quality metrics usually perform the quality assessment task by measuring similarities (when a reference is available) or distortion densities (when no reference is available) for partitions of the video and then combining the local partition measurements to an overall quality index in a process known as pooling. Quality pooling can be done spatially (for image quality assessment), or temporally (for video quality assessment). It is therefore possible, for these types of metrics, to incorporate saliency prediction results in the pooling stage of quality assessment pipeline. Fig. 5.1 shows the proposed saliency integration scenario. The rest of this section elaborates on various FR and NR quality metrics used in our experiments, and how to integrate saliency detection results in each quality metric. Fig. 5.2 demonstrates an example of saliency detection from stereoscopic video using 3D VAM of LBVS-3D, and how visual attention is drawn to salient objects.

### 5.1.1  Integration of Saliency Maps into FR Quality Metrics

Visual attention models provide a saliency map for each frame of a video. To be able to use the saliency prediction results in video quality assessment, we only consider those video quality metrics which produce a map of similarities, distortions, transform coefficients, errors, or in general the ones for which it is possible to apply the saliency map as a weighting mask.

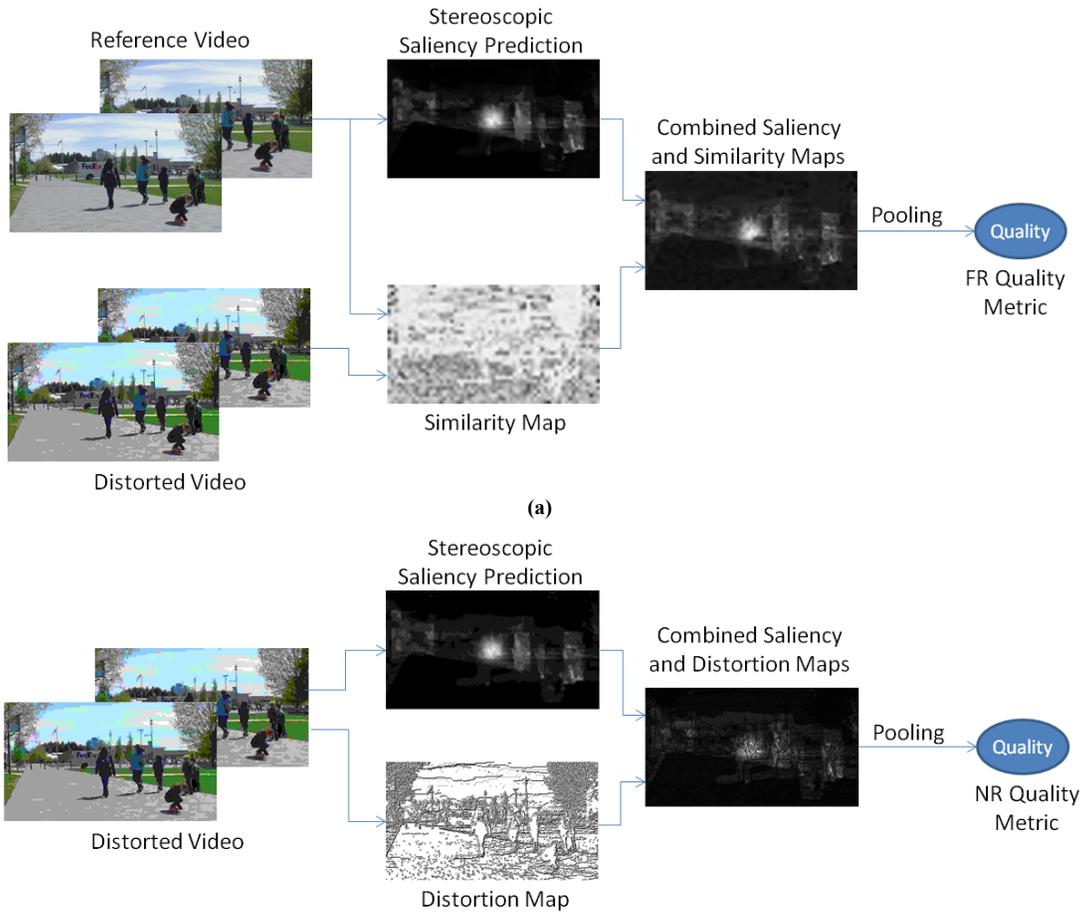

**Figure 5.1 Saliency inspired quality assessment for stereoscopic video: (a) Full-Reference (FR) and (b) No-Reference (NR) case**

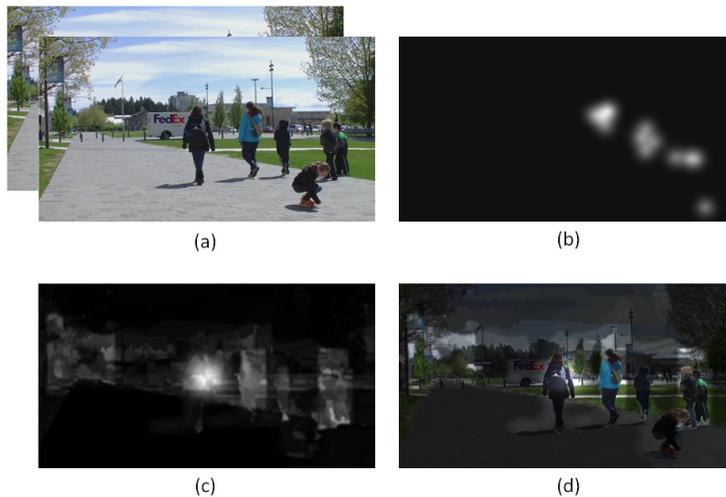

**Figure 5.2 Saliency prediction on stereoscopic video: (a) Original video, (b) eye fixation maps from eye tracking experiments, (c) saliency prediction using LBVS-3D method, and (d) saliency map super imposed on the original video**

Note that in the case of FR quality evaluation, saliency maps are generated using the reference stereo pair.

In this chapter, we integrate 3D VAM of LBVS-3D into the following FR 3D quality metrics: Ddl1 [16], OQ [17], CIQ [18], PHVS3D [19], PHSD [9], MJ3D [21], Q_Shao [20], and HV3D [126]. In addition, we follow what is considered to be common practice in 3D quality evaluation by using PSNR, SSIM [13], MS-SSIM [157], and VIF [135] for FR metric integration. In the case of 2D metrics, the overall 3D quality is resulted from averaging the frame qualities in the two views. The rest of this subsection elaborates on saliency integration for various FR metrics.

1) <u>PSNR</u>: PSNR is calculated based on the Mean Squared Error (MSE) as:

$$PSNR = 10\log\left(\frac{255^2}{MSE}\right) \quad (5.1)$$

We modify the MSE based on saliency maps as follows:

$$MSE_S = E_t\left\{E_{x,y}\left\{|I(x,y,t)-I'(x,y,t)|^2 \times S(x,y,t)\right\}\right\} \quad (5.2)$$

where $x$ and $y$ are pixel coordinates, $t$ denotes the frame number, $I$ and $I'$ are reference and distorted frames, $S$ is the normalized saliency map, and $E_{x,y}$ and $E_t$ denote spatial and temporal mean operators, respectively. Note that MSE and PSNR are calculated for left and right views separately (using the same saliency map), and the average PSNR is considered for each stereo pair.

2) <u>SSIM</u> [13]: Saliency based SSIM is computed as:

$$SSIM_S = E_t\left\{E_{x,y}\left\{SSIM\{I(x,y,t),I'(x,y,t)\}\times S(x,y,t)\right\}\right\} \quad (5.3)$$

where $SSIM\{I(x,y,t),I'(x,y,t)\}$ is the local structural similarity value at pixel location $(x,y)$ and time $t$. Similar to PSNR, the SSIM values are calculated for each view separately and averaged for the pair.

3) <u>MS-SSIM</u> [157]: Multi-Scale SSIM evaluates structural distortions for a pair of reference-distorted images at a number of scales [157]. In order to apply saliency prediction to MS-SSIM, we generate the saliency maps at each scale independently. Then saliency inspired MS-SSIM is evaluated as follows:

$$MSSSIM_S = E_t\left\{E_{x,y}\left\{l_M(x,y,t)\prod_{m=1}^{M}\{c_m(x,y,t)st_m(x,y,t)\times S_m(x,y,t)\}\right\}\right\} \quad (5.4)$$

where $l_m$, $c_m$, and $st_m$ assess the luminance, contrast, and structure distortions at scale $m$, $S_m$ is the generated saliency map at scale $m$, and $M$ is the number of decomposition scales. Since MS-SSIM is a 2D metric, the saliency based MS-SSIM is evaluated for each view separately and the average is calculated as the overall index.

4) <u>VIF</u> [135]: Visual Information Fidelity index evaluates the ratio of visual information between the reference and distorted images [135]. Pixel-wise implementation of VIF is used as follows:

$$VIF_S = E_t \left\{ \frac{\sum_{m=1}^{M} \left\{ \frac{1}{2} \sum_i \sum_k \log\left(1 + \frac{g^2 s_i^2 \lambda_k}{\sigma_v^2 + \sigma_n^2}\right) \times S_m(i,k) \right\}}{\sum_{m=1}^{M} \left\{ \frac{1}{2} \sum_i \sum_k \log\left(1 + \frac{s_i^2 \lambda_k}{\sigma_n^2}\right) \times S_m(i,k) \right\}} \right\} \quad (5.5)$$

where $M$ is the number of decomposition subbands, $S_m$ is saliency map generated at the size of subband $m$, $i$ and $k$ are spatial subband indices, $s_i$, $\lambda_k$, $\sigma_n$, and $\sigma_v$ are VIF parameters (See [135] for the details).

5) <u>Ddl1</u> [16]: This metric performs quality assessment on stereoscopic 3D images by weighting the structural similarity values of each view according to their Euclidean disparity differences. Here, we adjust this weighting by incorporating saliency values as follows:

$$\begin{aligned} Ddl1_{S-left} &= E_{x,y} \left\{ SSIM\{I_L(x,y,t), I'_L(x,y,t)\} \right. \\ &\quad \left. \times \left(1 - \frac{\sqrt{D_L(x,y,t)^2 - D'_L(x,y,t)^2}}{255}\right) \times S(x,y,t) \right\} \\ Ddl1_{S-right} &= E_{x,y} \left\{ SSIM\{I_R(x,y,t), I'_R(x,y,t)\} \right. \\ &\quad \left. \times \left(1 - \frac{\sqrt{D_R(x,y,t)^2 - D'_R(x,y,t)^2}}{255}\right) \times S(x,y,t) \right\} \\ Ddl1_S &= E_t \left\{ Ddl1_{S-left}(t) + Ddl1_{S-right}(t) \right\} \end{aligned} \quad (5.6)$$

where $D$ and $D'$ are the disparity maps and $L$ and $L'$ denote the left and right view, respectively.

6) <u>OQ</u> [17]: You et al. modeled the full-reference quality as combination of disparity map quality and quality of the views. They use Mean Absolute Differences (MAD) to measure the changes in the disparity map and SSIM for view quality measurement [17].

In order to modify this metric according to saliency map values, we apply the saliency map as weighting factor both in disparity map and view quality evaluations:

$$OQ_S = E_t\{a.IQ_S^d(t) + bDQ_S^e(t) + c.IQ_S^d(t).DQ_S^d(t)\} \tag{5.7}$$

where

$$\begin{aligned} IQ_S(t) &= SSIM_S(t) \\ DQ_S(t) &= E_{x,y}\{|D(x,y,t) - D'(x,y,t)| \times S(x,y,t)\} \end{aligned} \tag{5.8}$$

and *a, b, c, d,* and *e* are constant coefficients which are derived based on subjective experiments. To conduct a fair comparison, we use the same constants reported in [17].

7) <u>CIQ</u> [18]: Chen et al. proposed a FR quality assessment framework for stereo images based on generating cyclopean view (fusion of the left and right views in a single image) pictures for reference and distorted views. Then, the overall quality is evaluated as the SSIM between the cyclopean pictures from reference and distorted pairs. Saliency inspired CIQ is formulated as:

$$CIQ_S = E_t\{E_{x,y}\{SSIM\{CI(x,y,t), CI'(x,y,t)\} \times S(x,y,t)\}\} \tag{5.9}$$

where *CI* and *CI'* denote the cyclopean view images from the reference and distorted signals.

8) <u>PHVS3D</u> [19]: This metric takes into account the MSE of 3D block structures between the reference and distorted stereo pairs [19]. Saliency based PHVS3D is defined by:

$$PHVS3D_S = E_t\left\{10\log\left(\frac{255^2}{MSE_{3D-S}(t)}\right)\right\} \tag{5.10}$$

$$MSE_{3D-S}(t) = \frac{16}{H.W} \sum_{x=1}^{H-3} \sum_{y=1}^{W-3} MSE\{A_{xy}(t) - B_{xy}(t)\}.C_{4\times4}^2.S(x,y,t) \tag{5.11}$$

where *H* and *W* denote the height and width of the image, $A_{xy}$ and $B_{xy}$ are 3D-DCT coefficients for the reference and distorted views, and *C* is a Contrast Sensitivity Function (CSF) mask [19].

9) <u>PHSD</u> [9]: This metric is an improved version of PHVS3D (mentioned above) which considers the MSE between the depth maps in conjunction with the MSE of block structures [9]. Our modification of PHSD is in two levels of MSE, in both block structures and depth maps, and formulated as:

$$PHSD_S(t) = E_t\left\{10\log\left(\frac{255^2}{(1-\varepsilon).MSE_{i-S} + \varepsilon.MSE_{d-S}}\right)\right\} \qquad (5.12)$$

where

$$MSE_{d-S} = E_t\left\{E_{x,y}\left\{|D(x,y,t) - D'(x,y,t)|^2 \times S(x,y,t)\right\}\right\}$$
$$MSE_{i-S} = E_t\left\{E_{x,y}\left\{MSE_{bs}(x,y).\frac{MSE_{bs}(x,y)}{MSE_{bs}(x,y) + \alpha\sigma_d^2(x,y)} \times S(x,y,t)\right\}\right\} \qquad (5.13)$$

$\sigma_d^2$ is variance of depth map at spatial location $(x,y)$ and $MSE_{bs}$ is the error value calculated for 3D block structures [9].

10) MJ3D [21]: In this approach, Multi Scale SSIM is utilized for quality assessment of cyclopean view pictures constructed from reference and distorted stereo pairs [21]. We modify MJ3D based on saliency maps as follows:

$$MJ3D_S = E_t\left\{E_{x,y}\left\{l_{CI-M}(x,y,t).\prod_{m=1}^{M}(c_{CI-m}(x,y,t)st_{CI-m}(x,y,t) \times S_m(x,y,t))\right\}\right\} \qquad (5.14)$$

where $l_{CI}$, $c_{CI}$, and $st_{CI}$ are the luminance, contrast, and structure components of MS-SSIM in each scale, which are generated from the cyclopean view images.

11) Q_Shao [20]: Shao and colleagues proposed a quality assessment method for stereo images which is based on image region classification. In this method, each view (in both reference and distorted stereo pairs) is classified to three possible partitions: non corresponding ($n_c$), binocular fusion ($b_f$), and binocular suppression ($b_s$) regions. Three quality components are calculated based on the three regions and combined into an overall index as follows [20]:

$$Q = E_t\left\{\omega_{nc}Q_{nc} + \omega_{bf}Q_{bf} + \omega_{bs}Q_{bs}\right\} \qquad (5.15)$$

where $w_{nc}$, $w_{bf}$, and $w_{bs}$ are weighting coefficients for each quality component. Each component is computed as the average of per-pixel values over the corresponding region. Saliency information is therefore incorporated in each of the components, as a weighting factor to emphasize on visually important image regions.

12) HV3D [126]: As mentioned in Chapter 2, Human Visual system based quality measure for 3D video (HV3D) evaluates the perceived 3D quality of stereoscopic videos as a combination of depth map quality and quality of the views. HV3D is formulated as a combination of three terms [126]:

$$HV3D = E_t \left\{ \left( \sum_{i=1}^{N} \frac{SSIM(IDCT(XC_i), IDCT(XC_i'))}{N} \right)^{\beta_1} \cdot \right.$$
$$\left. (VIF(D, D'))^{\beta_2} \cdot \left( \sum_{i=1}^{N} \frac{\sigma_{d_i}^2}{N \cdot \max(\sigma_{d_j}^2 \mid j = 1,2,...,N)} \right)^{\beta_3} \right\}$$
(5.16)

where $XC_i$ is the cyclopean-view model for the ith matching block pair in the reference 3D view, $XC_i'$ is the cyclopean-view model for the ith matching block pair in the distorted 3D view, IDCT stands for inverse 2D discrete cosine transform, N is the total number of blocks in each view, $\beta_1$, $\beta_2$ and $\beta_3$ are constant exponents, and $\sigma_{di}^2$ is the local variance of block i in the disparity map of the 3D reference view. In order to incorporate the saliency information in each of the three quality components of HV3D, we use the saliency based $SSIM_S$ index (see (5.3)) for the first component and the $VIF_S$ index (see (5.5)) for the second component. For the third component (variances of the blocks), a single average saliency value (average of saliency map values) for each block is used as a weight for the variance term. The same constant parameter values are used as the ones reported in [126].

*5.1.2 Integration of Saliency Maps into NR Quality Metrics*

No reference quality assessment is generally a much more difficult task than full reference quality assessment as no information is available about the reference data. NR quality metrics usually aim at evaluating the quality when only a specific type of distortion is present. Due to widespread applications of image/video compression, most of the existing NR quality metrics assess sharpness, blurriness, or blockiness of images or videos. Compared to FR 3D video quality metrics, there is less number of NR 3D metrics proposed in the literature. We use QA3D [26], NOSPDM [27], and Q_Ryu [28] for saliency integration, as these are NR 3D metrics that can be modified according to the available saliency maps. In addition, we also apply our saliency maps to several other 2D metrics, which include: IQVG [116], GBIM [232], NRPBM [233], Q_blur_Farias [117], Q_block_Farias [117], Q_Sadaka [118], VQSM [119], and AQI [234]. In the case of 2D metrics, the overall quality is measured as the average quality of the frames for the two views. Note that in the case of NR quality assessment, saliency maps are generated using the distorted stereo pair as no reference is available. This requires accurate saliency prediction from distorted videos. LBVS-3D is capable of efficiently detect the salient regions in a video, even in the presence of distortions. Fig. 5.3 demonstrates examples of distorted video frames and how saliency maps are extracted using the LBVS-3D. The rest of this subsection elaborates on saliency integration for each metric (It is worth noting that the idea of saliency integration in quality metrics is also applicable in other media domains such as HDR or multiview quality assessment [235][237]).

1) <u>IQVG</u> [116]: Image Quality index based on Visual saliency guided sampling and Gabor filtering, IQVG, performs blind quality assessment of 2D images by applying Support Vector Regression (SVR) on features extracted from sampled image patches.

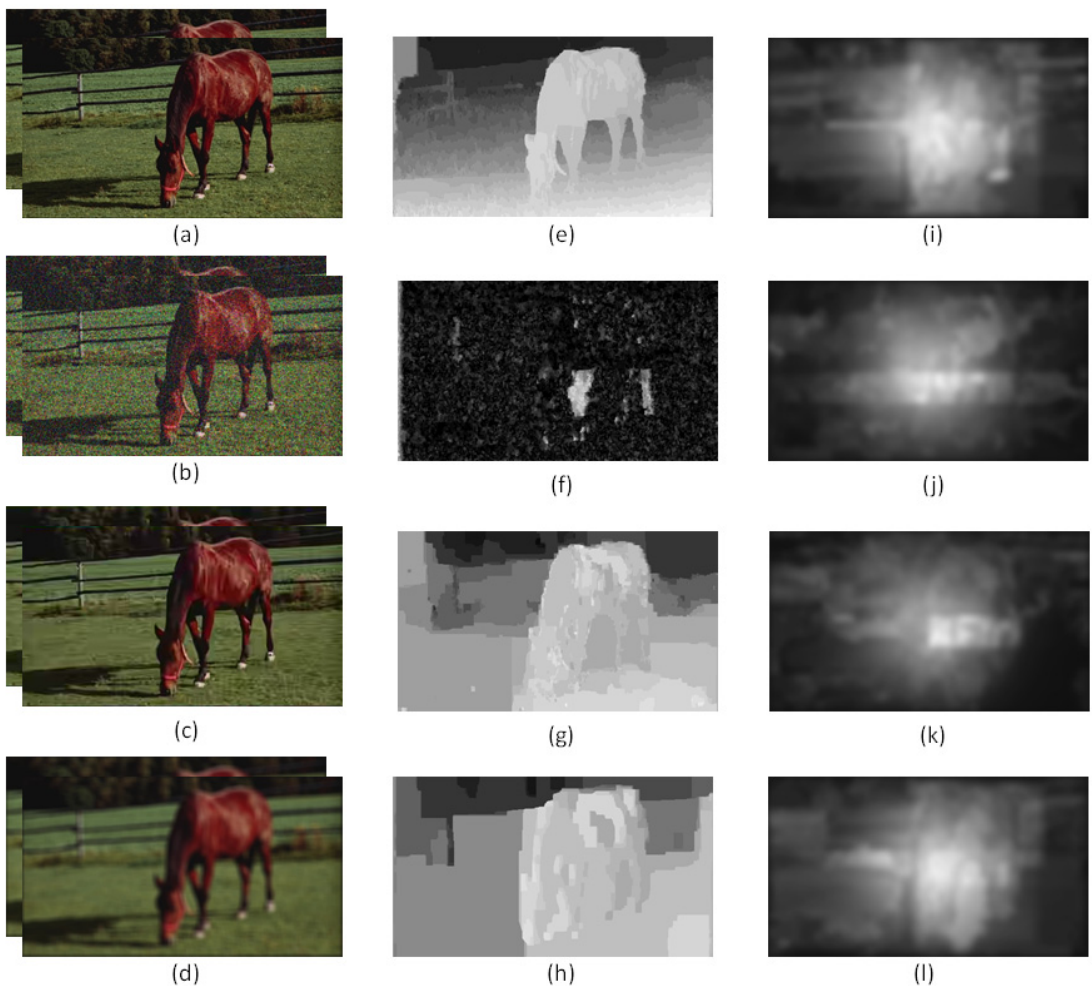

**Figure 5.3 Effect of distortion on depth map generation and saliency prediction; video frames: (a) reference video, (b) Additive White Gaussian Noise (AWGN), (c) 3D video compression, and (d) Gaussian blur, generated depth maps: (e), (f), (g), and (h), and predicted saliency maps using LBVS-3D method: (i), (j), (k), and (l)**

These patches are selected based on 2D saliency information. Here, we swap the 2D saliency maps used in IQVG with the stereo saliency maps of LBVS-3D. The rest of the process remains unchanged [116].

2) <u>GBIM</u> [232]: Generalized Block-Edge Impairment Metric, GBIM, measures blockiness artifacts present in digital video and image coding [232]. Blockiness across horizontal and vertical edges are averaged to formulated the GBIM as:

$$GBIM = E_t \left\{ \frac{M_h + M_v}{2 \times E} \right\} \qquad (5.17)$$

where $E$ is the average inter-pixel difference, and $M_h$ and $M_v$ measure the horizontal and vertical edge blockiness. 3D saliency based $M_h$ is defined by:

$$M_{h-S} = E_{x,y} \{ W(x,y).D_c F(x,y).S(x,y) \} \qquad (5.18)$$

where $W$ is diagonal weighting matrix and $D_c F$ is the inter-pixel difference between each of the horizontal block boundaries [232]. $M_{v-S}$ is defined similarly.

3) <u>NRPBM</u> [233]: Roffet et al. proposed a blurriness metric for 2D images based on the average horizontal and vertical block edge differences in the reference and distorted pictures [233]. We use 3D saliency map values as weights to these differences towards the overall index as follows:

$$NRPBM_S = E_t \left\{ 1 - Max \left( \frac{\sum_{x,y} DV_{ver}(x,y,t) \times S(x,y,t)}{\sum_{x,y} DF_{ver}(x,y,t) \times S(x,y,t)}, \frac{\sum_{x,y} DV_{hor}(x,y,t) \times S(x,y,t)}{\sum_{x,y} DV_{ver}(x,y,t) \times S(x,y,t)} \right) \right\} \qquad (5.19)$$

where $DV$ and $DF$ are differences between the original and blurred image, in vertical and horizontal directions [233].

4) <u>Q_Farias</u> [117]: Farias and Akamine proposed two NR metrics for quality assessment of 2D images. These metrics measure the blockiness and blurriness of the image [117]. In this approach, the blurriness metric is defined as the average edge with. The blockiness metric is defined by average horizontal and vertical differences. Saliency based versions of the mentioned metrics are formulated as follows:

$$Blur_S = E_t \{ E_{x,y} (width(x,y,t) \times S(x,y,t)) \} \qquad (5.20)$$

$$Block_S = E_t\left\{\frac{1}{H.W}\left(\frac{\sum_{i=1}^{H/8}\sum_{j=1}^{W}D_v^8(i,j,t)\times S(x,y,t)}{\sum_{i=1}^{H}\sum_{j=1}^{W}D_v(i,j,t)\times S(x,y,t)} + \frac{\sum_{i=1}^{H}\sum_{j=1}^{W/8}D_h^8(i,j,t)\times S(x,y,t)}{\sum_{i=1}^{H}\sum_{j=1}^{W}D_h(i,j,t)\times S(x,y,t)}\right)\right\}$$ (5.21)

where $H$ and $W$ are height and width of the image, $D_h$ and $D_v$ are the horizontal and vertical brightness differences, and $D_h^8$ and $D_v^8$ are brightness differences at the borders of $8\times 8$ block partitions.

5) <u>Q_Sadaka</u> [118]: Sadaka et al. designed an image sharpness metric based on Just Noticeable Blur (JNB) and 2D image saliency maps [118]. In this method, the sharpness metric is defined as:

$$M = E_t\left\{\left(\sum_{R\in I}\left|D_R\cdot\frac{\sum_{i\in R}S(x,y,t)}{\sum_{i\in I}S(x,y,t)}\right|^{\beta}\right)^{\frac{-1}{\beta}}\right\}$$ (5.22)

where $D_R$ indicates the amount of perceived blur in the region $R$ [118]. Here, we substitute the 2D image saliency maps used in (5.22) with our 3D video saliency maps, which are proven to provide superior stereo saliency detection [124].

6) <u>VQSM</u> [119]: Visual Quality Saliency based Metric (VQSM) is a 2D NR image quality metric which measures the sharpness and smoothness of an image [119]. We modify the sharpness and smoothness components using the available 3D saliency information as follows:

$$Q_{sh} = E_{x,y}\left\{\sqrt{g_x^2(x,y,t)+g_y^2(x,y,t)}\times S(x,y,t)\right\}$$ (5.23)

$$Q_{sm} = E_{x,y}\left\{\sigma_I^{5\times 5}\times \overline{S}^{5\times 5}\right\}$$ (5.24)

where $g_x$ and $g_y$ are the gradient values of image brightness component in horizontal and vertical directions, $\sigma_I^{5\times 5}$ is standard deviation over a $5\times 5$ window centered at $(x,y)$, and $\overline{S}$ is the average saliency values over the same window. The overall quality index is calculated using the same method as VQSM, and averaged over the frames:

$$VQSM_S = E_t\left\{\alpha_1 Q_{sh}^2 + \alpha_2 Q_{sh} + \alpha_3 Q_{sm}^2 + \alpha_4 Q_{sm} + \alpha_5\right\}$$ (5.25)

7) AQI [234]: Anisotropy Quality Index (AQI) is a 2D blind image quality metric which is based on measuring the variance of the expected entropy of an image upon a set of predefined directions [234]. We incorporate saliency probabilities of pixels into the entropy used by AQI index as weighting coefficients.

8) QA3D [26]: QA3D is a NR video quality metric for stereoscopic images which is designed to assess the transmission artifacts (blockiness, sharpness, edginess) [26]. In this method, first a hard threshold is applied to the disparity maps of the reference and distorted pair to set the disparity values smaller than threshold to zero. Then, a disparity index for frame n is defined for each frame by:

$$D_n = E_{x,y}\{D(x,y)\} \quad (5.26)$$

The disparity index is used to define a dissimilarity index between the views:

$$S_m = \frac{1}{10}\left(\sum_{i=n-p}^{n-1} D_i - D_n P\right) \times D_n \quad (5.27)$$

where $p$ is the number of previous frame used. This index along with an edge based difference measure ($D_E$) form the overall QA3D:

$$QA3D = E_t\left\{1 - \frac{S_m + D_E}{2}\right\} \quad (5.28)$$

Saliency information is incorporated in this metric both in the dissimilarity index $S_m$ and difference index $D_E$ as spatial weighting coefficients.

9) NOSPDM [27]: Gu et al. proposed a saliency based parallax compensation based distortion metric (NOSPDM) for JPEG compressed stereoscopic images [27] defined as (for each stereo pair):

$$NOSPDM = (2 - \mu_R)QJPEG_L + \mu_R QJPEG_R$$
$$- \lambda \max\{QJPEG_L, QJPEG_R\} + \cos^{-1}\left(\frac{L.R}{\|L\|_2 \|R\|_2}\right) + \omega_s . \cos^{-1}\left(\frac{S_L.S_R}{\|S_L\|_2 \|S_R\|_2}\right) \quad (5.29)$$

where $L$ and $R$ are the two view images, $S_L$ and $S_R$ are 2D saliency maps, $\mu_R$, $\omega_S$, and $\lambda$ are constant parameters, and QJPEG measures the sharpness of each view as follows:

$$QJPEG = \alpha + \beta\left(\frac{B_h + B_v}{2}\right)^{\gamma_1}\left(\frac{A_h + A_v}{2}\right)^{\gamma_2}\left(\frac{Z_h + Z_v}{2}\right)^{\gamma_3} \quad (5.30)$$

where α, β, $\gamma_1$, $\gamma_2$, $\gamma_3$ are constant parameters, $B_h$ and $B_v$ are blockiness across horizontal and vertical edges, $A_h$ and $A_v$ are the average absolute difference between in-block image samples in horizontal and vertical directions, and $Z_h$ and $Z_v$ are the horizontal and vertical zero crossing rates [27]. We use the 3D saliency information in the QJPEG terms for the two views in each of the three components as weights to the pixel values and zero crossings.

10) Q_Ryu [28]: Ryu and Sohn proposed a NR quality metric for stereoscopic images which takes into account blurriness and blockiness of an image pair [28]. In this approach, a pair of blurriness and blockiness maps are generated for each view, and combined with 2D saliency maps generated from each of the views. We substitute the 2D saliency maps used in this approach with our 3D saliency maps.

## 5.2 Experiments

We modify the FR and NR quality metrics described in Section 5.1 using stereo saliency information and evaluate the performance of the modified metrics in comparison to the original metrics. This section reviews the incorporated video database in the experiments, as well as the subjective tests procedure.

Since verifying the proposed saliency integration method requires subjective and objective quality values from stereoscopic videos along with their corresponding distorted sequences, we use the video database presented in Chapter 2, Section 2.2.1. The only difference is that, in Chapter 2, we divided the videos into training and validation sets for HV3D performance evaluation. Here, there is no need for dividing the video database into those sets as no training is involved. Therefore, we use the entire video database for our performance evaluations in this section. As a result 16 stereo videos and their corresponding distorted videos presented in Section 2.2.1 construct our dataset in this chapter. Table 5.1, Table 5.2, and Fig. 5.4 show the details about the videos.

In addition, we use the same subjective MOS values collected through the experiments of Section 2.2.2, for both the training and validation videos.

Table 5.1 Stereoscopic video database

| Sequence | Resolution | Frame Rate (fps) | Number of Frames | Spatial Complexity (Spatial Information) | Temporal Complexity (Temporal Information) | Depth Range (*cm*) |
|---|---|---|---|---|---|---|
| **Poznan_Hall2** | 1920×1088 | 25 | 200 | Low (35.4658) | Low (11.1460) | High (28.93) |
| **Undo_Dancer** | 1920×1088 | 25 | 250 | High (81.0423) | High (26.9021) | High (30.69) |
| **Poznan_Street** | 1920×1088 | 25 | 250 | High (95.3103) | High (26.5562) | High (34.01) |
| **GT_Fly** | 1920×1088 | 25 | 250 | Medium (58.8022) | High (33.0102) | High (31.02) |
| **Cokeground** | 1000×540 | 30 | 210 | High (86.9096) | Medium (15.9128) | Low (4.99) |
| **Ball** | 1000×540 | 30 | 150 | Medium (49.7701) | Low (13.3074) | Medium (15.53) |
| **Kendo** | 1024×768 | 30 | 300 | Medium (47.2172) | High (26.8791) | High (21.39) |
| **Balloons** | 1024×768 | 30 | 500 | Medium (48.6726) | High (21.4660) | Low (5.84) |
| **Lovebird1** | 1024×768 | 30 | 300 | Medium (59.2345) | Low (13.8018) | Medium (15.01) |
| **Newspaper** | 1024×768 | 30 | 300 | High (65.1173) | Medium (17.1297) | Low (5.09) |
| **Soccer2** | 720×480 | 30 | 450 | High (115.2781) | High (28.6643) | Medium (16.99) |
| **Alt-Moabit** | 512×384 | 30 | 100 | High (111.0437) | High (21.2721) | Medium (13.36) |
| **Hands** | 480×270 | 30 | 251 | High (114.6755) | High (25.2551) | Medium (15.86) |
| **Flower** | 480×270 | 30 | 112 | Medium (43.0002) | Low (13.5305) | Low (5.86) |
| **Horse** | 480×270 | 30 | 140 | High (85.4988) | High (22.3184) | Medium (13.56) |
| **Car** | 480×270 | 30 | 235 | Medium (49.6162) | Medium (16.0197) | High (24.21) |

Table 5.2 Different types of distortions

| Artifact / Distortion | Description | Parameters | Affects views separately | Affects both views together |
|---|---|---|---|---|
| **AWGN** | Additive White Gaussian Noise | zero mean and variance value 0.01 | X | |
| **Blur** | GLPF: Gaussian Low Pass Filter | size 4 and the standard deviation of 4 | X | |
| **Intensity Shift** | Increased brightness values | Increment by 20 (out of 255) | X | |
| **Simulcast Coding** | Simulcast compression of the views | HEVC HM 9.2 [150], GOP 4, QP 35, 40, Low Delay configuration profile | X | |
| **Disparity Map Compression** | Synthesizing views using a highly compressed disparity map | HEVC HM 9.2 [150], GOP 4, QP 25, 45, Low Delay configuration profile | | X |
| **3D Video Compression** | 3D video compression | HEVC based 3D HTM 9 [150], GOP 8, QP 25, 30, 35, 40, Random Access High Efficiency profile | | X |
| **View Synthesis** | Synthesizing one view | Using VSRS 3.5 [151] for synthesizing one view based on disparity map and the other view | | X |

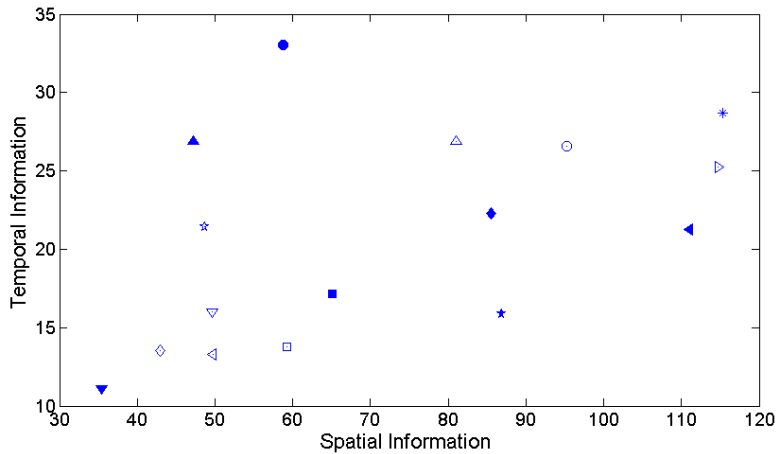

**Figure 5.4** Spatial and temporal information distribution for the stereo video database used in our experiments

## 5.3 Results and Discussions

Performance of objective FR and NR metrics are calculated for our stereo video database using the mentioned performance metrics and compared with that of the same metrics when stereo saliency maps are taken into account. Table 5.3 and Table 5.4 show the performance of FR and NR metrics for the original metrics and when the stereo saliency information is incorporated.

It is observed from these tables that saliency prediction in general improves the performance of both objective FR and NR metrics. In the case of FR metrics, the improvements are on average less than the NR case. This is due to the fact that information from the reference video is available for FR quality assessment and thus more accurate assessment is possible. It is worth noting that some of the NR metrics evaluated in our study incorporate 2D saliency maps in their original design. These metrics (IQVG [116], Q_Farias [117], Q_Sadaka [118], VQSM [119], NOSPDM [27], and Q_Ryu [28]) receive less improvement after being modified by stereo saliency information.

In addition to incorporating stereo saliency information generated using the LVBS-3D VAM in the FR and NR quality assessment tasks, we examine the added value of saliency information resulted from several other state-of-the-art 3D VAMs namely 3D VAMs of Fang et al. [68], Coria et al. [146], and Park et al. [172].

Following what is considered a common practice in saliency prediction studies, we also add to the evaluations the results from the 2D VAM of Itti et al. [57]. Fig. 5.5 shows the improvements in quality assessment achieved by using various VAMs. It is observed from Fig. 5.5 that 3D VAM of LBVS-3D has resulted in highest improvements in quality assessment. Moreover, NR metrics that already use saliency information receive less improvement compared to other NR metrics.

We further study the performance improvements by using stereo saliency information for each type of distortion separately. Table 5.5, Table 5.6, and Table 5.7 contain the PCC values for FR and NR metrics before and after incorporation of stereo saliency maps. Fig. 5.6 shows the PCC improvements in the case FR and NR quality assessment for different kinds of distortions. It is observed from Table 5, Table 5.6, Table 5.7, and Fig. 5.6 that different kinds of distortions receive a roughly similar amount of improvement, except AWGN for FR case, which could be due to less accurate disparity map estimation from the distorted videos for this type of distortion.

## 5.4 Conclusion

In this chapter we study the added value of using stereo saliency prediction in full-reference and no-reference quality assessment tasks. To this end, we leverage the stereo saliency prediction results to modify FR and NR quality metrics and re-evaluate their performance. We measure the performance improvements using a large database of stereoscopic videos with several representative types of distortions. Performance evaluations revealed that using stereo saliency in general improves the quality assessment accuracy. However, the improvements are more significant in the case of NR video quality assessment.

Table 5.3 Performance of different FR quality metrics with and without saliency map integration

| Quality Metric | Distortion | PCC | | SCC | | RMSE | | Outlier Ratio | |
|---|---|---|---|---|---|---|---|---|---|
| | | Original | Saliency Inspired | Original | Saliency Inspired | Original | Saliency Inspired | Original | Saliency Inspired |
| PSNR | | 0.6454 | 0.6800 | 0.6350 | 0.6646 | 10.388 | 9.671 | 0.0167 | 0.0083 |
| SSIM [13] | | 0.6844 | 0.7113 | 0.6213 | 0.6946 | 9.852 | 9.057 | 0.0083 | 0.0083 |
| MS-SSIM [157] | | 0.7071 | 0.7219 | 0.7180 | 0.7481 | 9.999 | 9.009 | 0.0083 | 0.0083 |
| VIF [135] | | 0.7257 | 0.7380 | 0.7204 | 0.7349 | 9.166 | 8.947 | 0 | 0 |
| Ddl1 [16] | | 0.7370 | 0.7638 | 0.7321 | 0.7557 | 8.732 | 8.556 | 0 | 0 |
| OQ [17] | | 0.7580 | 0.7709 | 0.7900 | 0.7993 | 8.610 | 8.500 | 0 | 0 |
| CIQ [18] | | 0.7200 | 0.7451 | 0.7080 | 0.7346 | 9.446 | 8.884 | 0.0083 | 0 |
| PHVS3D [19] | | 0.7837 | 0.8022 | 0.8233 | 0.8238 | 8.420 | 8.300 | 0 | 0 |
| PHSD [9] | | 0.7911 | 0.8234 | 0.7841 | 0.8010 | 8.321 | 8.067 | 0 | 0 |
| MJ3D [21] | | 0.8640 | 0.8698 | 0.8947 | 0.9033 | 7.229 | 7.178 | 0 | 0 |
| Q_Shao [20] | | 0.8348 | 0.8524 | 0.7988 | 0.8349 | 7.902 | 7.436 | 0 | 0 |
| HV3D [126] | | 0.9082 | 0.9231 | 0.9130 | 0.9343 | 6.433 | 6.267 | 0 | 0 |

Table 5.4 Performance of different NR quality metrics with and without saliency map integration

| Quality Metric | Distortion | PCC | | SCC | | RMSE | | Outlier Ratio | |
|---|---|---|---|---|---|---|---|---|---|
| | | Original | Saliency Inspired | Original | Saliency Inspired | Original | Saliency Inspired | Original | Saliency Inspired |
| IQVG [116] | | 0.6713 | 0.6805 | 0.6892 | 0.6956 | 9.923 | 9.901 | 0.0083 | 0.0083 |
| GBIM [232] | | 0.6065 | 0.6538 | 0.5897 | 0.6251 | 10.849 | 10.102 | 0.0167 | 0.0083 |
| NRPBM [233] | | 0.5980 | 0.6634 | 0.6001 | 0.6678 | 10.963 | 10.038 | 0.0167 | 0.0083 |
| Q_blur_Farias [117] | | 0.6312 | 0.6449 | 0.6229 | 0.6437 | 10.430 | 10.321 | 0.0083 | 0.0083 |
| Q_block_Farias [117] | | 0.6494 | 0.6523 | 0.6550 | 0.6591 | 10.432 | 10.277 | 0.0083 | 0.0083 |
| Q_Sadaka [118] | | 0.6668 | 0.6878 | 0.6790 | 0.6889 | 9.993 | 9.911 | 0.0083 | 0.0083 |
| VQSM [119] | | 0.6903 | 0.7009 | 0.6945 | 0.7178 | 8.987 | 8.690 | 0.0083 | 0 |
| AQI [234] | | 0.6882 | 0.7426 | 0.6721 | 0.7448 | 8.995 | 8.766 | 0.0083 | 0 |
| QA3D [26] | | 0.7127 | 0.7633 | 0.7089 | 0.7467 | 8.680 | 8.012 | 0 | 0 |
| NOSPDM [27] | | 0.7843 | 0.7919 | 0.7911 | 0.7999 | 7.943 | 7.903 | 0 | 0 |
| Q_Ryu [28] | | 0.8475 | 0.8533 | 0.8410 | 0.8557 | 7.687 | 7.559 | 0 | 0 |

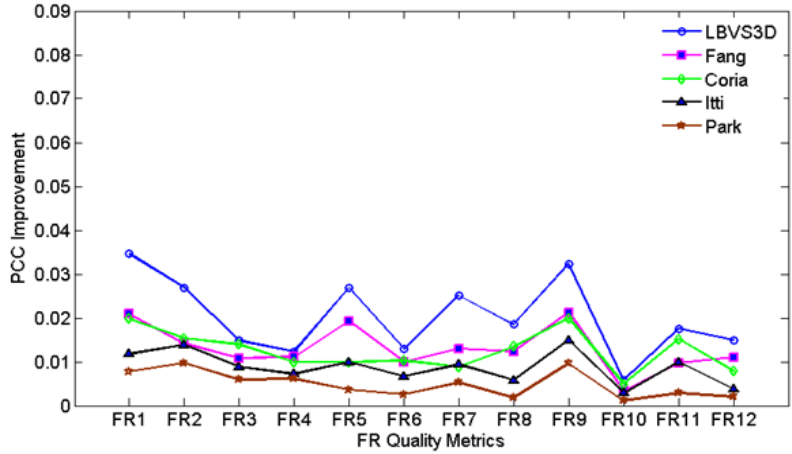

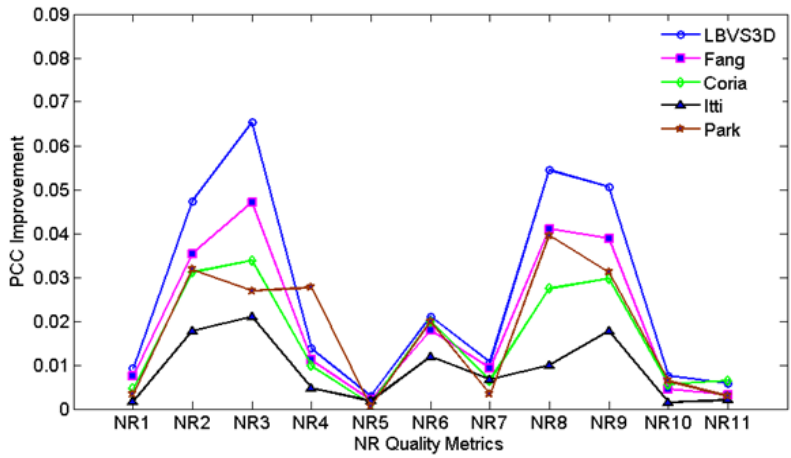

**Figure 5.5** PCC improvements by integration of different VAMs to: (a) FR and (b) NR metrics. FR metrics from left to right: PSNR, SSIM [13], MS-SSIM [157], VIF [135], Ddl1 [16], OQ [17], CIQ [18], PHVS3D [19], PHSD [9], MJ3D [21], Q_Shao [20], HV3D [129], and NR metrics from left to right: IQVG [116], GBIM [232], NRPBM [233], Q_blur_Farias [117], Q_block_Farias [117], Q_Sadaka [18], VQSM [119], AQI [234], QA3D [26], NOSPDM [27], Q_Ryu [28]

**Table 5.5** PCC values for FR quality metrics and for each specific type of distortion. 'Q-ref' denotes the original quality metric, while 'Q-sal' denotes the corresponding saliency inspired quality metric

| Quality Metric | Distortion | AWGN | | Simulcast Compression | | Blurring | | Brightness Shift | | 3D Video Compression | | View Synthesis | | Depth Map Compression | |
|---|---|---|---|---|---|---|---|---|---|---|---|---|---|---|---|
| | | Q-ref | Q-sal | Q-ref | Q-sal | Q-ref | Q-sal | Q-ref | Q-sal | Q-ref | Q-sal | Q-ref | Q-sal | Q-ref | Q-sal |
| PSNR | | 0.683 | 0.689 | 0.710 | 0.736 | 0.592 | 0.609 | 0.493 | 0.510 | 0.740 | 0.749 | 0.639 | 0.640 | 0.650 | 0.659 |
| SSIM [13] | | 0.772 | 0.781 | 0.753 | 0.773 | 0.716 | 0.734 | 0.765 | 0.787 | 0.670 | 0.721 | 0.654 | 0.668 | 0.670 | 0.683 |
| MS-SSIM [157] | | 0.789 | 0.789 | 0.777 | 0.789 | 0.821 | 0.835 | 0.837 | 0.845 | 0.720 | 0.727 | 0.623 | 0.641 | 0.749 | 0.764 |
| VIF [135] | | 0.797 | 0.807 | 0.769 | 0.776 | 0.823 | 0.840 | 0.852 | 0.872 | 0.731 | 0.742 | 0.606 | 0.610 | 0.754 | 0.761 |
| Ddl1 [16] | | 0.623 | 0.632 | 0.762 | 0.782 | 0.743 | 0.754 | 0.776 | 0.781 | 0.720 | 0.728 | 0.726 | 0.737 | 0.813 | 0.826 |
| OQ [17] | | 0.713 | 0.718 | 0.686 | 0.693 | 0.704 | 0.718 | 0.652 | 0.661 | 0.741 | 0.750 | 0.682 | 0.693 | 0.707 | 0.711 |
| CIQ [18] | | 0.777 | 0.783 | 0.774 | 0.796 | 0.833 | 0.838 | 0.724 | 0.729 | 0.756 | 0.771 | 0.712 | 0.724 | 0.770 | 0.792 |
| PHVS3D [19] | | 0.692 | 0.699 | 0.796 | 0.818 | 0.724 | 0.731 | 0.608 | 0.618 | 0.819 | 0.829 | 0.737 | 0.746 | 0.784 | 0.800 |
| PHSD [9] | | 0.648 | 0.655 | 0.852 | 0.876 | 0.752 | 0.764 | 0.625 | 0.635 | 0.833 | 0.857 | 0.753 | 0.768 | 0.829 | 0.834 |
| MJ3D [21] | | 0.828 | 0.833 | 0.845 | 0.860 | 0.812 | 0.819 | 0.860 | 0.868 | 0.801 | 0.820 | 0.722 | 0.739 | 0.702 | 0.715 |
| Q_Shao [20] | | 0.799 | 0.806 | 0.823 | 0.848 | 0.828 | 0.836 | 0.781 | 0.796 | 0.792 | 0.810 | 0.709 | 0.725 | 0.725 | 0.745 |
| HV3D [126] | | 0.799 | 0.808 | 0.831 | 0.858 | 0.811 | 0.818 | 0.841 | 0.850 | 0.897 | 0.907 | 0.888 | 0.900 | 0.860 | 0.881 |

Table 5.6 PCC values for NR quality metrics and for each specific type of distortion. 'Q-ref' denotes the original quality metric, while 'Q-sal' denotes the corresponding saliency inspired quality metric

| Quality Metric Distortion | AWGN | | Simulcast Compression | | Blurring | | Brightness Shift | | 3D Video Compression | | View Synthesis | | Depth Map Compression | |
|---|---|---|---|---|---|---|---|---|---|---|---|---|---|---|
| | Q-ref | Q-sal | Q-ref | Q-sal | Q-ref | Q-sal | Q-ref | Q-sal | Q-ref | Q-sal | Q-ref | Q-sal | Q-ref | Q-sal |
| IQVG [116] | 0.557 | 0.561 | 0.681 | 0.690 | 0.693 | 0.700 | 0.488 | 0.501 | 0.650 | 0.663 | 0.558 | 0.571 | 0.688 | 0.694 |
| GBIM [232] | 0.525 | 0.575 | 0.598 | 0.634 | 0.606 | 0.644 | 0.479 | 0.512 | 0.601 | 0.634 | 0.572 | 0.607 | 0.616 | 0.655 |
| NRPBM [233] | 0.512 | 0.558 | 0.598 | 0.645 | 0.606 | 0.656 | 0.495 | 0.535 | 0.603 | 0.638 | 0.577 | 0.560 | 0.567 | 0.588 |
| Q_blur_Farias [117] | 0.599 | 0.602 | 0.623 | 0.630 | 0.666 | 0.670 | 0.589 | 0.591 | 0.635 | 0.639 | 0.611 | 0.618 | 0.616 | 0.620 |
| Q_block_Farias [117] | 0.580 | 0.589 | 0.646 | 0.651 | 0.645 | 0.650 | 0.568 | 0.570 | 0.634 | 0.638 | 0.605 | 0.613 | 0.657 | 0.673 |
| Q_Sadaka [118] | 0.613 | 0.618 | 0.698 | 0.704 | 0.702 | 0.709 | 0.636 | 0.640 | 0.663 | 0.669 | 0.634 | 0.651 | 0.667 | 0.672 |
| VQSM [119] | 0.645 | 0.656 | 0.687 | 0.691 | 0.700 | 0.709 | 0.634 | 0.642 | 0.680 | 0.687 | 0.657 | 0.666 | 0.687 | 0.691 |
| AQI [234] | 0.662 | 0.693 | 0.671 | 0.713 | 0.682 | 0.726 | 0.635 | 0.657 | 0.668 | 0.670 | 0.611 | 0.646 | 0.689 | 0.721 |
| QA3D [26] | 0.689 | 0.713 | 0.727 | 0.768 | 0.759 | 0.793 | 0.689 | 0.702 | 0.712 | 0.745 | 0.689 | 0.734 | 0.668 | 0.701 |
| NOSPDM [27] | 0.747 | 0.756 | 0.809 | 0.817 | 0.816 | 0.821 | 0.737 | 0.744 | 0.777 | 0.781 | 0.751 | 0.758 | 0.783 | 0.793 |
| Q_Ryu [28] | 0.761 | 0.770 | 0.867 | 0.875 | 0.861 | 0.881 | 0.691 | 0.702 | 0.808 | 0.816 | 0.673 | 0.680 | 0.751 | 0.759 |

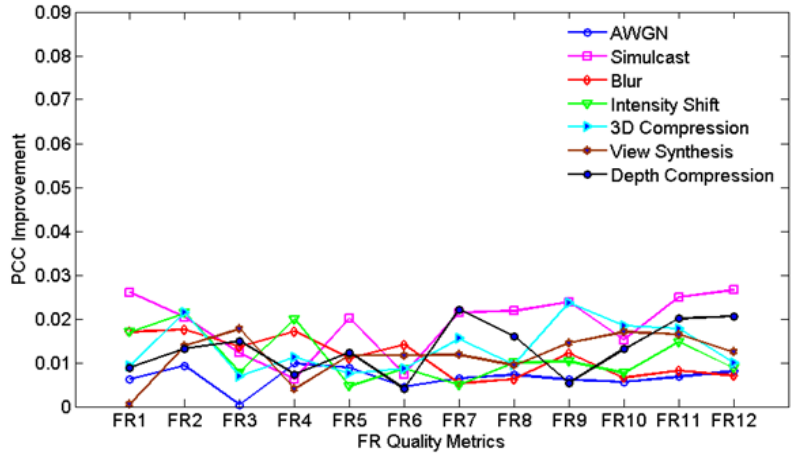

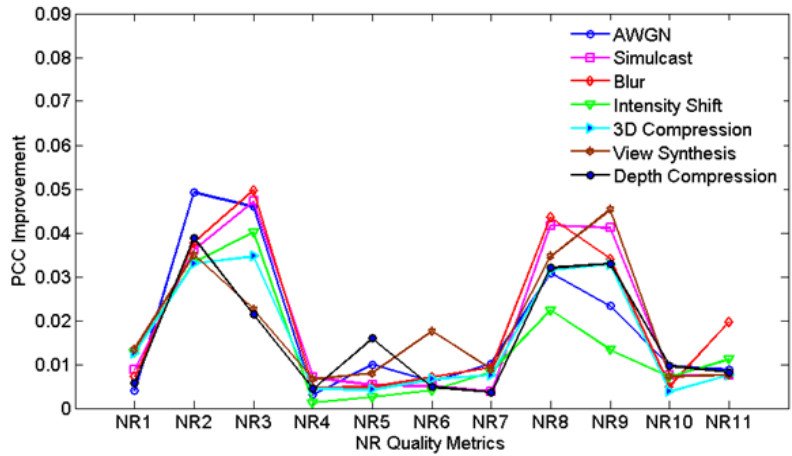

Figure 5.6 PCC improvements resulting from the integration of LBVS-3D VAM to: (a) FR and (b) NR quality metrics. FR metrics from left to right: PSNR, SSIM [13], MS-SSIM [157], VIF [135], Ddl1 [16], OQ [17], CIQ [18], PHVS3D [19], PHSD [9], MJ3D [21], Q_Shao [20], HV3D [129], and NR metrics from left to right: IQVG [116], GBIM [232], NRPBM [233], Q_blur_Farias [117], Q_block_Farias [117], Q_Sadaka [18], VQSM [119], AQI [234], QA3D [26], NOSPDM [27], Q_Ryu [28]

**Table 5.7 Average PCC improvement for FR and NR quality metrics and for each specific type of distortion**

| Quality Metric | Distortion | AWGN | Simulcast Compression | Blurring | Brightness Shift | 3D Video Compression | View Synthesis | Depth Map Compression |
|---|---|---|---|---|---|---|---|---|
| FR | | 0.0066 | 0.0189 | 0.0113 | 0.0113 | 0.0133 | 0.0118 | 0.0132 |
| NR | | 0.0183 | 0.0192 | 0.0202 | 0.0142 | 0.0162 | 0.0188 | 0.0161 |

# 6 Conclusions and Future Work

## 6.1 Summary of the Contributions

This thesis investigates various aspects of the quality assessment of stereoscopic 3D video. We propose a novel method for full-reference quality assessment of stereo video in Chapter 2. This method relies on the combination of the quality of cyclopean view and that of the depth map. The performance of the proposed quality metric is verified through large-scale subjective experiments.

To study the no-reference aspects of quality assessment in 3D video, various attributes are studied separately in Chapter 3 and Chapter 4, and the results are incorporated in the no-reference quality assessment method proposed in Chapter 5. Due to the importance that temporal aspects have on viewing quality, we investigate the frame rate separately in Chapter 3. We study the effect of 3D video frame rate on visual quality and bit rate in Chapter 3, where we analyze the relationship between frame rate, bitrate, and the 3D QoE through extensive subjective tests. We capture a database of stereo videos at different frame rates and used this database to derive conclusions and make proposals for 3D video transmission. Our database is made publicly available to facilitate further research in this area.

In Chapter 4, we design a visual attention model for stereo video content that starts with a rough segmentation and quantifies several intuitive observations such as the effects of visual discomfort level, depth abruptness, motion acceleration, elements of surprise and size, compactness, and sparsity of the salient regions. To calculate local and global features describing these observations, a new fovea-based model of spatial distance between the image regions is used. Then, a random forest based algorithm is utilized to learn a model of stereoscopic video saliency so that the various conspicuity maps generated by our method are efficiently fused into one single saliency map, which delivers high correlation with the eye-fixation data. The performance of the proposed saliency model is evaluated against the results of a large-scale eye-tracking experiment, which involves 24 subjects and an in-house database of 61 captured stereoscopic videos. The video saliency benchmark database is publicly available, for further research and development by the research community [125].

Finally, in Chapter 5, we use the results of our findings in Chapters 3 and 4, to integrate the stereo saliency prediction in the design of various quality metrics. Saliency information is not only integrated to various no-reference quality metrics, but it is also used to improve the performance of full-reference quality metrics. We verify the efficiency of the proposed framework through subjective tests.

## 6.2 Significance and Potential Applications of the Research

The research presented in this thesis aims at assessing visual quality of stereoscopic 3D video content. The full-reference quality metric proposed in Chapter 2 has the potential to be used in the compression of 3D video, where it can help customize video coding to ensure efficient resource management at the encoder side. Its implementation in H.264/AVC 3D standard known as 3D-AVC or the HEVC 3D video extension and the HEVC Multiview Video plus Depth (MVD) format. In addition, any automatic stereo video quality evaluation process can incorporate this metric to gain improvements on performance.

The findings of our study about 3D video frame rate, in Chapter 3, are helpful in defining bitrate adaptation guidelines for 3D video delivery over variable bitrate networks. These guidelines will allow network providers to change 3D content frame rate in order to deal with bandwidth capacity changes so that viewers' quality of experience is not significantly affected.

Our visual attention model of Chapter 4 has various potential applications in: 1) robot vision, where automatic 3D saliency prediction is required, 2) video surveillance, where a salient object is to be tracked, 3) region-of-interest 3D video compression where bit allocation is done according to objects' degree of saliency, 4) reframing the 3D video, 5) stereo cameras, and finally 6) quality assessment of stereo video, where it is integrated to the framework of full-reference and no-reference quality metrics (Chapter 5).

## 6.3 Future Work

Regarding the proposed full-reference quality metric for stereo video in Chapter 2, future work includes extending the proposed frame work for multiview video streams. Combining the quality of different views and depth maps is a challenge. In addition, the proposed quality metric can be used to customize 3D video compression, to ensure an efficient resource allocation in the encoder.

Our findings in Chapter 3 help us understand the effect of each frame rate on the bit rate and subjective 3D quality. These findings can be used in 3D video transmission, to design efficient bit rate adaptation schemes for variable bit rate channels. Another interesting problem is to study the effect of frame rate in auto-stereo and free-viewing displays, as higher volume of data is presented to viewers in these cases.

Future work related to Chapter 4 includes incorporation of the proposed 3D visual attention model in region-of-interest video compression. In addition, studying the impact of depth and other saliency attributes in multiview and free-view displays is a challenging problem, which can be based on our work presented in Chapter 4.

Chapter 5 can be extended by using the saliency information to boost the overall quality of users' experience for immersive media technologies such as gaming, $360^\circ$ displays such as oculus, and in general virtual reality application. For each one of these technologies, special considerations have to be taken into account so that the saliency information extracted from the content is efficiently integrated to the system, yielding an improved quality of experience.

[137] K. Lee, J. Park, S. Lee, and A. C. Bovik, "Temporal pooling of video quality estimates using perceptual motion models," *IEEE 17th International on Image Processing, ICIP* 2010, Hong Kong.

[138] K. Seshadrinathan, and A. C. Bovik, "Temporal hysteresis model of time varying subjective video quality," *IEEE International Conference on Acoustic, Speech, and Signal Processing, ICASSP* 2011.

[139] 3D video database at Digital Multimedia Lab, University of British Columbia, available since May 2014 at: http://dml.ece.ubc.ca/data/hv3d/

[140] Hamberg and H. de Ridder, "Time-varying image quality: modeling the relation between instantaneous and overall quality," *Society of Motion Picture & Television Engineers (SMTPE) Journal*, Vol. 108 (1999), pp. 802-811.

[141] A. Banitalebi-Dehkordi, M. T. Pourazad, and P. Nasiopoulos, "Effect of eye dominance on the perception of stereoscopic 3D video," *International Conference on Image Processing, ICIP*, 2014.

[142] ISO/IEC JTC1/SC29/WG11, "Common Test Conditions for HEVC- and AVC-based 3DV", Doc. N12352, Switzerland, November 2011.

[143] Recommendation ITU P.910, "Subjective video quality assessment methods for multimedia applications," *ITU*, 1999.

[144] I. Sobel, "History and definition of the Sobel operator", 2014.

[145] D. Xu, L. E. Coria, and P. Nasiopoulos, "Guidelines for an improved quality of experience in 3D TV and 3D mobile displays," *Journal of the Society for Information Display*, vol. 20, no. 7, pp. 397-407, July 2012, doi:10.1002/jsid.99.

[146] L. Coria, D. Xu, and Panos Nasiopoulos, "Automatic stereoscopic video reframing," *3DTV-Con: The True Vision - Capture, Transmission and Display of 3D Video*, Oct. 2012.

[147] ISO/IEC JTC1/SC29/WG11, "3D-HEVC Test Model 3," *N13348*, January 2013, Geneva.

[148] G. J. Sullivan, J. Ohm, W. J. Han, T. Wiegand, "Overview of the high efficiency video coding (HEVC) standard," *IEEE Transactions of Circuits and Systems on Video Technology*, vol. 22, issue 12, pp. 1649-1668, 2012.

[149] M. T. Pourazad, C. Doutre, M. Azimi, and P. Nasiopoulos, "HEVC: The new gold standard for video compression: How does HEVC compare with H.264/AVC," *IEEE Consumer Electronics Magazine*, vol. 1, issue 3, pp. 36-46, 2012.

[150] HEVC at HHI Fraunhofer (Retrieved at May 2015): https://hevc.hhi.fraunhofer.de/

[151] "View synthesis reference software (VSRS) 3.5," wg11.sc29.org, March 2010.

[152] Recommendation ITU-R BT.500-13, "Methodology for the subjective assessment of the quality of the television pictures", 2012.

[153] ISO/IEC JTC1/SC29/WG11 (MPEG), Document N12036, "Call for proposals on 3D video coding technology," *96th MPEG meeting*, Geneva, March 2011.
134

[172] Y. Park, B. Lee, W. Cheong, and N. Hur, "Stereoscopic 3D visual attention model considering comfortable viewing," *IET Conference on Image Processing (IPR 2012)*, pp. 1-5, 2012.

[173] I. Iatsun, M-C. Larabi, Ch. Fernandez-Maloigne, "Using monocular depth cues for modeling stereoscopic 3D saliency," *ICIP* 2014.

[174] R. Ju, L. Ge, W. Geng, T. Ren, and G. Wu, "Depth saliency based on anisotropic center-surround difference," *ICIP* 2014.

[175] X. Fan, Z. Liu, and G. Sun, "Salient region detection for stereoscopic images," *19$^{th}$ International Conference on Digital Signal Processing*, 2014, Hong Kong.

[176] A. Banitalebi-Dehkordi, E. Nasiopoulos, M. T. Pourazad, and Panos Nasiopoulos, "Benchmark 3D eye-tracking dataset for visual saliency prediction on stereoscopic 3D video," *SPIE Journal of Electronic Imaging*, Under Review, 2015.

[177] D. C. Bourassa, I. C. McManus, and M. P. Bryden, "Handedness and eye-dominance: a meta-analysis of their relationship," *Laterality*, vol. 1, no. 1, 1996, pp. 5-34.

[178] D. Comanicu and P. Meer, "Mean shift: A robust approach toward feature space analysis," *IEEE PAMI,* vol. 24, pp. 603-619, May 2002.

[179] F.M. Adams and C.E. Osgood, "A cross-cultural study of the affective meanings of color," *Journal of Cross-Cultural Psychology*, vol. 4, 1973.

[180] E.F. Schubert, "Light emitting diodes," *2$^{nd}$ Ed. Cambridge University Press*, 2006.

[181] M. Baik et al., "Investigation of eye-catching colors using eye tracking," Proc. of *SPIE-IS&T Electronic Imaging, SPIE* vol. 8651, 2013.

[182] M. Tian, S. Wan, and L. Yue, "A Color saliency model for salient objects detection in natural scenes," *Advances in Multimedia Modeling, Lecture Notes in Computer Science* vol. 5916, 2010, pp 240-250.

[183] E. Lübbe, "Colours in the mind - colour systems in reality," 2010.

[184] T. Erdogan, How to calculate luminosity, dominant wavelength, and excitation purity," *Semrock White Paper Series*.

[185] E.D. Gelasca, D. Tomasic, and T. Ebrahimi, "Which colors best catch your eyes: a subjective study of color saliency," *ISCAS* 2005.

[186] M. Drulea and S. Nedevschi, "Motion estimation using the correlation transform," *IEEE TIP*, vol.22, no.8, pp.3260-3270, Aug. 2013.

[187] A. Banitalebi-Dehkordi, M.T. Pourazad, and Panos Nasiopoulos, "The effect of frame rate on 3D video quality and bitrate," *Springer Journal of 3D Research*, vol. 6:1, pp. 5-34, March 2015, DOI 10.1007/s13319-014-0034-3.

[188] J.Y. Lin, S. Franconeri, and J.T. Enns, "Objects on a collision path with the observer demand attention," *Psychology Science,* vol. 19, 2008.

[189] Y. Liu, L.K. Cormack, and A.C. Bovik, "Natural scene statistics at stereo fixations," *2010 Symposium on Eye-Tracking Research & Applications, ETRA* 2010.

[209] Y. Rubner, C. Tomasi, and L. J. Guibas, "The earth movers distance as a metric for image retrieval," *International Journal of Computer Vision*, 40:2000, 2000.

[210] O. Pele and M. Werman, "Fast and robust earth mover's distances," *ICCV* 2009.

[211] K. P. Burnham and D. R. Anderson, Model selection and multi-model inference: A practical information-theoretic approach. *Springer*. (2nd ed.), p.51.2002.

[212] N. Bruce and J. Tsotsos, "Saliency, attention, and visual search: An information theoretic approach," *Journal of Vision*, 2009.

[213] M. Cerf, E.P. Frady, and C. Koch, "Faces and text attract gaze independent of the task: Experimental data and computer model," *Journal of Vision*, 2009.

[214] L. Itti, "Quantifying the contribution of low-level saliency to human eye movements in dynamic scenes," *Visual Cognition*, vol. pp. 1093-1123, 2005.

[215] D. Berg et al., "Free viewing of dynamic stimuli by humans and monkeys," *Journal of Vision*, vol. 9, pp. 1-15, 2009.

[216] B. Tatler, "The central fixation bias in scene viewing: Selecting an optimal viewing position independently of motor biases and image feature distributions," *Journal of Vision,* vol. 7, pp. 1-17, 2007.

[217] Q. Zhao and C. Koch, "Learning a saliency map using fixated locations in natural scenes," *Journal of Vision*, 11(3), 2011.

[218] Y. Fang, Z. Chen, W. Lin, C.-W. Lin, "Saliency detection in the compressed domain for adaptive image retargeting," *IEEE Transactions on Image Processing*, vol. 21, no. 9, pp. 3888-3901, 2012.

[219] Y. Xie, H. Lu, and M-H. Yang, "Bayesian saliency via low and mid level cues," *IEEE Transactions on Image Processing (TIP)*, vol. 22, no. 5, pp. 1689-1698, 2013.

[220] S-h. Zhong, Y. Liu, F. Ren, J. Zhang, and T. Ren, "Video saliency detection via dynamic consistent spatio-temporal attention modelling," *27th AAAI Conference on Artificial Intelligence*, 2013.

[221] Y. Fang, W. Lin, B-S. Lee, C.T. Lau, Zh. Chen, Ch-W. Lin, "Bottom-up saliency detection model based on human visual sensitivity and amplitude spectrum," *IEEE Transactions on Multimedia* 14(1): 187-198 (2012).

[222] Ch. Yang, L. Zhang, H. Lu, M-H. Yang, and X. Ruan, "Saliency detection via graph-based manifold ranking," *IEEE Conference on Computer Vision and Pattern Recognition (CVPR 2013)*, Portland, June, 2013.

[223] R. Margolin, L. Zelnik-Manor, and A. Tal. "What makes a patch distinct," *CVPR* 2013.

[224] T.N. Vikram, M. Tscherepanow, and B. Wrede, "A saliency map based on sampling an image into random rectangular regions of interest," *Pattern Recognition*, 2012.

[225] L. Zhang, Zh. Gu, and H. Li, "SDSP: a novel saliency detection method by combining simple priors," *ICIP*, 2013.

[226] N. Murray, M. Vanrell, X. Otazu, and C.A. Parraga, "Saliency estimation using a non-parametric vision model," *CVPR* 2011.
138